\documentclass[10pt]{article}

\usepackage{amssymb,amsmath,amsthm,mathrsfs} 
\usepackage{mathtools}
\usepackage{graphicx}
\usepackage{lineno}
\usepackage[font=small,labelfont=bf]{caption} 
\usepackage[authoryear,sort&compress]{natbib}
\usepackage{url}
\usepackage{enumerate}
\usepackage{colordvi,color,pspicture}
\usepackage{psfrag}
\usepackage{rotating}        
\usepackage{epstopdf}
\usepackage{algorithm}       
\usepackage{algorithmic}
\usepackage{multirow}        

\newcommand{\Z}{\mathcal{Z}}
\newcommand{\Y}{\mathcal{Y}}
\newcommand{\X}{\mathcal{X}}

\newcommand{\y}{\mathsf{y}}
\newcommand{\F}{\mathcal{F}}
\newcommand{\mF}{\mathscr F}
\newcommand{\g}{\gamma}

\newcommand{\N}{\mathcal{N}}
\newcommand{\V}{\mathcal{V}}
\newcommand{\A}{\mathcal{A}}
\newcommand{\R}{\mathbb{R}}
\newcommand{\mS}{\mathcal{S}}

\newcommand{\D}{\mathscr D}
\newcommand{\mcS}{\mathscr S}

\newcommand{\lam}{\lambda}
\newcommand{\al}{\alpha}

\DeclareMathOperator{\argmin}{argmin}
\DeclareMathOperator{\argmax}{argmax}

\DeclareMathOperator{\vol}{Vol}

\DeclarePairedDelimiter{\ceil}{\lceil}{\rceil}  
\DeclarePairedDelimiter\floor{\lfloor}{\rfloor}

\theoremstyle{plain}
\newtheorem{theorem}{Theorem}[subsection]
\newtheorem{proposition}{Proposition}[subsection]

\newcommand{\algorithmicbreak}{\textbf{break}}
\newcommand{\BREAK}{\STATE \algorithmicbreak}
\begin{document}

\title{Technical Report \#2: \\ Classification Using Proximity Catch Digraphs}
\author{Artur Manukyan\thanks{Ko\c{c} University, Sar{\i}yer, 34450, Istanbul, Turkey}   
        \thinspace\thinspace \&         
        Elvan Ceyhan\thanks{University of Pittsburgh, Department of Statistics, WWPH 1829, 230 S Bouquet Street, Pittsburgh PA 15260 USA}
}
\date{Mar 15, 2017}
\maketitle

\pagenumbering{arabic} \setcounter{page}{1}

\begin{abstract}
\noindent We employ random geometric digraphs to construct semi-parametric classifiers. These data-random digraphs are from parametrized random digraph families called proximity catch digraphs (PCDs). A related geometric digraph family, class cover catch digraph (CCCD), has been used to solve the class cover problem by using its approximate minimum dominating set. CCCDs showed relatively good performance in the classification of imbalanced data sets, and although CCCDs have a convenient construction in $\R^d$, finding minimum dominating sets is NP-hard and its probabilistic behaviour is not mathematically tractable except for $d=1$. On the other hand, a particular family of PCDs, called \emph{proportional-edge} PCDs (PE-PCDs), has mathematical tractable minimum dominating sets in $\R^d$; however their construction in higher dimensions may be computationally demanding. More specifically, we show that the classifiers based on PE-PCDs are prototype-based classifiers such that the exact minimum number of prototypes (equivalent to minimum dominating sets) are found in polynomial time on the number of observations. We construct two types of classifiers based on PE-PCDs. One is a family of hybrid classifiers depend on the location of the points of the training data set, and another type is a family of classifiers solely based on class covers. We assess the classification performance of our PE-PCD based classifiers by extensive Monte Carlo simulations, and compare them with that of other commonly used classifiers. We also show that, similar to CCCD classifiers, our classifiers are relatively better in classification in the presence of class imbalance.
\end{abstract}

\noindent
{\small {\it Keywords:}  Class cover problem, Delaunay tessellation, Digraph, Domination, Prototype selection, Separability, Support estimation, Delaunay tessellation

\vspace{.25 in}

$^*$corresponding author.\\
\indent {\it e-mail:} artur-man@hotmail.com}


\newpage


\section{Introduction} \label{sec:intro}

Classification methods based on set covering algorithms received considerable attention because of their use in prototype selection \citep{bien2011,cannon2004,angiulli2012}. Prototypes are selected members of a data set so as to attain various tasks including reducing, condensing or summarizing a data set. Many learning methods aim to carry out more than one of these tasks, thereby building efficient learning algorithms \citep{pekalska2006,bien2011}. A desirable prototype set reduces the data set in order to decrease running time, condenses the data set to preserve information, and summarizes the data set for better exploration and understanding. The methods we discuss in this work are considered as decision boundary generators where decisions are made based on class conditional regions, or \emph{class covers}, that are composed of a collection of convex sets, each associated with a prototype \citep{toussaint2002}. The union of such convex sets constitute a region for the class of interest, estimating the support of this class \citep{scholkopf2001}. Support estimates have uses in both supervised and unsupervised learning schemes offering solutions to many problems of machine learning literature \citep{marchette2004}. We propose supervised learning methods, or classifiers, based on these estimates of the supports constructed with a random geometric digraph family called \emph{proximity catch digraphs}. 

Proximity Catch Digraphs (PCDs) are closely related to Class Cover Catch Digraphs (CCCDs) introduced by \cite{priebe2001}, and are vertex-random digraphs defined by the relationship between class-labeled observations. They introduced CCCDs to find graph theoretic solutions to the Class Cover Problem (CCP), and provided some results on the minimum dominating sets and the distribution of the domination number of such digraphs for one dimensional data. The goal of CCP is to find a set of hyperballs (usually Euclidean balls) such that their union encapsulates, or \emph{covers}, a subset of the training data set associated with a particular class, called the target class \citep{cannon2004}. In addition, \cite{priebe:2003b} showed that approximate dominating sets of CCCDs, which were obtained by a greedy algorithm, can be used to establish efficient semi-parametric classifiers. Moreover, \cite{devinney2002} defined random walk CCCDs (RW-CCCD) where balls of class covers are defined in a relaxed manner compared to the previously introduced CCCDs. These digraphs have been used, e.g. in face detection \citep{eveland2005} and in latent class discovery for gene expression data \citep{priebe2003dna}. CCCDs also show robustness to data sets with imbalanced class priors \citep{manukyan2016}. This phenomenon often occurs in real data sets; that is, some classes of the data sets have a large number of members whereas the remaining classes only have few, resulting a bias towards the majority class (the class with more members) which drastically decreases the classification performance.   

Class covers with Euclidean balls have been extended to allow the use of different type of regions to cover a class of interest. \cite{serafini2014} uses sets of boxes to find a cover of classes, and also defines the maximum redundancy problem. This is an optimization problem of covering as many points as possible by each box where the total number of boxes are kept to a (approximately) minimum. \cite{hammer2004} investigates CCP using boxes with applications to the logical data analysis. Moreover, \cite{bereg2012} extend covering boxes to rectilinear polygons to cover classes, and they report on the complexity of the CCP algorithms using such polygonal covering regions. \cite{takigawa2009} incorporate balls and establish classifiers similar to the ones based on CCCDs, and they also use sets of convex hulls. \cite{ceyhan2005pcd} uses sets of triangles relative to the tessellation of the opposite class to analytically compute the minimum number of triangles required to establish a class cover. In this work, we study class covers with particular triangular regions (simplical regions in higher dimensions).  

CCCDs can be generalized using \emph{proximity maps} \citep{jaromczyk1992}. \cite{ceyhan2005pcd} defined PCDs and introduced three families of PCDs to analytically compute the distribution of the domination number of such digraphs in a two class setting. Domination number and, another graph invariant, the arc density (the ratio of number of arcs in a  digraph to the total number of arcs possible) of these PCDs have been used for testing spatial patterns of segregation and association \citep{ceyhan2005,ceyhan2006,ceyhan2007}. In this article, we employ PCDs in statistical classification and investigate their performance. The PCDs of concern in this work are based on a particular family of proximity maps called \emph{proportional-edge} (PE) proximity maps. The corresponding PCDs are called PE-PCDs, and are defined for target class (i.e. the class of interest) points inside the convex hull of non-target points \citep{ceyhan2005pcd}. However, this construction ignores the target class points outside the convex hull of the non-target class. We mitigate this shortcoming by partitioning the region outside of the convex hull into unbounded regions, called outer simplices, which may be viewed as extensions of outer intervals in $\R$ (e.g. intervals with infinite endpoints) to higher dimensions. We attain proximity regions in these outer simplices by extending PE proximity maps to outer simplices. We establish two types of classifiers based on PE-PCDs, namely \emph{hybrid} and \emph{cover} classifiers. The first type incorporates the PE-PCD covers of only points in the convex hull and use other classifiers for points outside the convex hull of the non-target class, hence we have some kind of a hybrid classifier; the second type is further based on two class cover models where the first is a hybrid of PE-PCDs and CCCDs (composite covers) whereas the second is purely based on PE-PCDs (standard covers). 

One common property of most class covering (or set covering) methods is that none of the algorithms find the exact minimum number of covering sets in polynomial time, and solutions are mostly provided by approximation algorithms \citep{vazirani2001}. However, for PE-PCDs, the exact minimum number of covering sets (equivalent to prototype sets) can be found much faster; that is, the exact minimum solution is found in a running time polynomial in size of the data set but exponential in dimensionality. PE-PCDs have computationally tractable (exact) minimum dominating sets in $\R^d$ \citep{ceyhan2010}. Although the complexity of class covers based on this family of proximity maps exponentially increases with dimensionality, we apply dimension reduction methods (e.g. principal components analysis) to substantially reduce the number of features and to reduce the dimensionality. Hence, based on the transformed data sets in the reduced dimensions, the PE-PCD based hybrid and, in particular, cover classifiers become more appealing in terms of both prototype selection and classification performance (in the reduced dimension). We use simulated and real data sets to show that these two types of classifiers based on PE-PCDs have either comparable or slightly better classification performance than other classifiers when the data sets exhibit the class imbalance problem. 

The article is organized as follows: in Section~\ref{sec:auxiliary}, we introduce some auxiliary tools for the defining PCDs, and in particular Section~\ref{sec:pcds}, we describe the PE-PCDs. In Section~\ref{sec:pcdcovers}, we introduce two types of class cover models that are called composite and standard covers. In Section~\ref{sec:classification}, we introduce two types statistical classifiers based on PE-PCDs which are called hybrid and cover PE-PCD classifiers. The latter type is defined for both class cover models described in Section~\ref{sec:pcdcovers}. In Section~\ref{sec:simulations}, we assess the performance of PE-PCD classifiers and compare them with existing methods (such as $k$-nearest neighbors and support vector machine classifiers) on simulated data sets. Finally, in Section~\ref{sec:realdata}, we assess our classifiers on real data sets, and in Section~\ref{sec:discussion}, we present discussion and conclusions as well as future research directions. 

\section{Tessellations in $\R^d$ and the Auxiliary Tools} \label{sec:auxiliary}

In this section, we introduce tools for constructing PE-PCD classifiers. Let $(\Omega,\mathcal{M})$ be a measurable space, and let the training data set be composed of two non-empty sets, $\X_0$ and $\X_1$, that are sets of $\Omega$-valued random variables with class conditional distributions $F_0$ and $F_1$, with supports $s(F_0)$ and $s(F_1)$, and with sample sizes $n_0:=|\X_0|$ and $n_1:=|\X_1|$, respectively. We develop rules to define proximity maps and regions for the class of interest, i.e. \emph{target class}, $\X_j$, for $j=0,1$, with respect to the \emph{Delaunay tessellation} of the class of non-interest, i.e. \emph{non-target class} $\X_{1-j}$. 

A tessellation in $\R^d$ is a collection of non-intersecting (actually intersecting possibly only on boundaries) convex $d$-polytopes such that their union covers a region. We partition $\R^d$ into non-intersecting $d$-simplices and $d$-polytopes to construct PE-PCDs that tend to have multiple disconnected components. We show that such a partitioning of the domain provides digraphs with computationally tractable minimum dominating sets. In addition, we use the barycentric coordinate system to characterize the points of the target class with respect to the Delaunay tessellation of the non-target class. Such a coordinate system simplifies the definitions of many tools associated with PE-PCD classifiers in $\R^d$, including minimum dominating sets of PE-PCDs and convex distance functions. 
  
\subsection{Delaunay Tessellation of $\R^d$}

\begin{figure}[!h]
\centering
\begin{tabular}{ccc}
\includegraphics[scale=0.32]{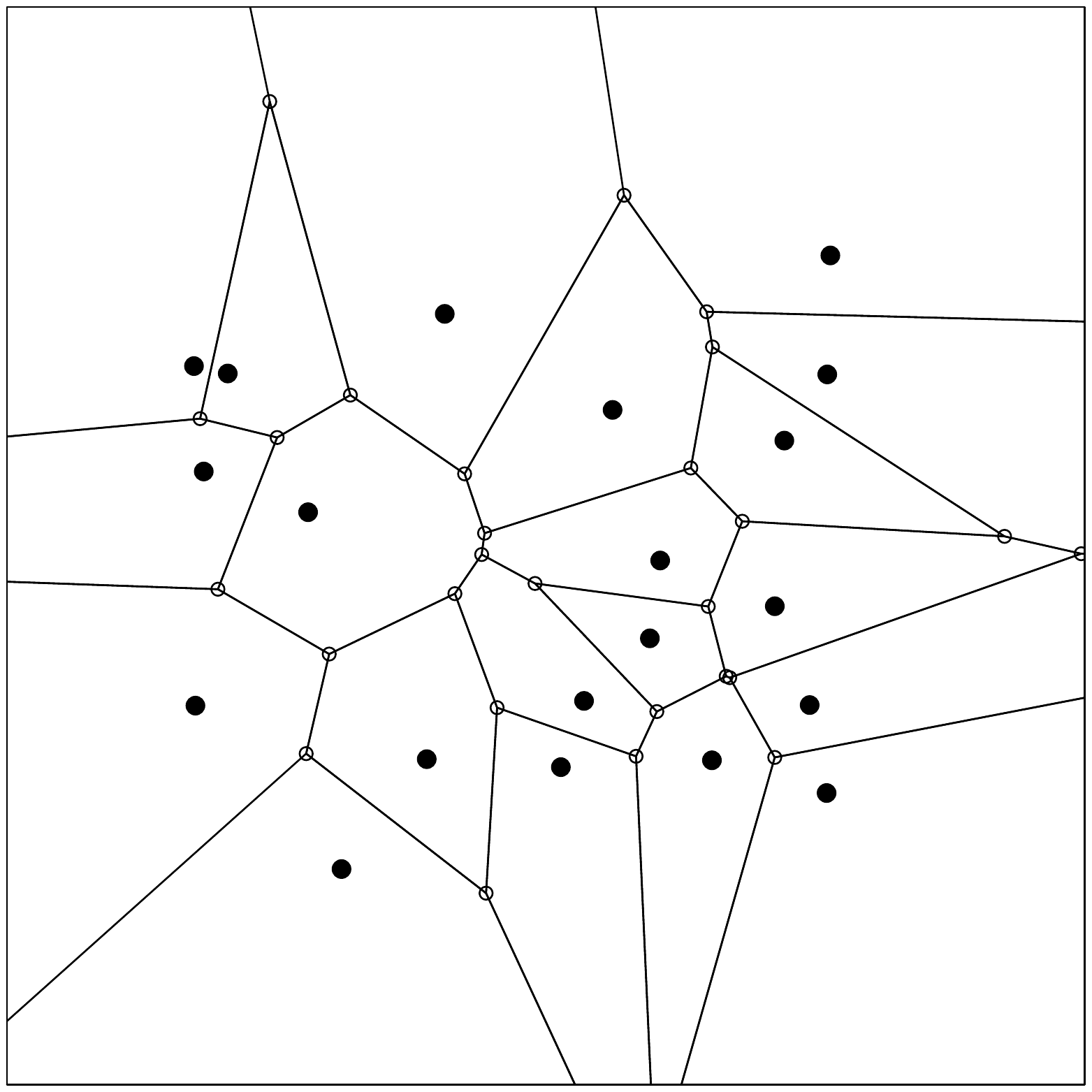} & \includegraphics[scale=0.32]{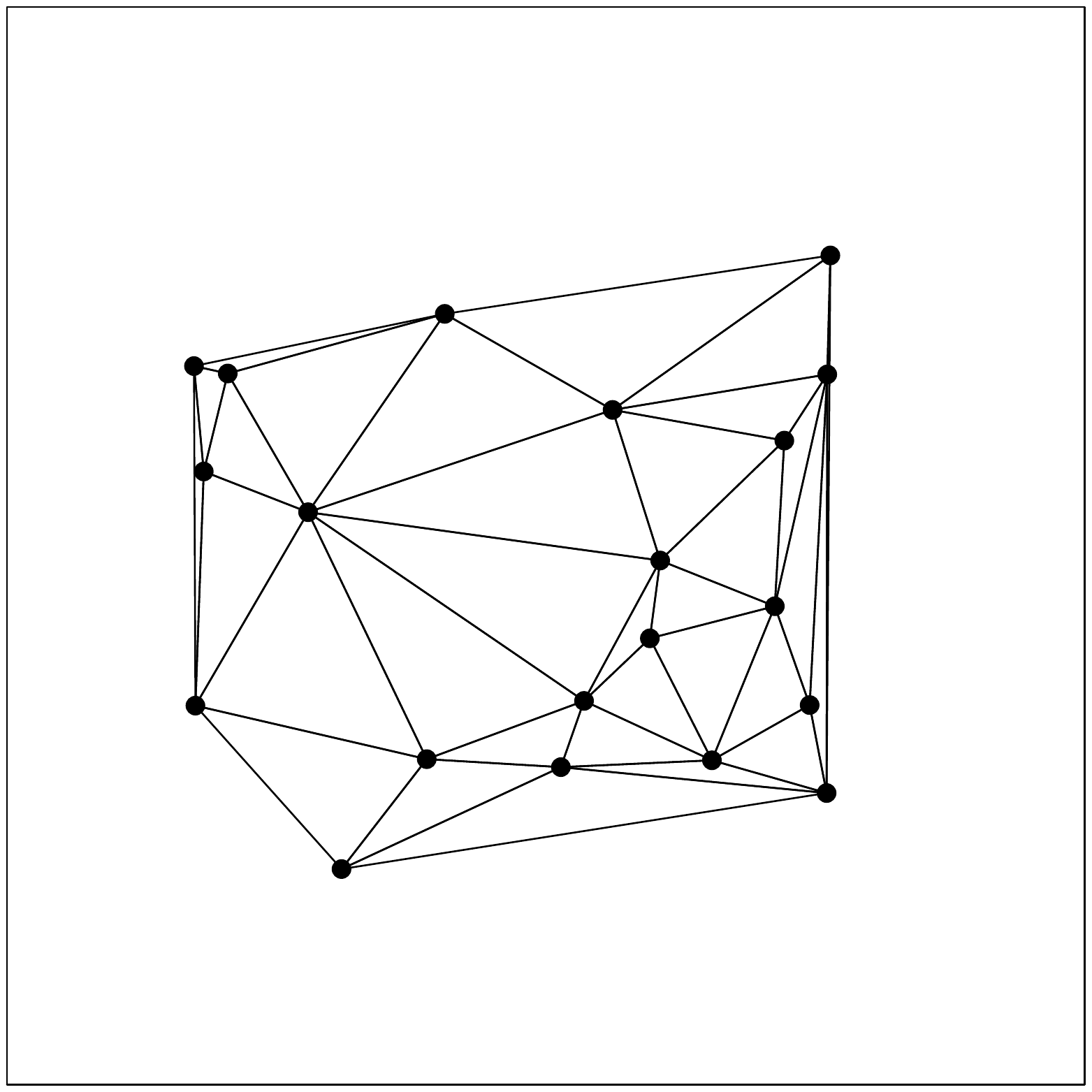} &
\includegraphics[scale=0.32]{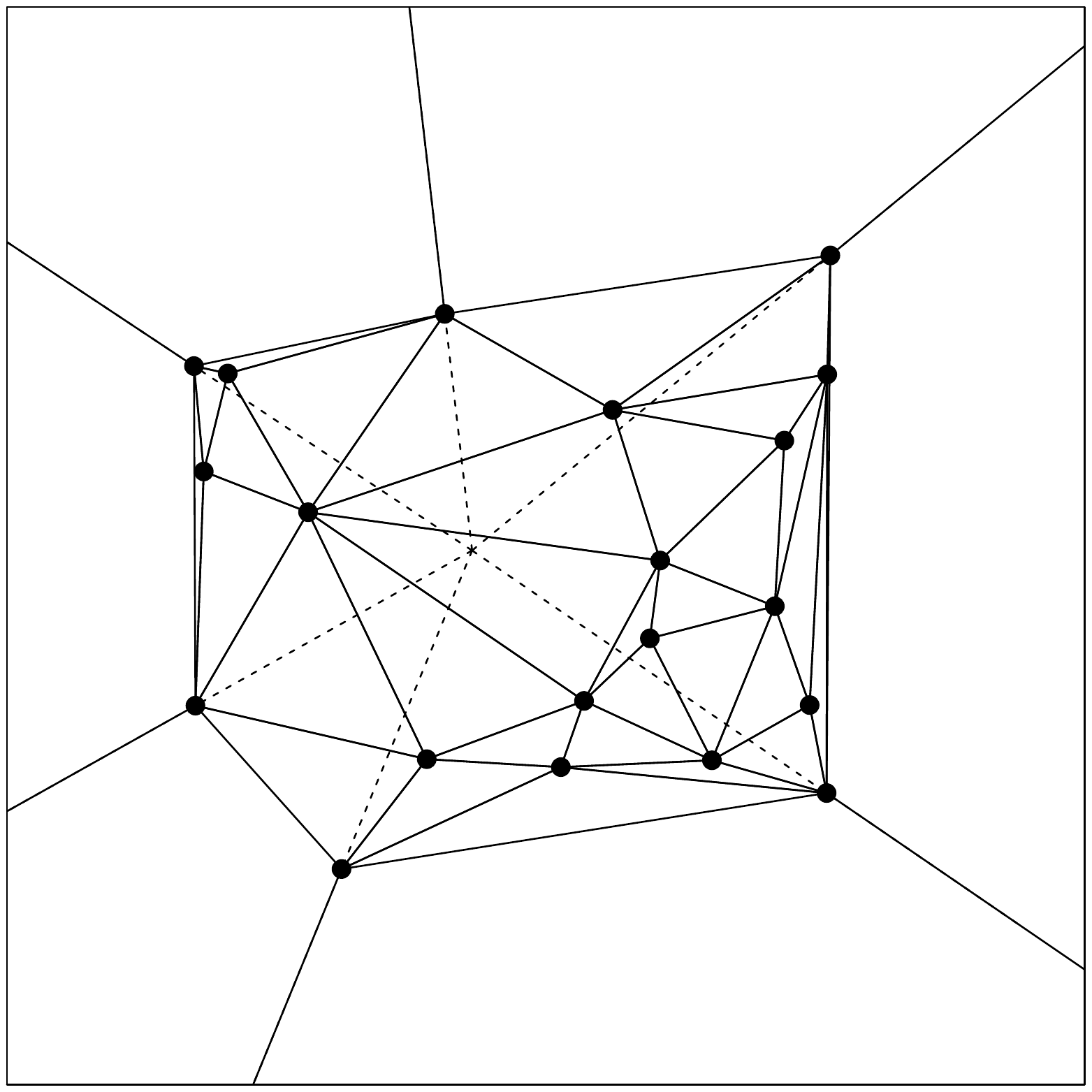} \\
(a) & (b) & (c)\\
\end{tabular}
\caption{ (a) A Voronoi diagram of points $\X_{1-j} \subset \R^2$ and (b) the associated the Delaunay triangulation, partitioning $C_H(\X_{1-j})$. (c) The Delaunay tessellation of $\X_{1-j}$ with rays $\protect\overrightarrow{C_M p_i}$ for $i=1,\ldots,6$ that yield a partitioning of $\R^2 \setminus C_H(\X_{1-j})$. The dashed lines illustrate the direction of these rays where they meet at the point $C_M$, center of mass of $C_H(\X_{1-j})$.}
\label{fig:enddel}
\end{figure}

The convex hull of the non-target class $C_H(\X_{1-j})$ can be partitioned into \emph{Delaunay cells} through the Delaunay tessellation of $\X_{1-j} \subset \R^2$. The Delaunay tessellation becomes a triangulation which partitions $C_H(\X_{1-j})$ into non intersecting triangles. For the points in the general position, the triangles in the Delaunay triangulation satisfy the property that the circumcircle of a triangle contain no points from $\X_{1-j}$ except for the vertices of the triangle. In higher dimensions, Delaunay cells are $d$-simplices (for example, a tetrahedron in $\R^3$). Hence, the $C_H(\X_{1-j})$ is the union of a set of disjoint $d$-simplices $\{\mathfrak S_k\}_{k=1}^K$ where $K$ is the number of $d$-simplices, or Delaunay cells. Each $d$-simplex has $d+1$ non-coplanar vertices where none of the remaining points of $\X_{1-j}$ are in the interior of the circumsphere of the simplex (except for the vertices of the simplex which are points from $\X_{1-j}$). Hence, simplices of the Delaunay tessellations are more likely to be acute (simplices with no substantially small inner angles). Note that Delaunay tesselation is the dual of the \emph{Voronoi diagram} of the set $\X_{1-j}$. A Voronoi diagram is a partitioning of $\R^d$ into convex polytopes such that the points inside each polytope is closer to the point associated with the polytope than any other point in $\X_{1-j}$. Hence, a polytope $\V(\y)$ associated with a point $\y \in \X_{1-j}$ is defined as $$\V(\y)=\{v \in \R^d: \lVert v-\y \rVert \leq \lVert v-z \rVert \ \text{ for all } z \in \X_{1-j} \setminus \{\y\}\}.$$ Here, $\lVert\cdot\rVert$ stands for the usual Euclidean norm. Observe that the Voronoi diagram is unique for a fixed set of points $\X_{1-j}$. A Delaunay graph is constructed by joining the pairs of points in $\X_{1-j}$  whose boundaries of voronoi polytopes are intersecting. The edges of the Delaunay graph constitute a partitioning of $C_H(\X_{1-j})$, hence the Delaunay tessellation. By the uniqueness of the Voronoi diagram, the Delaunay tesselation is also unique (except for cases where $d+1$ or more points lie on the same circle of hypersphere). An illustration of the Voronoi diagram and the corresponding Delaunay triangulation in $\R^2$ are given in Figure~\ref{fig:enddel}(a) and (b). 

A Delaunay tessellation partitions only $C_H(\X_{1-j})$ and do not offer a partitioning of the complement $\R^d \setminus C_H(\X_{1-j})$ unlike the Voronoi diagrams. As we will see in the following sections, this drawback makes the definition of our semi-parametric classifiers more difficult. Let \emph{facets} of $C_H(\X_{1-j})$ be the simplices on the boundary of $C_H(\X_{1-j})$. To partition $\R^d \setminus C_H(\X_{1-j})$, we define unbounded regions associated with each facet of $C_H(\X_{1-j})$, namely \emph{outer simplices} in $\R^d$ or \emph{outer triangles} in $\R^2$. Each outer simplex is constructed by a single facet of $C_H(\X_{1-j})$, denoted by $\F_l$ for $l=1,\cdots,L$. Here, $L$ is the number of boundary facets and, note that, each facet is a $(d-1)$-simplex. Let $\{p_1,p_2,\cdots,p_N\} \subseteq \X_{1-j}$ be the set of points on the boundary of $C_H(\X_{1-j})$, and let $C_M:=\sum_{i=1}^N p_i/N$ be the center of mass of  $C_H(\X_{1-j})$. We use the bisector rays of \cite{deng1999} as frameworks for constructing outer simplices, however such rays are not well defined for convex hulls in $\R^d$ for $d>2$. Let the ray emanating from $C_M$ through $p_i$ be denoted as $\overrightarrow{C_{M} p_i}$. Hence, we define the outer simplices by rays emanating from each boundary points $p_i$ to outside of $C_H(\X_{1-j})$ in the direction of $\overrightarrow{C_{M} p_i}$. Each facet $\F_l$ has $d$ boundary points adjacent to it, and the rays associated with these boundary points establish an unbounded region together with the facet $\F_l$. Such a region can be viewed as an infinite ``drinking glass" with $\F_l$ being the bottom while top of the glass reaching infinity, similar to intervals in $\R$ with infinite endpoints. Let $\mF_l$ denote the outer simplex associated with the facet $\F_l$. An illustration of outer triangles in $\R^2$ has been given in Figure~\ref{fig:enddel}(c) where the $C_H(\X_{1-j})$ has six facets, hence $\R^2 \setminus C_H(\X_{1-j})$ is partitioned into six disjoint unbounded regions. 

\subsection{Barycentric Coordinate System}

The \emph{barycentric coordinate system} was introduced by A.F. M\"{o}bius in his book ``The Barycentric Calculus" in 1837. The idea is to define weights $w_1$, $w_2$ and $w_3$ associated with points $\y_1$, $\y_2$ and $\y_3$ which constitute a triangle $T$ in $\R^2$, respectively \citep{ungar2010}. Hence the center of mass, or the \emph{barycenter}, for $w_1+w_2+w_3 \neq 0$ is given by
\begin{equation} \label{equ:bary_2d_P}
		P=\frac{w_1\y_1+w_2\y_2+w_3\y_3}{w_1+w_2+w_3}.
\end{equation}
\noindent Similarly, let $\mathfrak S = \mathfrak S(\Y)$ be a $d$-simplex defined by the non-coplanar points $\Y=\{\y_1,\y_2,\cdots,\y_{d+1}\} \subset \R^d$ with weights $(w_1,w_2,\cdots,w_{d+1})$. Thus, the barycenter $W \in \R^d$ is given by  
\begin{equation}
	W= \frac{\sum_{i=1}^{d+1} w_i \y_i}{\sum_{i=1}^{d+1} w_i} \quad \text{with} \quad \sum_{i=1}^{d+1} w_i \neq 0.
\end{equation}  
\noindent The $(d+1)$-tuple $\mathbf{w}=(w_1,w_2,\cdots,w_{d+1})$ can also be viewed as a set of coordinates of $W$ with respect to the set $\Y = \{\y_1,\y_2,\cdots,\y_{d+1}\}$ for $d>0$. Hence, the name \emph{barycentric coordinates}. Observe that $W$ in Equation~(\ref{equ:bary_2d_P}) is scale invariant (i.e. invariant under scaling of the weights of $W$). Therefore, the set of barycentric coordinates, also denoted as $(w_1:w_2:\ldots:w_{d+1})$, are homogeneous, i.e., for any $\lam \in \R_+$,
\begin{equation} \label{equ:homo_bary}
	(w_1:w_2:\ldots:w_{d+1}) = (\lam w_1: \lam w_2:\ldots: \lam w_{d+1}).
\end{equation}
\noindent This gives rise to \emph{special barycentric coordinates} $\mathbf{w}'=(w'_1,w'_2,\cdots,w'_{d+1})$ of a point $x \in \R^d$ with respect to the set $\Y$ as follows: 
\begin{equation}
	\sum_{i=1}^{d+1} w'_i = \sum_{i=1}^{d+1} \frac{w_i}{w_{tot}} = 1,
\end{equation} 
\noindent where $w_{tot}:=\sum_{j=1}^{d+1} w_j$. For the sake of simplicity, we refer to the special (or normalized) barycentric coordinates just as ``barycentric coordinates" throughout this work, and use $\mathbf{w}$ to denote the set of this coordinates of $x$. Hence, the vector $\mathbf{w}$ is the solution to the linear systems of equations 
\begin{equation}
\mathbf{A}\mathbf{w}=\left[\begin{array}{cccc}
\y_2-\y_1 & \y_3-\y_1 & \cdots & \y_{d+1} - \y_1 \end{array}\right]\left[\begin{array}{c}
w_{1}\\
w_{2}\\
\vdots\\
w_{d}
\end{array}\right]=x-\y_1
\end{equation}
\noindent where $\mathbf{A} \in \R^{d \times d}$ is a matrix whose columns are vectors defined by $\y_k - \y_1$ in $\R^d$ for $k=2,\cdots,d+1$. Note that $w_{d+1}=1-\sum_{i=1}^d w_i$. The set $\mathbf{w}$ is unique since vectors $\y_k - \y_1$ are linearly independent but $w_i$ are not necessarily in $(0,1)$. Barycentric coordinates define whether the point $x$ is in $\mathfrak S(\Y)$ or not, as follows: 
\begin{itemize}
\item $x \in \mathfrak S(\Y)^o$ if $w_i \in (0,1)$ for all $i=0,1,\cdots,d+1$: the point $x$ is inside of the $d$-simplex $\mathfrak S(\Y)$ where $\mathfrak S(\Y)^o$ denotes the interior of $\mathfrak S(\Y)$,  
\item $x \in \mathfrak \partial(\mathfrak S(\Y))$, the point $x$ is on the boundary of $\mathfrak S(\Y)$, if $w_i=0$ and $w_j=(0,1]$ for some $I$ such that $i \in I \subset \{0,1,\cdots,d+1\}$ and $j \in \{0,1,\cdots,d+1\} \setminus I$, 
\item $x=\y_i$ if $w_i=1$ and $w_j=0$ for any $i=0,1,\cdots,d+1$ and $j \neq i$: the point $x$ is at the a corner of $\mathfrak S(\Y)$,
\item $x \not\in \mathfrak S(\Y)$ if $w_i \not\in [0,1]$ for some $i \in \{0,1,\cdots,d+1\}$: the point $x$ is outside of $\mathfrak S(\Y)$.
\end{itemize}
\noindent Barycentric coordinates of a point $x \in \mathfrak S(\Y)$ can also be viewed as the convex combination of the points of $\Y$, the vertices on the boundary of $\mathfrak S(\Y)$. 

\subsection{Vertex Regions in  $\R^2$}

\begin{figure} [ht]
\centering
\begin{tabular}{cc}
\includegraphics[scale=0.5]{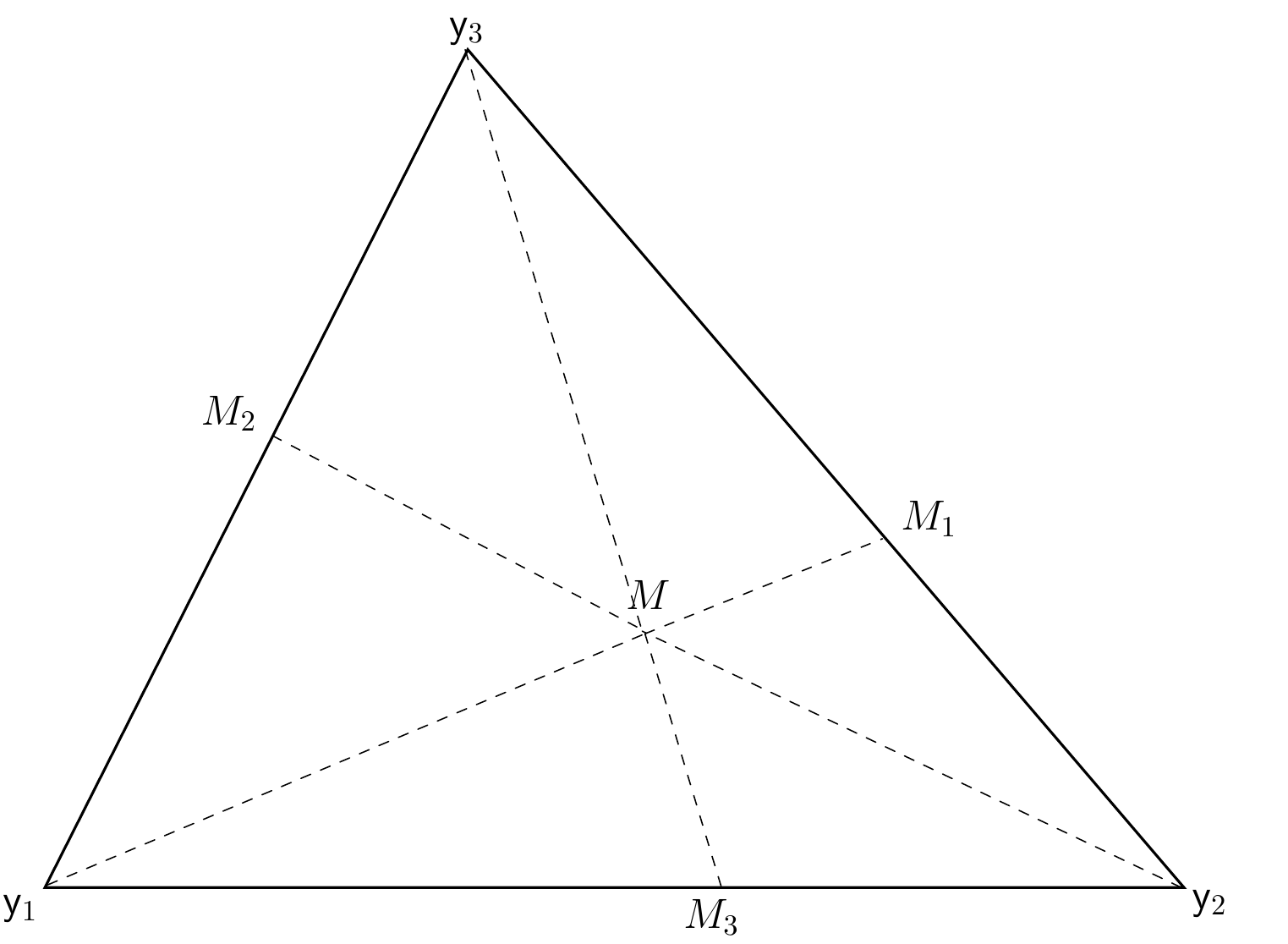} & \includegraphics[scale=0.5]{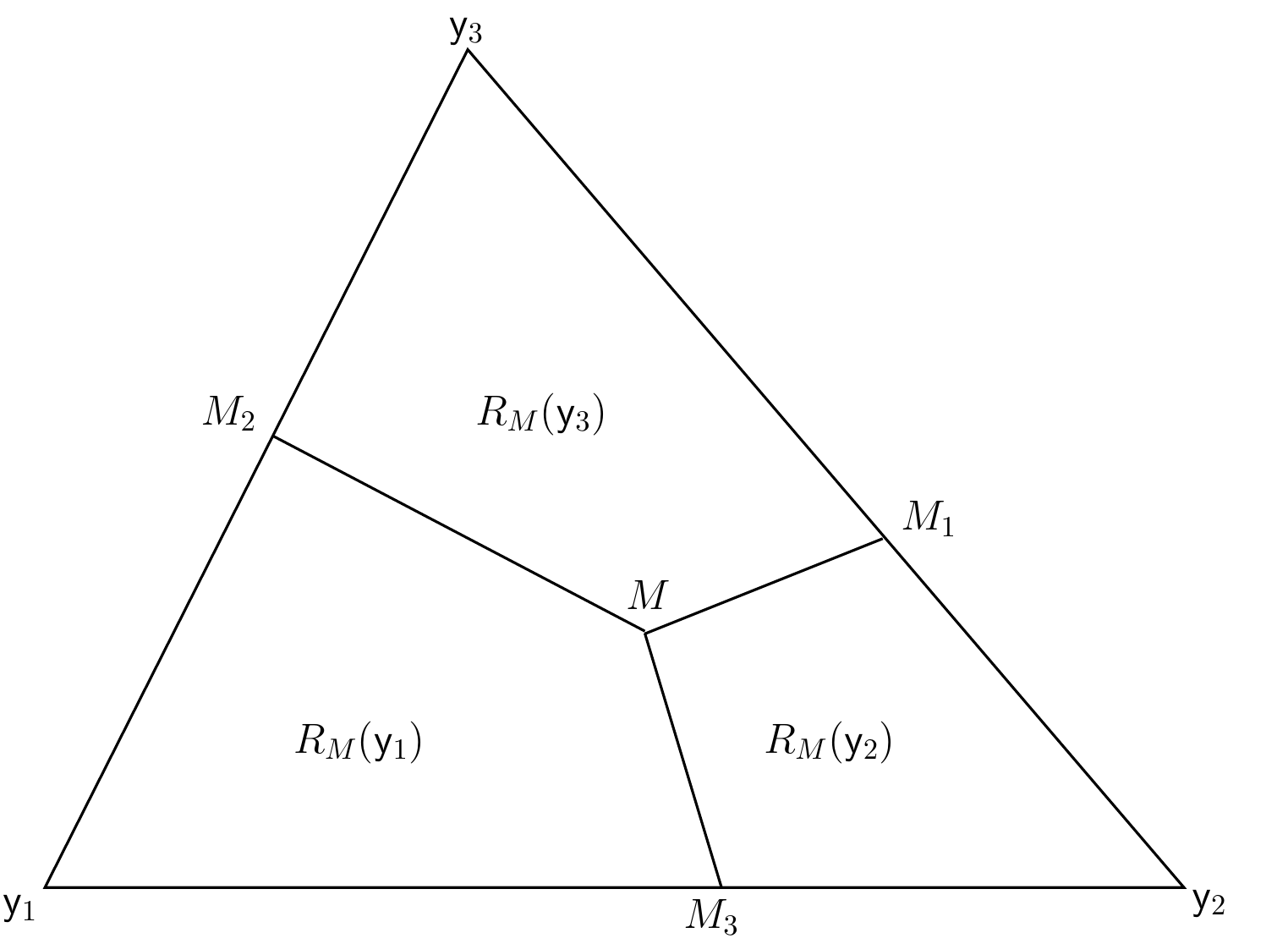} \\ 
\end{tabular}
\caption{$M$-vertex regions of an acute triangle $T(\Y)=T(\y_1,\y_2,\y_3)$ with a center $M \in T(\Y)^o$. (a) The dashed lines constitute the vertex regions. (b) Each $M$-vertex region is associated with a vertex $\y_i$ for $i=1,2,3$.}
\label{fig:vertex_reg}
\end{figure}

We first define vertex regions in $\R^2$, and later, we generalize them to vertex regions in $\R^d$ for $d>2$. Let $\Y=\{\y_1,\y_2,\y_3\} \subset \R^2$ be three non-collinear points, and let $T=T(\Y)$ be the triangle formed by these points. Also, let $e_i$ be the edge of $T$ opposite to the vertex $\y_i$ for $i=1,2,3$. We partition the triangle $T$ into regions, called \emph{vertex regions}. These regions are constructed based on a point, preferably a \emph{triangle center} $M \in T^o$. Vertex regions partition $T$ into disjoint regions (only intersecting on the boundary) such that each vertex region has only one of vertices $\{\y_1,\y_2,\y_3\}$ associated with it. In particular, $M$-vertex regions are classes of vertex regions, which are constructed by the lines from each vertex $\y_i$ to $M$. These lines cross the edge $e_i$ at point $M_i$. By connecting $M$ with each $M_i$, we attain regions associated with vertices $\{\y_1,\y_2,\y_3\}$. $M$-vertex region of $\y_i$ is denoted by $R_M(\y_i)$ for $i=1,2,3$. For the sake of simplicity, we will refer $M$-vertex regions as vertex regions. Figure~\ref{fig:vertex_reg} illustrates the vertex regions of an acute triangle in $\R^2$.

\cite{ceyhan2005} introduced the vertex regions as auxiliary tools to define proximity regions. They also gave the explicit functional forms of these regions as a function of the coordinates of vertices $\{\y_1,\y_2,\y_3\}$. However, we characterize these regions based on barycentric coordinates as given in Propositions~\ref{prop:vertex_r2_bary}, as this coordinate system will be more convenient for computation in higher dimensions.

\begin{proposition} \label{prop:vertex_r2_bary}
Let $\Y=\{\y_1,\y_2,\y_3\} \subset \R^2$ be a set of three non-collinear points, and let the set of vertex regions $\{R_M(\y_i)\}_{i=1,2,3}$ partitions $T(\Y)$. Hence for $x,M \in T(\Y)^o$, we have $x \in R_M(\y_i)$ if and only if 
\begin{align*}
 w_T^{(i)}(x) > \max_{\substack{j=1,2,3 \\ j \neq i}} \frac{m_i w_T^{(j)}(x)}{m_j}
\end{align*}
for $i=1,2,3$ where $\mathbf{w}_T(x)=\left(w_T^{(1)}(x),w_T^{(2)}(x),w_T^{(3)}(x)\right)$ and $\mathbf{m}=(m_1,m_2,m_3)$ are barycentric coordinates of $x$ and $M$ with respect to $T(\Y)$, respectively.
\end{proposition}

\noindent {\bf Proof:}
It is sufficient to show the result for $i=1$ (as others follow by symmetry). Hence we show that, for $x \in R_M(\y_1)$, we have 
\begin{align*}
w_T^{(1)}(x) > \max \Bigg\{ \frac{m_1 w_T^{(2)}(x)}{m_2},\frac{m_1 w_T^{(3)}(x)}{m_3} \Bigg\}.  
\end{align*}
\noindent Let $T_2(\Y)$ and $T_3(\Y)$ be the interiors of two triangles given by sets of points $\{\y_1,\y_2,M_{2}\}$ and $\{\y_1,\y_3,M_{3}\}$, respectively. Let $z \in T_2(\Y)$ and let $w_{T_2}(z)=(\al_1,\al_2,\al_3)$ be the barycentric coordinates of $z$ with respect to $T_2(\Y)$. Then 
\begin{align*}
	z&=\al_1 \y_1 + \al_2 \y_2 + \al_3 M_{2} \\
	 &=\al_1 \y_1 + \al_2 \y_2 + \al_3 (b \y_1 + (1-b) \y_3) \\ 
	 &=(\al_1+\al_3 b) \y_1 + \al_2 \y_2 + \al_3 (1-b) \y_3, \
\end{align*}
since $M_2$ lies on edge $e_2$, we can write it as $M_2=b \y_1 + (1-b) \y_3$ for some $b \in (0,1)$. By the uniqueness of $\mathbf{w}_T(z)$, we have $w_T^{(1)}(z) = \al_1+\al_3 b$ and $w_T^{(3)}(z) = \al_3 (1-b)$. Hence, 
\begin{align*}
	\frac{w_T^{(1)}(z)}{w_T^{(3)}(z)} = \frac{\al_1+\al_3 b}{\al_3 (1-b)} > \frac{b}{(1-b)} = \frac{m_1}{m_3}
\end{align*}
since $\al_i > 0$ for $i=1,2,3$. Also, since $M_{2}$ and $M$ are on the same line which crosses the edge $e_2$, for some $c \in (0,1)$:
\begin{align*}
	M&=c \y_2 + (1-c) M_{2} \\
	 &=c \y_2 + (1-c) (b y_1 + (1-b) y_3) \\
	 &=b(1-c) \y_1 + c \y_2 + (1-b)(1-c) \y_3, \
\end{align*}
Hence, $b(1-c) = m_1$ and $(1-b)(1-c) = m_3$, and observe that $m_1/m_3=b/(1-b)$. Then, $T_2(\Y)=\{x \in T(\Y)^o: w_T^{(1)}(x) > (m_1/m_3) w_T^{(3)}(x)\}$, and similarly, $T_3(\Y)=\{x \in T(\Y)^o: w_T^{(1)}(x) > (m_1/m_2) w_T^{(2)}(x)\}$. Thus, 
\begin{align*}
R_M(\y_1) = T_2(\Y) \cap T_3(\Y) = \Bigg\{x \in T(\Y)^o: w_T^{(1)}(x) > \max \Bigg\{ \frac{m_1 w_T^{(2)}(x)}{m_2},\frac{m_1 w_T^{(3)}(x)}{m_3} \Bigg\} \Bigg\}. \quad \blacksquare
\end{align*}

	Note that, when $M:=M_C$ the median (or the center of mass) of the triangle $T(\Y)$, we can simplify the result of Proposition~\ref{prop:vertex_r2_bary}; that is, for any point $x \in T(\Y)^o$, we have $x \in R_{M_C}(\y_i)$ if and only if $w_T^{(i)}(x)=\max_{j=1,2,3} w_T^{(j)}(x)$ since the set of (special) barycentric coordinates of $M_C$ is $\mathbf{m}_C=(1/3,1/3,1/3)$. 
	
\subsection{$M$-Vertex Regions in $\R^d$ with $d>2$}

The definitions of vertex regions in $\R^2$ can be extended to the ones in $\R^d$ for $d>2$. A $d$-simplex is the smallest convex polytope in $\R^d$ constructed by a set of non-coplanar vertices $\Y=\{\y_1,\y_2,\cdots,\y_{d+1}\}$. The boundary of a $d$-simplex consists of $k$-simplices called $k$-faces for $0 \leq k < d$. Each $k$-face is a simplex defined by a subset of $\Y$ with $k$ elements, hence there are $\binom{d+1}{k+1}$ $k$-faces in a $d$-simplex. Let $\mathfrak S(\Y)$ be the simplex defined by the set of points $\Y$. Given a \emph{simplex center} $M \in \mathfrak S(\Y)^o$ (e.g. a triangle center in $\R^2$), there are $d+1$ $M$-vertex regions constructed by the set $\Y$. The $M$-vertex region of the vertex $\y_i$ is denoted by $R_M(\y_i)$ for $i=1,2,\cdots,d+1$. 

\begin{figure} [ht]
\centering
\begin{tabular}{cc}
\includegraphics[scale=0.45]{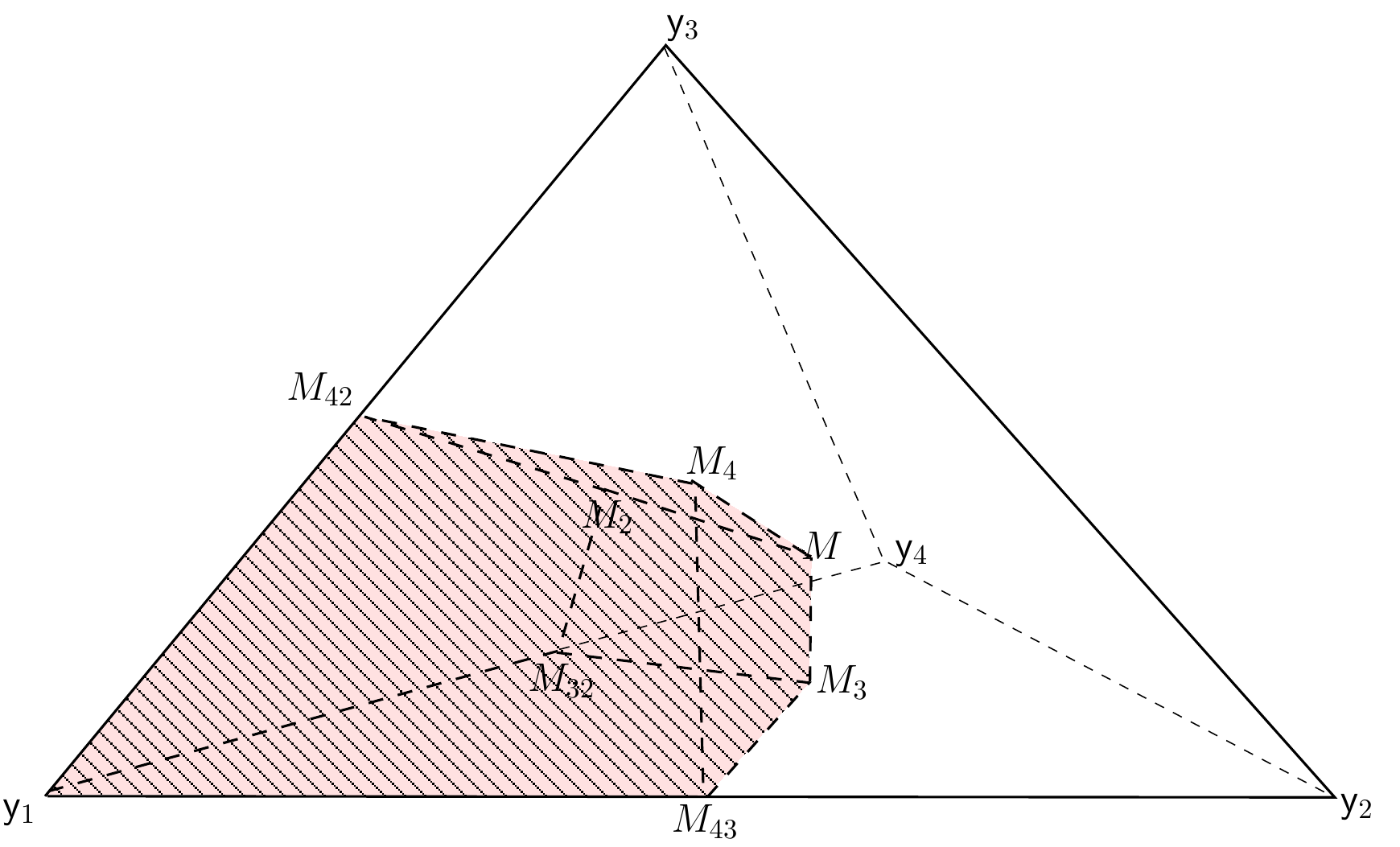} & 
\includegraphics[scale=0.45]{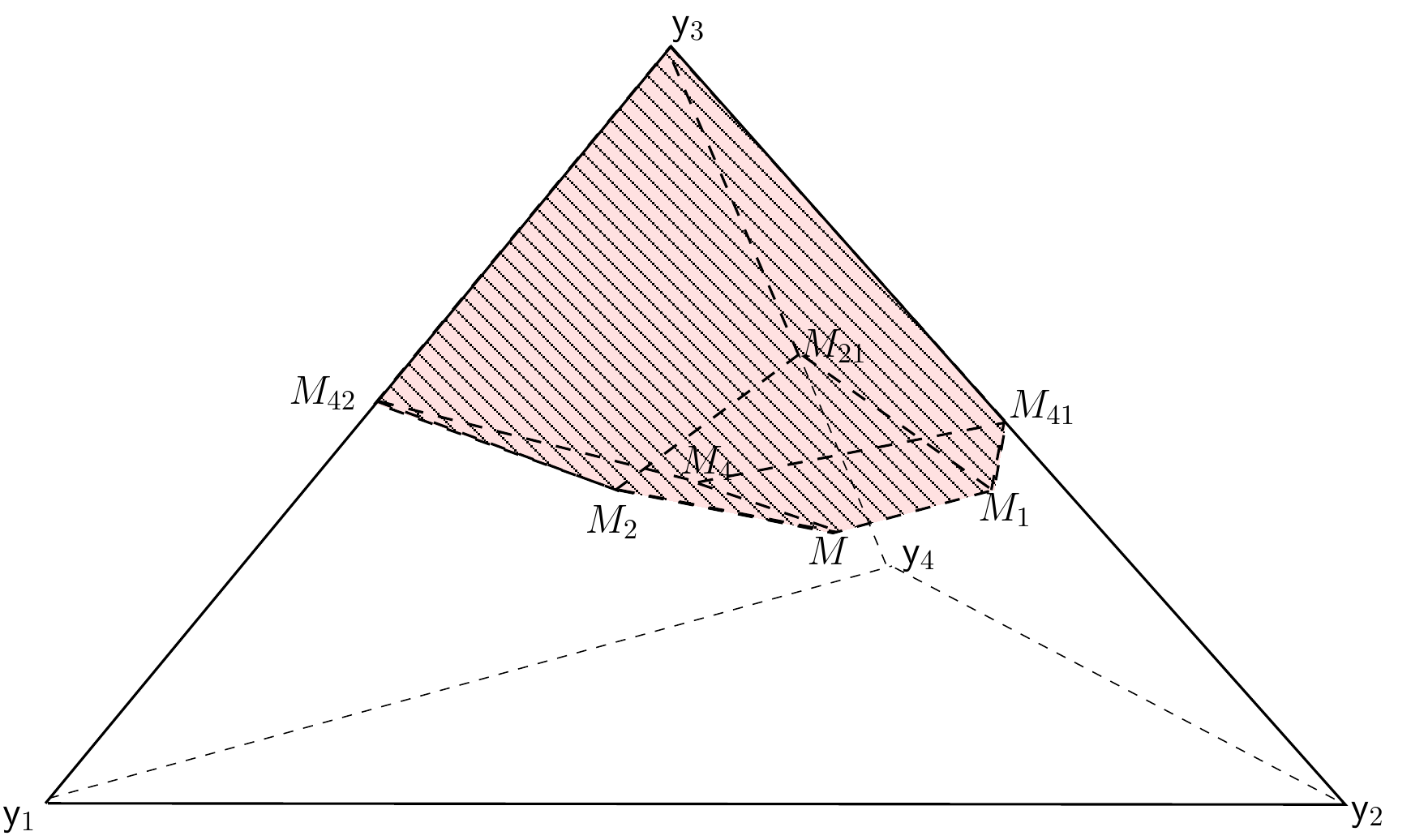} \\ 
(a) & (b) \\
\end{tabular}
\caption{(a) $M$-vertex region $R_M(\y_1)$ of vertex $\y_1$ and (b) $R_M(\y_3)$ of vertex $\y_3$ of a 3-simplex, or a tetrahedron. $M$-vertex regions are shaded. }
\label{fig:vertex_reg_nd}
\end{figure}

For $i=1,\ldots,d+1$, let $f_i$ denote the $(d-1)$-face opposite to the vertex $\y_i$. Observe that the lines through the points $\y_i$ and $M$ cross the face $f_i$, a ($d-1$)-face, at the points $M_i$. Similarly, since the face $f_i$ is a ($d-1$)-simplex with a center $M_i$ for any $i=1,\ldots,d+1$, we can find the centers of $(d-2)$-faces of this ($d-1$)-simplex. Note that both $M_i$ and $M$ are of same type of centers of their respective simplices $f_i$ and $\mathfrak S(\Y)$. The vertex region $R_M(\y_i)$ is the convex hull of the points $\y_i$, $\{M_j\}^{d+1}_{j=1;j \neq i}$, and centers of all $k$-faces (which are also $k$-simplices) adjacent to $\y_i$ for $k=1,\ldots,d-2$. Illustration of the vertex regions $R_M(\y_1)$ and $R_M(\y_3)$ of a 3-simplex (tetrahedron) are given in Figure~\ref{fig:vertex_reg_nd}. Each 2-face of this 3-simplex is a 2-simplex (a triangle). For example, in Figure~\ref{fig:vertex_reg_nd}(a), the points $M_2$, $M_3$ and $M_4$ are centers of $f_2$, $f_3$ and $f_4$, respectively. Moreover, these 2-simplices also have faces (1-faces or edges of the $3$-simplex), and the centers of these faces are $\{M_{ij}\}^4_{i,j=1;i \neq j}$. Hence, the vertex region $R_M(\y_1)$ is a convex polytope of points $\{\y_1,M,M_2,M_3,M_4,M_{32},M_{42},M_{43}\}$ and $R_M(\y_3)$ is a convex polytope of points $\{\y_3,M,M_2,M_4,M_1,M_{42},M_{41},M_{21}\}$. The following theorem is an extension of the Proposition~\ref{prop:vertex_r2_bary} to higher dimensions. 

\begin{theorem} \label{thm:vertex_rd_bary}
Let $\Y=\{\y_1,\y_2,\cdots, \y_{d+1}\} \subset \R^d$ be a set of non-coplanar points for $d>0$, and let the set of vertex regions $\{R_M(\y_i)\}_{i=1}^{d+1}$ partitions $\mathfrak S(\Y)$. Hence, for $x,M \in \mathfrak S(\Y)^o$, we have $x \in R_M(\y_i)$ if and only if 
\begin{equation}
w_{\mathfrak S}^{(i)}(x) > \max_{\substack{j=1,\cdots,d+1 \\ j \neq i}} \frac{m_i w_{\mathfrak S}^{(j)}(x)}{m_j}
\end{equation}
\noindent where $\mathbf{w}_{\mathfrak S}(x)=\left(w_{\mathfrak S}^{(1)}(x),\cdots,w_{\mathfrak S}^{(d+1)}(x)\right)$ and $\mathbf{m}=(m_1,\ldots,m_{d+1})$ are the barycentric coordinates of $x$ and $M$ with respect to $\mathfrak S(\Y)$, respectively. 
\end{theorem}

\noindent See Appendix for the proof. 
	
	For $M=M_C$ and for any point $x \in \mathfrak S(\Y)^o$, we have $x \in R_{M_C}(\y_i)$ if and only if $w_T^{(i)}(x)=\max_{j} w_T^{(j)}(x)$ since the set of barycentric coordinates of $M_C$ is $\mathbf{m}_C=(1/(d+1),1/(d+1), \ldots, 1/(d+1))$. The $M_C$-vertex regions are particularly appealing for our proportional-edge proximity regions. 

\section{Proximity Regions and Proximity Catch Digraphs} \label{sec:pcds}

We consider proximity regions for the (supervised) two-class classification problem, then perform complexity reduction via minimum dominating sets of the associated proximity catch digraphs. For $j=0,1$, the {\em proximity map} $\N(\cdot): \Omega \rightarrow 2^{\Omega}$ associates with each point $x \in \X_j$, a {\em proximity region} $\N(x) \subset \Omega$. Consider the data-random (or vertex-random) proximity catch digraph $D_j=(\V_j,\A_j)$ with vertex set $\V_j=\X_j$ and arc set $\A_j$ defined by $(u,v) \in \A_j \iff $ $\{u,v\}\subset \X_j$ and $v \in \N(u)$, for $j=0,1$. The digraph $D_j$ depends on the (joint) distribution of the sets of  points $\X_0$ and $\X_1$, and on the map $\N(\cdot)$. The adjective {\em proximity} --- for the digraph $D_j$ and for the map $\N(\cdot)$ --- comes from thinking of the region $\N(x)$ as representing those points in $\Omega$ ``close'' to $x$ \citep{toussaint1980,jaromczyk1992}. Our proximity catch digraphs (PCDs) for $\X_j$ against $\X_{1-j}$ are defined by specifying $\X_j$ as the target class and $\X_{1-j}$ as the non-target class. Hence, in the definitions of our PCDs, the only difference is switching the roles of $\X_0$ and $\X_1$. For $j=0$, $\X_0$ becomes the target class, and for $j=1$, $\X_1$ becomes the target class.

The proximity regions associated with PCDs introduced by \cite{ceyhan2005} are \emph{simplicial} proximity regions (regions that constitute simplices in $\R^d$) defined for the points of the target class $\X_j$ in the convex hull of the non-target class, $C_H(\X_{1-j})$. However, by introducing the \emph{outer simplices} associated with the facets of $C_H(\X_{1-j})$, we extend the definition of the simplical proximity regions to $\R^d \setminus C_H(\X_{1-j})$. Such simplical regions are $d$-simplices in $C_H(\X_{1-j})$ (triangles in $\R^2$ and tetrahedrons in $\R^3$) and $d$-polytopes for $\R^d \setminus C_H(\X_{1-j})$. After partitioning $\R^d$ into disjoint regions, we further partition each simplex $\mathfrak S_k$ (only the ones inside $C_H(\X_{1-j})$) into vertex regions, and define the simplical proximity regions $\N(x)$ for $x \in \mathfrak S_k$. Here, we define the regions $\N(x)$ as open sets in $\R^d$.
 
\subsection{Class Cover Catch Digraphs} \label{sec:cccds}

Class Cover Catch Digraphs (CCCDs) are graph theoretic representations of the CCP \citep{priebe2001,priebe:2003b}. In a CCCD, for $x,y \in \X_j$; let $x$ be the center of a ball $B=B(x,\varepsilon)$ with radius $\varepsilon=\varepsilon(x)$. A CCCD is a digraph $D_j=(\V_j,\A_j)$ with vertex set $\V_j=\mathcal{X}_j$ and the arc set $\A_j$ where $(x,y) \in \A_j$ iff $y \in B$. One particular family of CCCDs are called pure-CCCDs wherein, for all $x \in \X_j$, no non-target class point lies in $B$. Hence, for some $\theta \in (0,1]$ and for all $x \in \X_j$, the open ball $B$ is denoted by $B_{\theta}(x,\varepsilon_{\theta}(x))$ with the radius $\varepsilon_{\theta}(x)$ given by
\begin{equation} \label{equtheta}
	\varepsilon_{\theta}(x):=(1-\theta)d(x,l(x)) + \theta d(x,u(x)),
\end{equation}
where
\begin{equation*}
	u(x):=\underset{y \in \X_{1-j}}{\argmin} \thinspace d(x,y)
\end{equation*}
and
\begin{equation*}
	l(x):=\underset{z \in \X_j}{\argmax} \{d(x,z): d(x,z) < d(x,u(x))\}.
\end{equation*}

\begin{figure}[h]
\centering
\begin{tabular}{cc}
\includegraphics[scale=0.4]{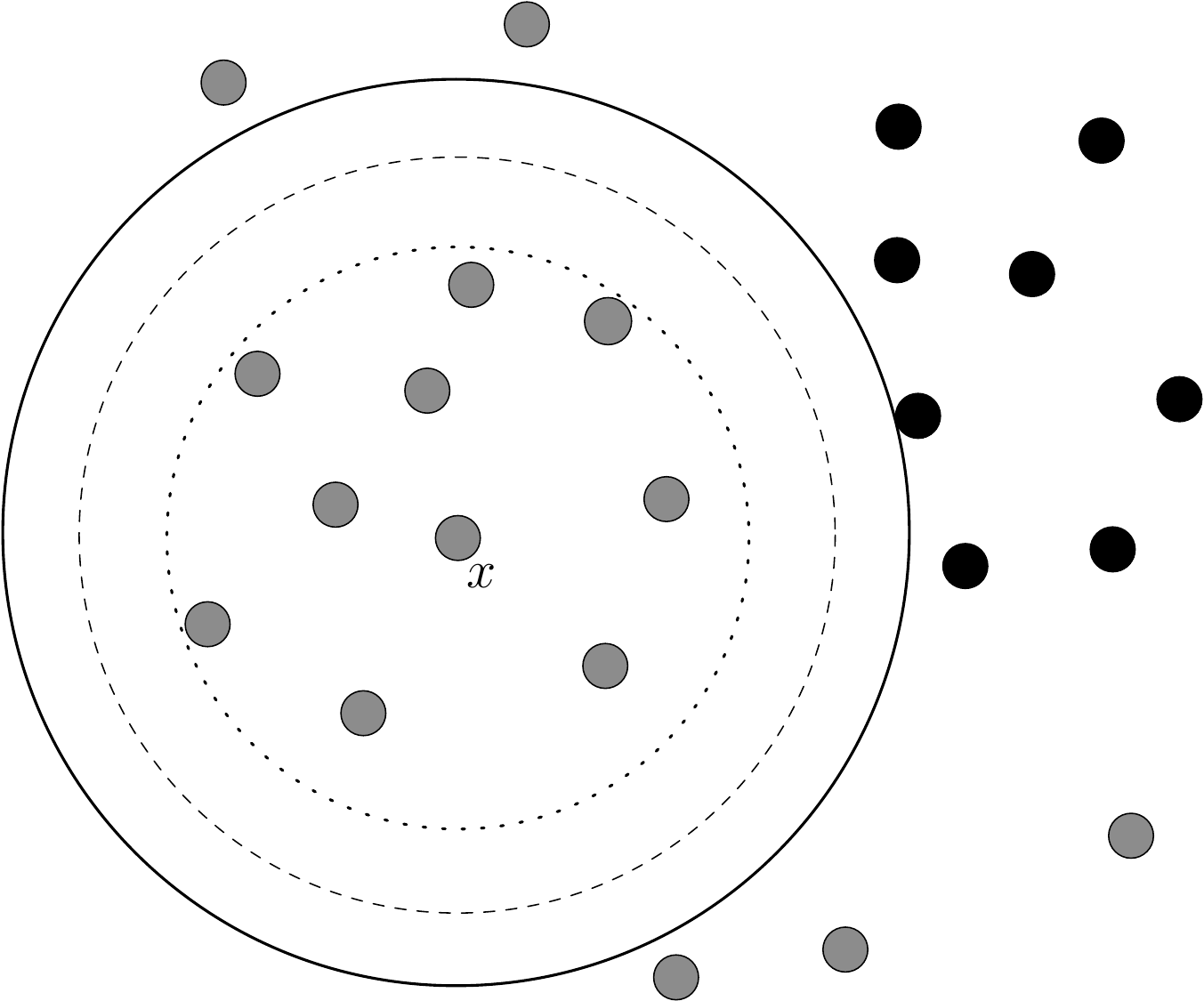}
\end{tabular}
\caption{The radius $\varepsilon_{\theta}(x)$ of a single target class point $x$ in a two-class setting. Grey and black points represent the points of the target class $\X_j$ and the non-target class $\X_{1-j}$, respectively. The solid circle is constructed with the radius $\varepsilon_{\theta}(x)$ given by $\theta=1$, dashed one by $\theta=0.5$ and the dotted one by $\theta=\epsilon$, where $\epsilon$ is the machine epsilon.}
\label{theta}
\end{figure}

\noindent Here, $d(.,.)$ can be any dissimilarity measure but we use the Euclidean distance henceforth. For all $x \in \X_j$, the definition of the radius $\varepsilon_{\theta}(x)$ keeps any non-target class point $v \in \X_{1-j}$ out of the ball $B$; that is, $\X_{1-j} \cap B = \emptyset$. We say the CCCD $D_j$ is ``pure" since the balls include only the target class points and none of the non-target class points. The CCCD $D_j$ is invariant to the choice of $\theta$, but this parameter affects the classification performance. This parameter potentially establishes classifiers with increased performance \citep{priebe:2003b}. An illustration of the effect of parameter $\theta$ on the radius of $B_{\theta}(x,\varepsilon_{\theta}(x))$ is given in Figure~\ref{theta} \citep{devinney2003}. In fact, CCCDs can also be viewed as a family of PCDs using \emph{spherical} proximity maps, letting $\N(x):=B(x,\varepsilon(x))$. We denote the proximity regions associated with pure-CCCDs as $\N_S(x,\theta)=B_{\theta}(x,\varepsilon_{\theta}(x))$. For simplicity, we refer to pure-CCCDs as CCCDs throughout this article.

\subsection{Proportional-Edge Proximity Maps} \label{sec:PCDs}

We use a type of proximity map with expansion parameter $r$, namely \emph{proportional-edge} (PE) proximity map, denoted by $\N_{PE}(\cdot,r)$. The PE proximity map and the associated digraphs, PE-PCDs, are defined in \cite{ceyhan2005}. Currently, PE-PCDs are only defined for the points in $\X_j \cap C_H(\X_{1-j})$. Hence, for the remaining points of the target class $\X_j$, i.e. $\X_j \setminus C_H(\X_{1-j})$, we extend the definition of PE proximity maps to the outer simplices. Hence, we will be able to show later that the resulting PCDs have computationally tractable \emph{minimum dominating sets} which are equivalent to the exact minimum prototype sets of PE-PCD classifiers for the entire data set. 

\subsubsection{Proximity Maps of $d$-Simplices}

\begin{figure} [ht]
\centering
\includegraphics[scale=0.33]{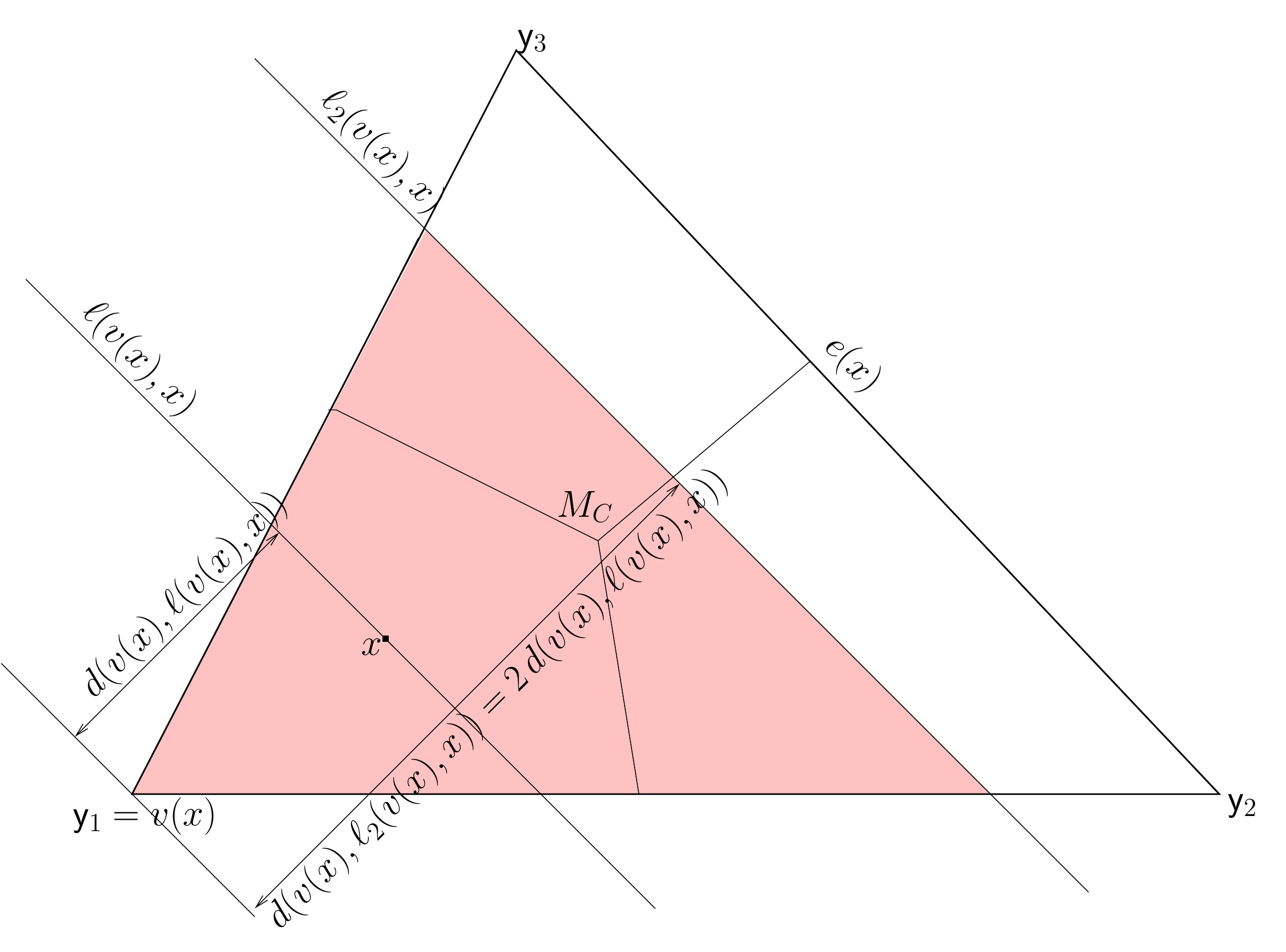}
\caption{The PE proximity region (shaded), $\N_{PE}(x,r=2)$, in a triangle $T \subseteq \R^2$.}
\label{fig:ProxMap}
\end{figure}

For $r \in [1,\infty)$, we define $\N_{PE}(\cdot,r)$ to be the PE proximity map associated with a triangle $T=T(\Y)$ formed by the set of non-collinear points $\Y = \{\y_1,\y_2,\y_3\} \subset \mathbb{R}^2$. Let $R_{M_C}(\y_1)$, $R_{M_C}(\y_2)$ and $R_{M_C}(\y_3)$ be the vertex regions associated with vertices $\y_1$,$\y_2$ and $\y_3$. Note that the barycentric coordinates of $M_C$ are $(1/3:1/3:1/3)$. For $x \in T^o$, let $v(x) \in \Y$ be the vertex whose region contains $x$; hence $x \in R_{M_C}(v(x))$. If $x$ falls on the boundary of two vertex regions, or on $M_C$, we assign $v(x)$ arbitrarily. Let $e(x)$ be the edge of $T$ opposite to $v(x)$. Let $\ell(v(x),x)$ be the line parallel to $e(x)$ through $x$. Let $d(v(x),\ell(v(x),x))$ be the Euclidean (perpendicular) distance from $v(x)$ to $\ell(v(x),x)$. For $r \in [1,\infty)$, let $\ell_r(v(x),x)$ be the line parallel to $e(x)$ such that $d(v(x),\ell_r(v(x),x)) = rd(v(x),\ell(v(x),x))$. Let $T_r(x)$ be the triangle similar to and with the same orientation as $T$ where $T_r(x)$ has $v(x)$ as a vertex and $\ell_r(v(x),x)$ as edge opposite of $v(x)$. Then the {\em proportional-edge} proximity region $\N_{PE}(x,r)$ is defined to be $T_r(x) \cap T$. Figure~\ref{fig:ProxMap} illustrates a PE proximity region $\N_{PE}(x,r)$ of a point $x$ in an acute triangle. 

The extension of $\N_{PE}(\cdot,r)$ to $\mathbb{R}^d$ for $d > 2$ is straightforward. Now, let $\Y = \{\y_1,\y_2,\cdots,\y_{d+1}\}$ be a set of $d+1$ non-coplanar points, and represent the simplex formed by the these points as $\mathfrak S=\mathfrak S(\Y)$. We define the PE proximity map as follows. Given a point $x \in \mathfrak S^o$, let $v(x)$ be the vertex in whose region $x$ falls (if $x$ falls on the boundary of two vertex regions or on $M_C$, we assign $v(x)$ arbitrarily.) Let $\varphi(x)$ be the face opposite to vertex $v(x)$, and $\eta(v(x),x)$ be the hyperplane parallel to $\varphi(x)$ which contains $x$. Let $d(v(x),\eta(v(x),x))$ be the (perpendicular) Euclidean distance from $v(x)$ to $\eta(v(x),x)$. For $r \in [1,\infty)$, let $\eta_r(v(x),x)$ be the hyperplane parallel to $\varphi(x)$ such that $d(v(x),\eta_r(v(x),x))=r\,d(v(x),\eta(v(x),x))$. Let $\mathfrak S_r(x)$ be the polytope similar to and with the same orientation as $\mathfrak S$ having $v(x)$ as a vertex and $\eta_r(v(x),x)$ as the opposite face. Then the proportional-edge proximity region is given by $\N_{PE}(x,r):=\mathfrak S_r(x) \cap \mathfrak S$. 

Notice that, so far, we assumed a single $d$-simplex for simplicity. For $n_{1-j}=d+1$, the convex hull of the non-target class $C_H(\X_{1-j})$ is a $d$-simplex. If $n_{1-j}>d+1$, then we consider the Delaunay tessellation (assumed to exist) of $\X_{1-j}$ where $\mS^{(1)}_{1-j} =\{\mathfrak S_1,\ldots,\mathfrak S_K\}$ denotes the set of all Delaunay cells (which are $d$-simplices). We construct the proximity region $\N_{PE}(x,r)$ of a point $x \in \X_j$ depending on which $d$-simplex $\mathfrak S_k$ this point reside in. Observe that, this construction pertains to points in $\X_j \cap C_H(\X_{1-j})$ only.

\subsubsection{Proximity Maps of outer simplices}

\begin{figure} [ht]
\centering
\includegraphics[scale=0.43]{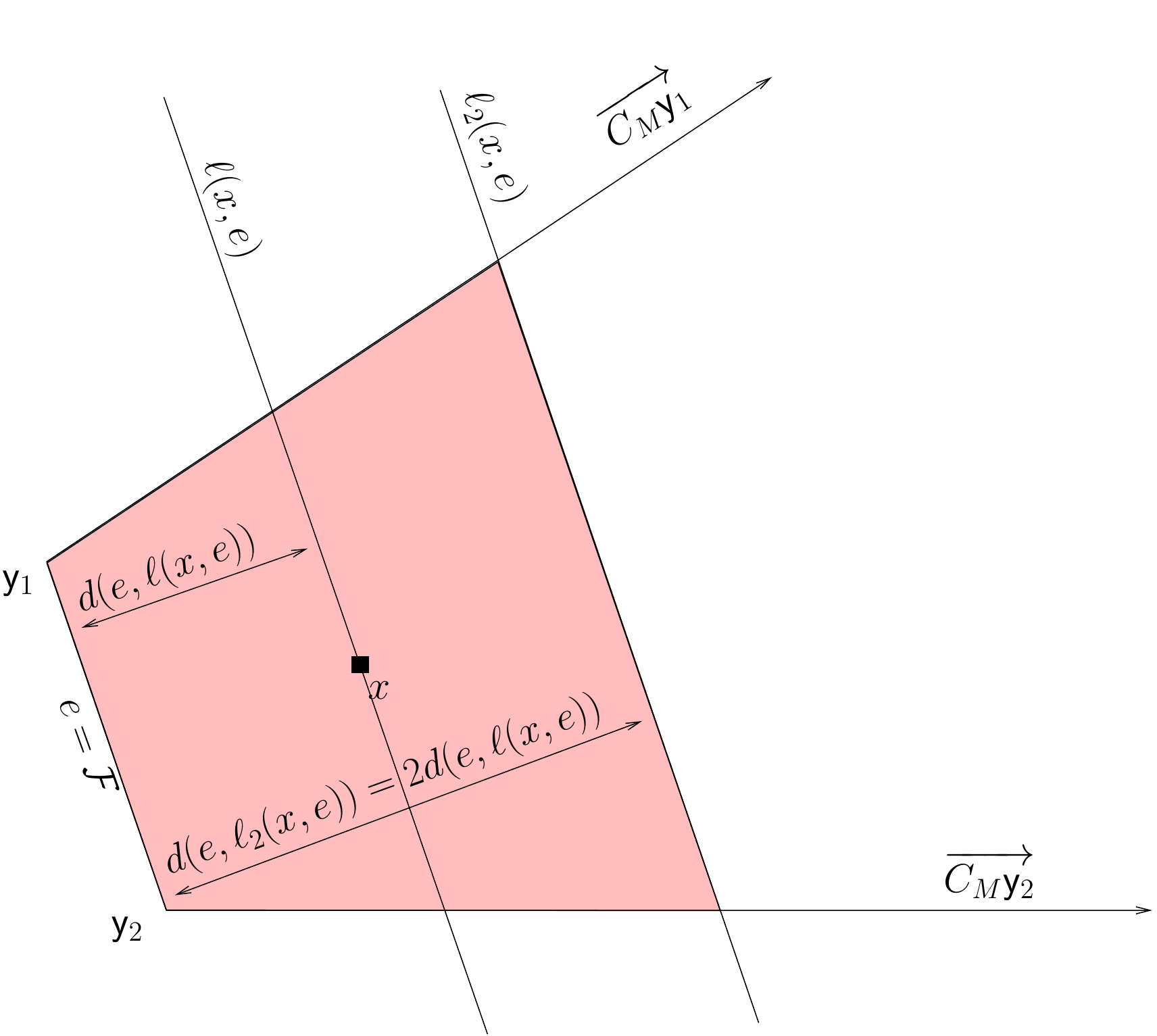} 
\caption{The proportional-edge proximity region, $\N_{PE}(x,r=2)$ (shaded), in an outer triangle $\mF \subseteq \R^2$.}
\label{fig:ProxMap_out}
\end{figure}

For points of the target class $\X_j$ outside of the convex hull of the non-target class $\X_{1-j}$, i.e. $\X_j \setminus C_H(\X_{1-j})$, we define the PE proximity maps similar to the ones defined for $d$-simplices. Let $\mF \subset \R^2$ be an outer triangle defined by the adjacent boundary points $\{\y_1,\y_2\} \subset \R^2$ of $C_H(\X_{1-j})$ and by rays $\overrightarrow{C_{M} \y_1}$ and $\overrightarrow{C_{M} \y_2}$ for $C_M$ being the median of the boundary points of $C_H(\X_{1-j})$. Also, let $e=\F$ be the edge (or facet) of $C_H(\X_{1-j})$ adjacent to vertices $\{\y_1,\y_2\}$. Note that there is no center in an outer triangle, and hence no vertex regions. For $r \in [1,\infty)$, we define $\N_{PE}(\cdot,r)$ to be the PE proximity map of the outer triangle. For $x \in \mF^o$, let $\ell(x,e)$ be the line parallel to $e$ through $x$, and let $d(e,\ell(x,e))$ be the Euclidean distance from $e$ to $\ell(x,e)$. For $r \in [1,\infty)$, let $\ell_r(x,e)$ be the line parallel to $e$ such that $d(e,\ell_r(x,e)) = rd(e,\ell(x,e))$. Let $\mF_r(x)$ be a polygon similar to the outer triangle $\mF$ such that $\mF_r(x)$ has $e$ and $e_r(x)=\ell_r(x,e) \cap \mF$ as its two edges, however $\mF_r(x)$ is a bounded region whereas $\mF$ is not. Then, the proximity region $\N_{PE}(x,r)$ is defined to be $\mF_r(x)$. Figure~\ref{fig:ProxMap_out} illustrates a PE proximity region $\N_{PE}(x,r)$ of a point $x$ in an outer triangle. 

The extension of $\N_{PE}(\cdot,r)$ of outer triangles to $\mathbb{R}^d$ for $d > 2$ is straightforward. Let $\mF \subset \R^d$ be an outer simplex defined by the adjacent boundary points $\{\y_1,\ldots,\y_d\} \subset \R^d$ of $C_H(\X_{1-j})$ and by rays $\{\overrightarrow{C_{M} \y_1},\ldots,\overrightarrow{C_{M} \y_d}\}$. Also, let $\F$ be the facet of $C_H(\X_{1-j})$ adjacent to vertices $\{\y_1,\ldots,\y_d\}$. We define the PE proximity map as follows. Given a point $x \in \mathfrak \mF^o$, let $\eta(x,\F)$ be the hyperplane parallel to $\F$ through $x$ and let $d(\F,\eta(x,\F))$ be the Euclidean distance from $\F$ to $\eta(x,\F)$. For $r \in [1,\infty)$, let $\eta_r(x,\F)$ be the hyperplane parallel to $\F$ such that $d(\F,\eta_r(x,\F)) = rd(\F,\eta(x,\F))$. Let $\mF_r(x)$ be the polytope similar to the outer simplex $\mF$ such that $\mF_r(x)$ has $\F$ and $\F_r(x)=\eta_r(x) \cap \mF$ as its two faces. Then, the proximity region $\N_{PE}(x,r)$ is defined to be $\mF_r(x)$.

The convex hull $C_H(\X_{1-j})$ has at least $d+1$ facets (exactly $d+1$ when $n_{1-j}=d+1$), and since each outer simplex is associated with a facet, the number of outer simplices is at least $d+1$. Let $\mS^{(2)}_{1-j} =\{\mF_1,\ldots,\mF_L\}$ denotes the set of all outer simplices. This construction handles the points in $\X_j \setminus C_H(\X_{1-j})$ only. Together with the points inside $C_H(\X_{1-j})$, the PE-PCD $D_j$, whose vertex set is $\V_j=\X_j$, has at least $$\sum_{k=1}^{K}I(\X_j \cap \mathfrak S_k \neq \emptyset) + \sum_{l=1}^{L}I(\X_j \cap \mF_l \neq \emptyset)$$ many components. 

\subsection{Minimum Dominating Sets} \label{sec:dominating-sets}

Our main contribution is the development of prototype-based classifiers with computationally tractable exact minimum prototype sets. We model the target class with a digraph $D$ such that prototype sets of the target class are equivalent to dominating sets of $D$. \cite{ceyhan2010} determined the appealing properties of \emph{minimum dominating set} of CCCDs in $\R$ as a guideline in defining new parametric digraphs relative to the Delaunay tessellation of the non-target class. In $\R$, CCCDs have computationally tractable minimum dominating sets, and the exact distribution of \emph{domination number} is known for target class points which are uniformly distributed within each cell. However, there is no polynomial time algorithm providing the exact minimum dominating sets of CCCDs in $\R^d$ for $d>1$. In this section, we provide a characterization of minimum dominating sets of PE-PCDs with barycentric coordinate systems and use them to introduce algorithms for finding these sets in polynomial time.   

We model the support of the class conditional distribution, i.e. $s(F_j)$, by a mixture of proximity regions. Our estimate for the support of the class $\X_j$ is $Q_j:=\cup_{x \in \X_j}\N(x)$ such that $\X_j \subset Q_j$. Nevertheless, the support of the target class $\X_j$ can be estimated by a cover with lower complexity (fewer proximity regions). For that purpose, we wish to reduce the model complexity by selecting an appropriate subset of proximity regions that still gives approximately the same estimate as $Q_j$; that is, let this cover be defined as $C_j:=\cup_{x \in S_j} \N_{PE}(x,r)$, where $S_j$ is a prototype set of points $\X_j$ such that $\X_j \subset C_j$. A reasonable choice of prototype sets for our class covers are the minimum dominating sets of PE-PCDs, whose elements are often more ``central" than the arbitrary sets of the same size. Dominating sets of minimum size are appealing since the size of the prototype sets determine the complexity of the model; that is, the smaller the set in cardinality (i.e. the model is lower in complexity), the higher the expected classification performance \citep{mehta1995,rissanen1989,gao2013}. 


In general, a digraph $D=(\V,\A)$ of order $n=|\V|$, a vertex $v$ {\em dominates} itself and all vertices of the form $\{u:\,(v,u) \in \A\}$. A {\em dominating set}, $S_D$, for the digraph $D$ is a subset of $\V$ such that each vertex $v \in \V$ is dominated by a vertex in $S_D$. A {\em minimum dominating set} (MDS), $S_{MD}$, is a dominating set of minimum cardinality, and the {\em domination number}, $\g(D)$, is defined as $\g(D):=|S_{MD}|$. If a minimum dominating set is of size one, we call it a {\em dominating point}. Finding a minimum dominating set is, in general, an NP-hard optimization problem \citep{karr1992,arora1996}. However, an approximately minimum dominating set can be obtained in $O(n^2)$ using a well-known greedy algorithm as in Algorithm~\ref{dom_greedy} \citep{chvatal:1979,parekh:1991}. PCDs using $\N_S(\cdot,\theta)$ (or CCCDs with parameter $\theta$) are examples of such digraphs. But, (exact) MDS of PCDs of maps $\N_{PE}(\cdot,r)$ are computationally tractable unlike PCDs with maps $\N_S(\cdot,\theta)$. Many attributes of these PE proximity maps and the proof of the existence of an algorithm to find a set $S_{MD}$ are conveniently implemented through the barycentric coordinate system. Before proving the results on the MDS, we give the following proposition.

\begin{algorithm}
\begin{algorithmic}[h]
 \REQUIRE A digraph $D=(\V,\A)$
 \ENSURE An approximate minimum dominating set, $S$
 \STATE  \textbf{set} $H=\V$ and $S = \emptyset$ 
 \WHILE{$H \neq \emptyset$}
  \STATE $v^{*} \leftarrow \argmax_{v \in \V(D)} |\{u \in \V(D): (v,u) \in \A(D)\}|$ 
  \STATE $S \leftarrow S \cup \{v^{*}\}$ 
  \STATE $H \leftarrow \V(D) \setminus \{u \in \V(D): (v^{*},u) \in \A(D)\}$
  \STATE $D \leftarrow D[H]$ 
 \ENDWHILE
\end{algorithmic}
\caption{The greedy algorithm for finding an approximate minimum dominating set of a digraph $D$. Here, $D[H]$ is the digraph induced by the set of vertices $H \subseteq \V$ \citep[see][]{west2000}.}
\label{dom_greedy}
\end{algorithm}

\begin{proposition} \label{prop:bary_line}
Let $\Y=\{\y_1,\y_2,\ldots, \y_{d+1}\} \subset \R^d$ be a set of non-coplanar points for $d>0$. For $x,x^* \in \mathfrak S=\mathfrak S(\Y)^o$, we have $d(x,f_i) < d(x^*,f_i)$ if and only if $w^{(i)}_{\mathfrak S}(x) < w^{(i)}_{\mathfrak S}(x^*)$ for all $i=1,\ldots,d+1$, where $d(x,f_i)$ is the distance between point $x$ and the face $f_i$.
\end{proposition}

\noindent {\bf Proof:}
For $i=1,\ldots,d+1$, note that $f_i$ is the face of the simplex $\mathfrak S$ opposite to the vertex $\y_i$. Let $L(\y_i,x)$ be the line through points $x$ and $\y_i$, and let $z \in f_i$ be the point that $L(\y_i,x)$ and $f_i$ cross at. Also, recall that $\eta(\y_i,x)$ denotes the hyperplane through the point $x$, and parallel to $f_i$. Hence, for $\al \in (0,1)$, $$ x= \al \y_i + (1-\al) z,$$ and since $z$ is a convex combination of the set $\{\y_k\}_{k \neq i}$,
\begin{align*}
 x = \al \y_i + \left( \sum_{k=1;k\neq i}^{d+1} (1-\al)\beta_k \y_k \right),
\end{align*}  
\noindent for $\beta_k \in (0,1)$. Thus, $w^{(i)}_{\mathfrak S}(x)=\al$ by the uniqueness of $\mathbf{w}_{\mathfrak S}(x)$. Observe that $\al=d(x,z)/d(\y_i,z)=d(x,f_i)/d(\y_i,f_i)$ since distances $d(x,z)$ and $d(x,f_i)=d(\eta(\y_i,x),f_i)$ are directly proportional. In fact, points that are on the same line parallel to $f_i$ have the same $i$'th barycentric coordinate $w^{(i)}_{\mathfrak S}(x)=\al$ corresponding to the vertex $\y_i$. Also, recall that with decreasing $\al$, the point $x$ gets closer to $f_i$ ($x \in f_i$ if $\al=0$, and $x=\y_i$ if $\al=1$). Then, for any two points $x,x^* \in \mathfrak S^o$, we have $w^{(i)}_{\mathfrak S}(x) < w^{(i)}_{\mathfrak S}(x^*)$ if and only if $d(\eta(\y_i,x),f_i) < d(\eta(\y_i,x^*),f_i)$ if and only if $d(x,f_i) < d(x^*,f_i)$. $\blacksquare$

Barycentric coordinates of a set of points in $\mathfrak S(\X_{1-j})$ help one characterize the set of \emph{local extremum} points, where a subset of local extremum points constitute the minimum dominating set $S_{MD}$. We use the Proposition~\ref{prop:bary_line} to prove the following theorem on $S_{MD}$ of a PE-PCD $D$.

\begin{theorem} \label{thm:dom_APE}
Let $\Z=\{z_1,z_2,\ldots, z_n\} \subset \R^d$ and $\Y=\{\y_1,\y_2,\ldots, \y_{d+1}\}\subset \R^d$ for $d>0$, and let $\mathfrak S=\mathfrak S(Y)$ be the $d$-simplex given by the set $\Y$ such that $\Z \subset \mathfrak S^o$. Hence, given the map $\N_{PE}(\cdot,r)$, we have $\g(D) \leq d+1$ for PE-PCD $D$ with vertex set $\V=\Z$. 
\end{theorem}

\noindent {\bf Proof:} 
Let $x,x^*,M \in \mathfrak S^o$. For $i=1,\ldots,d+1$, we show that there exists a point $x_{[i]} \in \Z \cap R_{M}(\y_i)$ such that $\Z \cap R_{M}(\y_i) \subset N_{PE}(x_{[i]},r)$ for all $r \in (1,\infty)$. It is easy to see that $d(x^*,f_i) < d(x,f_i)$ if and only if $\N_{PE}(x^*,r) \subset N_{PE}(x,r)$. Hence, $d(x_{[i]},f_i) = \min_{z \in \Z} d(z,f_i)$ if and only if $\N_{PE}(z,r) \subset N_{PE}(x_{[i]},r)$ for all $z \in \Z \cap R_{M}(\y_i)$. Also, by Proposition~\ref{prop:bary_line}, note that $d(x_{[i]},f_i) \leq \min_{z \in \Z} d(z,f_i)$ if and only if $w^{(i)}_{\mathfrak S}(x_{[i]}) \leq \min_{z \in \Z} w^{(i)}_{\mathfrak S}(z)$. Thus, the \emph{local extremum} point $x_{(i)}$ is given by $$x_{[i]} := \underset{x \in \Z \cap R_{M}(\y_i)}{\argmin} w^{(i)}_{\mathfrak S}(x).$$ Finally, observe that $\Z \subset \cup_{i=1}^{d+1} \N_{PE}(x_{[i]},r)$. Hence, the set of all local extremum points $\{x_{[1]},\ldots,x_{[d+1]}\}$ is a dominating set of the points $\Z \subset \mathfrak S^o$, so $\g(D) \leq d+1$. $\blacksquare$ 

MDSs of PE-PCDs are found by locating the \emph{local extremum} point $x_{[i]}$ of the vertex region $R_{M_C}(\y_i)$ for all $i=1,\ldots,d+1$. By Theorem~\ref{thm:dom_APE}, in $R_{M_C}(\y_i)$, the point $x_{[i]}$ is the closest points to the face $f_i$. For a set of $d$-simplices given by the Delaunay tesselation of $\X_{1-j}$, Algorithm~\ref{dom_NPE} identifies all the local extremum points of each $d$-simplex in order to find the (exact) minimum dominating set $S_j=S_{MD}$.

Let $D_j=(\V_j,\A_j)$ be a PE-PCD with vertex $\V=\X_j$. In Algorithm~\ref{dom_NPE}, we partition $\X_j$ into such subsets that each subset falls into a single $d$-simplex of the Delaunay tesselation of the set $\X_{1-j}$.  Let $\mS_{1-j}$ be the set of all $d$-simplices associated with $\X_{1-j}$. Moreover, for each $\mathfrak S \in \mS_{1-j}$, we further partition the subset $\X_j \cap \mathfrak S$ into subsets that each subset falls into a single vertex region of $\mathfrak S$. In each vertex region $R_{M_C}(\y_i)$, we find the \emph{local extremum} point $x_{[i]}$. Let $S(D)$ denote the minimum dominating set and $\g(D)$ denote the domination number of a digraph D. Also, let $D_j[\mathfrak S]$ be the digraph induced by points of $\X_j$ inside the $d$-simplex $\mathfrak S$, i.e. $\X_j \cap \mathfrak S$. Recall that, as a result of Theorem~\ref{thm:dom_APE}, $\g(D_j[\mathfrak S]) \leq d+1$ since $\X_j \cap \mathfrak S \subset  \cup_{i=1}^{d+1} \N_{PE}(x_{[i]},r)$. To find $S(D_j[\mathfrak S])$, we check all subsets of the set of local extremum points, from smallest cardinality to highest, and check if $\X_j \cap \mathfrak S$ is in the union of proximity regions of these subsets of local extremum points. For example, $S(D_j[\mathfrak S])=\{x_{[l]}\}$ and $\g(D_j[\mathfrak S])=1$ if $\X_j \cap R_{M_C}(\y_i) \subset N_{PE}(x_{[l]},r)$ for some $l=1,2,3$; else $S(D_j[\mathfrak S])=\{x_{[l_1]},x_{[l_2]}\}$ and $\g(D_j[\mathfrak S])=2$ if $\X_j \cap R_{M_C}(\y_i) \subset N_{PE}(x_{[l_1]},r) \cup N_{PE}(x_{[l_2]},r)$ for some $\{l_1,l_2\} \in \binom{\{1,2,3\}}{2}$; or else $S(D_j[\mathfrak S])=\{x_{[1]},x_{[2]},x_{[3]}\}$ and $\g(D_j[\mathfrak S])=3$ if $\X_j \cap R_{M_C}(\y_i) \subset \cup_{l=1,2,3} \N_{PE}(x_{[l]},r)$. The resulting minimum dominating set of $D_j$ for $\X_j \cap C_H(\X_{1-j})$ is the union of these sets, i.e., $S_j=\cup_{\mathfrak S \in \mS_{1-j}} S(D_j[\mathfrak S])$ and $\g(D_j)=|S_j|$. Observe that $S(D_j[\mathfrak S]) = \emptyset$ if $\X_j \cap \mathfrak S = \emptyset$. This algorithm is guaranteed to terminate, as long as $n_0$ and $n_1$ are both finite.

\begin{algorithm}[h]
\begin{algorithmic}[1]
 \REQUIRE The target class $\X_j$, a set of $d$-simplices of the non-target class $\mS_{1-j}$, and the PE proximity map $\N_{PE}(\cdot,r)$.
 \ENSURE The minimum dominating set, $S_j$
 \STATE  $S_{j} = \emptyset$ 
 \FORALL{$\mathfrak S \in \mathcal{S}_{1-j}$ where $ \X_j \cap \mathfrak S \neq \emptyset$ }
  \STATE $\X^{*}_{j} \leftarrow \X_j \cap \mathfrak S$ and let $\{\y_1,\ldots,\y_{d+1}\}$ be the vertices of $\mathfrak S.$
  \FOR{$i=1,\ldots,d+1$}
  	\STATE Let $x_{[i]} \leftarrow \underset{x \in \X^{*}_{j} \cap R_{M_C}(\y_i)}{\argmin} w^{(i)}_{\mathfrak S}(x).$
  \ENDFOR
  \FOR{$t=1,\ldots,d+1$}
  	\IF{there exists a set $\{l_1,\ldots,l_t\} \in \dbinom{\{1,\ldots,d+1\}}{t}$ s.t. $\X^{*}_{j} \subset \cup_{a=1}^{t} \N_{PE}(x_{[l_a]},r)$}
  		\STATE $S_{j} \leftarrow S_{j} \cup \{x_{[l_1]},\ldots,x_{[l_t]}\}$
  		\BREAK 
    \ENDIF
  \ENDFOR 
 \ENDFOR
\end{algorithmic}
\caption{The algorithm for finding the (exact) minimum dominating set $S_{j}$ of a PE-PCD $D_j$ induced by $\X_j \cap C_H(\X_{1-j})$.}
\label{dom_NPE}
\end{algorithm}

The level of reduction depends also on the magnitude of the expansion parameter $r$. In fact, the larger the magnitude of $r$, the more likely the $S(D_j[\mathfrak S])$ have smaller cardinality, i.e. the more the reduction in the data set. Thus, we have a stochastic ordering as follows:

\begin{theorem} \label{thm:rparam}
Let $\g(D_j[\mathfrak S],r)$ be the domination number the PE-PCD $D_j(\mathfrak S)$ with expansion parameter $r$. Then for $r_1<r_2$, we have $\g(D_j[\mathfrak S],r_2) \leq^{ST} \g(D_j[\mathfrak S],r_1)$ where $\leq^{ST}$ stands for ``stochastically smaller than". 
\end{theorem}

\noindent {\bf Proof:}
Suppose $r_1<r_2$. Then in a given simplex $\mathfrak S_k$ for $k=1,\ldots,K$, let $\g_k(r):=\g(\mathfrak S_k,r)$ be the domination number of the component of the PE-PCD $D_j$ whose vertices are restricted to the interior of $\mathfrak S_k$. Let $\Z=\{Z_1,Z_2,\ldots,Z_n\}$ be a set of i.i.d. random variables drawn from a continuous distribution $F$ whose support is $\mathfrak S_k$, and let $Z_{[i]}$ be the local extremum point of $\Z \cap R_{M_C}(\y_i)$ where $\y_i$ being the $i$'th vertex of $\mathfrak S_k$. Also, let $\vol(N_{PE}(x,r))$ be the volume of the $\N_{PE}(x,r)$ of a point $x \in \mathfrak S_k^o$. Note that, $$\vol(N_{PE}(x,r_1)) < \vol(N_{PE}(x,r_2)).$$ Hence, since $\N_{PE}(x,r_1) \subset \N_{PE}(x,r_2)$, $$\vol(\N_{PE}(Z_{[i]},r_1)) \leq^{ST} \vol(\N_{PE}(Z_{[i]},r_2)).$$ Now, let $\{l_1,\ldots,l_t\} \in \dbinom{\{1,\ldots,d+1\}}{t}$ be any set of indices associated with a subset of all local extremum points $t=1,\ldots,d+1$. Thus, $$\vol \left( \cup_{b=1}^t \N_{PE}(Z_{[l_b]},r_1) \right) \leq^{ST} \vol \left( \cup_{b=1}^t \N_{PE}(Z_{[l_b]},r_2) \right).$$ Hence, given that the event $\Z \subset \left( \cup_{b=1}^t \N_{PE}(Z_{[l_b]},r) \right)$ implies $\g_k(r) \leq t$, we can show that $$ P\left(\Z \subset \left( \cup_{b=1}^t \N_{PE}(Z_{[l_b]},r_2) \right) \right) \leq  P \left( \Z \subset \left( \cup_{b=1}^t \N_{PE}(Z_{[l_b]},r_1) \right) \right),$$ and $$P(\g_k(r_2) \leq t) \leq P(\g_k(r_1) \leq t)$$ for $t=1,\ldots,d+1$. $\blacksquare$

	Algorithm~\ref{dom_NPE} ignores the target class points outside the convex hull of the non-target class. This is not the case with Algorithm~\ref{dom_greedy}, since the map $\N_S(\cdot,\theta)$ is defined over all points $\X_j$ whereas the original PE proximity map $\N_{PE}(\cdot,r)$ is not. Hence, the prototype set $S_j$ only yields a reduction in the set $\X_j \cap C_H(\X_{1-j})$. Solving this issue requires different approaches. One solution is to define covering methods with two proximity maps that are the PE proximity map and the other which does not require the target class points to be inside the convex hull of the non-target class points, e.g. spherical proximity regions (proximity maps $N_S(\cdot,\theta)$). 
	
	Algorithm~\ref{dom_NPEandNS} uses both maps $\N_{PE}(\cdot,r)$ and $\N_S(\cdot,\theta)$ to generate a prototype $S_j$ for the target class $\X_j$. There are two separate MDSs, $S^{(1)}_j$ which is exactly minimum, and $S^{(2)}_j$ which is approximately minimum. Each of the two maps is associated with two distinct digraphs such that $\X_j \cap C_H(\X_{1-j})$ constitutes the vertex set of one digraph and $\X_j \setminus C_H(\X_{1-j})$ constitute the vertex of another, where the non-target class is always $\X_{1-j}$. Algorithm~\ref{dom_NPE} finds a prototype set $S^{(1)}_j$ for $\X_j \cap C_H(\X_{1-j})$, and then the prototype set $S^{(2)}_j$ for $\X_j \setminus C_H(\X_{1-j})$ is appended to the overall prototype set $S_j=S^{(1)}_j \cup S^{(2)}_j$ as in Algorithm~\ref{dom_NPEandNS}. Note that the set $S_j$ is an approximate minimum dominating set since $S^{(2)}_j$ is approximately minimum. 
	
\begin{algorithm}[h]
\begin{algorithmic}[1]
 \REQUIRE The target class $\X_j$, a set of $d$-simplices of the non-target class $\mS_{1-j}$, and the proximity maps $\N_{PE}(\cdot,r)$ and $\N_S(\cdot,\theta)$.
 \ENSURE The approximate minimum dominating set, $S_j$
 \STATE  $S^{(1)}_{j} = \emptyset$ and $S^{(2)}_{j} = \emptyset$
 \STATE  Find the minimum dominating set of $\X_j \cap C_H(X_{1-j})$ in Algorithm 2 and assign it to $S^{(1)}_j$
 \STATE  $\X'_j = \X_j \setminus C_H(\X_{1-j})$. 
 \STATE  Find the approximate minimum dominating set as in Algorithm 1 where the target class is $\X'_j$ and the non-target class is $\X_{1-j}$, and assign it to $S^{(2)}_{j}$
 \STATE $S_j = S^{(1)}_{j} \cup S^{(2)}_{j}$
\end{algorithmic}
\caption{The algorithm for finding the minimum dominating set $S_{j}$ of PCD $D_j$ defined by the proximity maps $\N_{PE}(\cdot,r)$ and $\N_S(\cdot,\theta)$.}
\label{dom_NPEandNS}
\end{algorithm}

Algorithm~\ref{dom_NPE_out} uses only the PE proximity map $\N_{PE}(\cdot,r)$ with the original version inside $C_H(\X_{1-j})$ and extended version outside $C_H(\X_{1-j})$. The cover is a mixture of $d$-simplices and $d$-polytopes. Given a set of $d$-simplices $\mS^{(1)}_{1-j}$ and a set of outer simplices $\mS^{(2)}_{1-j}$, we find the respective local extremum points of each $d$-simplex and outer simplex. Local extremum points of $d$-simplices are found as in Algorithm~\ref{dom_NPE}, and then we find the local extremum points of the remaining points to get the prototype set of the entire target class $\X_j$. The following theorem provides a result on the local extremum points in an outer simplex $\mF$. Note that, in Algorithm~\ref{dom_NPE_out}, the set $S_j$ is the exact minimum dominating set since both $S^{(1)}_j$ and $S^{(2)}_j$ are exact MDSs for the PE-PCDs induced by $\X_j \cap C_H(\X_{1-j})$ and $\X_j \setminus C_H(\X_{1-j})$, respectively. 

\begin{theorem} \label{thm:dom_ES}
Let $\Z=\{z_1,z_2,\ldots, z_n\} \subset \R^d$, let $\F$ be a facet of the $C_H(\X_{1-j})$ and let $\mF$ be the associated outer simplex such that $\Z \subset \mF^o$. Hence, the local extremum point and the $S_{MD}$ of the PE-PCD $D$ restricted to $\mF^o$ is found in linear time and is equal to 1.
\end{theorem}

\noindent {\bf Proof:}
We show that there is a point $s \in \Z$ such that $\X \subset \N_{PE}(s,r)$ for all $r \in (1,\infty)$. As a remark, note that $\eta(x,\F)$ denotes the hyperplane through $x$, and is parallel to $\F$. Thus, for $x,x^* \in \F^o$, observe that $d(x,\F) < d(x^*,\F)$ if and only if $d(\eta(x,\F),\F) < d(\eta(x^*,\F),\F)$ if and only if $\N_{PE}(x,r) \subset \N_{PE}(x^*,r)$. Thus, the \emph{local extremum} point $s$ is given by $$s := \underset{x \in \Z}{\argmax} \> d(x,\F).$$ Therefore, $S_{MD}=\{s\}$ yields the result. $\blacksquare$ 

\begin{algorithm}[h]
\begin{algorithmic}[1]
 \REQUIRE The target class $\X_j$, the set $\mS^{(1)}_{1-j}$, the set $\mS^{(2)}_{1-j}$.
 \ENSURE The minimum dominating set, $S_j$
 \STATE  $S^{(1)}_{j} = \emptyset$ and $S^{(2)}_{j} = \emptyset$
 \STATE  Find the minimum dominating set in Algorithm 2 and assign it to $S^{(1)}_j$
 \STATE  $\X'_j = \X_j \setminus C_H(\X_{1-j})$. 
 \FORALL{$\mF \in \mS^{(2)}_{1-j}$ where $\X'_j \cap \mF \neq \emptyset$ }
  \STATE $\X^{*}_{j} \leftarrow \X'_j \cap \mF$
  \STATE Let $s \in \X^{*}_{j}$ be the local extremum point in $\mF$
  \STATE $S^{(2)}_{j} \leftarrow S^{(2)}_{j} \cup \{s\}$
 \ENDFOR 
 \STATE $S_j = S^{(1)}_{j} \cup S^{(2)}_{j}$
\end{algorithmic}
\caption{The algorithm for finding the (exact) minimum dominating set $S_{j}$ of PE-PCD $D_j$ with vertex set $\X_j$.}
\label{dom_NPE_out}
\end{algorithm}

	Given Theorems~\ref{thm:dom_APE} and~\ref{thm:dom_ES}, Algorithm~\ref{dom_NPE_out} may be the most appealing algorithm since it gives the exact minimum dominating set for the complete target class $\X_j$. However, the following theorem show that the cardinality of such sets increase exponentially on dimensionality of the data set, even though it is polynomial on the number of observations. 

\begin{theorem} \label{thm:complexity}
	Algorithm~\ref{dom_NPE_out} finds an exact minimum dominating set $S_j$ of the target class $\X_j$ in $\mathcal{O}(d^k n^2_{1-j} + 2^d n_{1-j}^{\ceil{d/2}})$ time for $k >1$ where $|S_j| = \mathcal{O}(dn_{1-j}^{\ceil{d/2}})$.
\end{theorem}	 

\noindent {\bf Proof:} A Delaunay tesselation of the non-target class $\X_{1-j} \subset \R^d$ is found in $\mathcal{O}(d^k n^2_{1-j})$ time with the Bowyer-Watson algorithm for some $k>1$, depending on the complexity of the algorithm that finds the circumcenter of a $d$-simplex \citep{watson1981}. The resulting tesselation with $n_{1-j}$ vertices has at most $\mathcal{O}(n_{1-j}^{\ceil{d/2}})$ simplices and at most $\mathcal{O}(n_{1-j}^{\floor{d/2}})$ facets \citep{seidel1995}. Hence the union of sets of $d$-simplices $\mS^{(1)}_{j}$ and outer simplices $\mS^{(2)}_{1-j}$ is of cardinality at most $\mathcal{O}(n_{1-j}^{\ceil{d/2}})$. Now, for each simplex $\mathfrak S \in \mS^{(1)}_{j}$ or each outer simplex $\F \in \mS^{(2)}_{j}$, the local extremum points are found in linear time. Each simplex is divided into $d+1$ vertex regions with each having their own set of local extremum points. A minimum cardinality subset of the set of local extremum points is of cardinality at most $d+1$ and found in a brute force fashion. For outer simplices, however, the local extremum point is the farthest point to the associated facet of the Delaunay tesselation. Thus, it takes at most $\mathcal{O}(2^d)$ and $\mathcal{O}(1)$ time to find the exact minimum subsets of local extremum points for each simplex and outer simplex, respectively. Then the result follows. $\blacksquare$

Theorem~\ref{thm:complexity} shows the exponential increase of the number of prototypes as  dimensionality increases. Thus, the complexity of the class cover model also increases exponentially, which might lead to overfitting. We will investigate this issue further in Sections~\ref{sec:simulations} and~\ref{sec:realdata}.

\section{PCD covers} \label{sec:pcdcovers}

We establish class covers with the PE proximity map $\N_{PE}(\cdot,r)$ and spherical proximity map $\N_{S}(\cdot)$. We define two types of class covers: one type is called \emph{composite covers} where we cover the points in $\X_j \cap C_H(\X_{1-j})$ with PE proximity maps and the points in $\X_j \setminus C_H(\X_{1-j})$ with spherical proximity maps, and the other is called \emph{standard cover} incorporating the PE proximity maps for all points in $\X_j$. We use these two types of covers to establish a specific type of classifier that is more appealing in the sense of prototype selection. 

\begin{figure}[!h]
\centering
\begin{tabular}{ccc}
\includegraphics[scale=0.32]{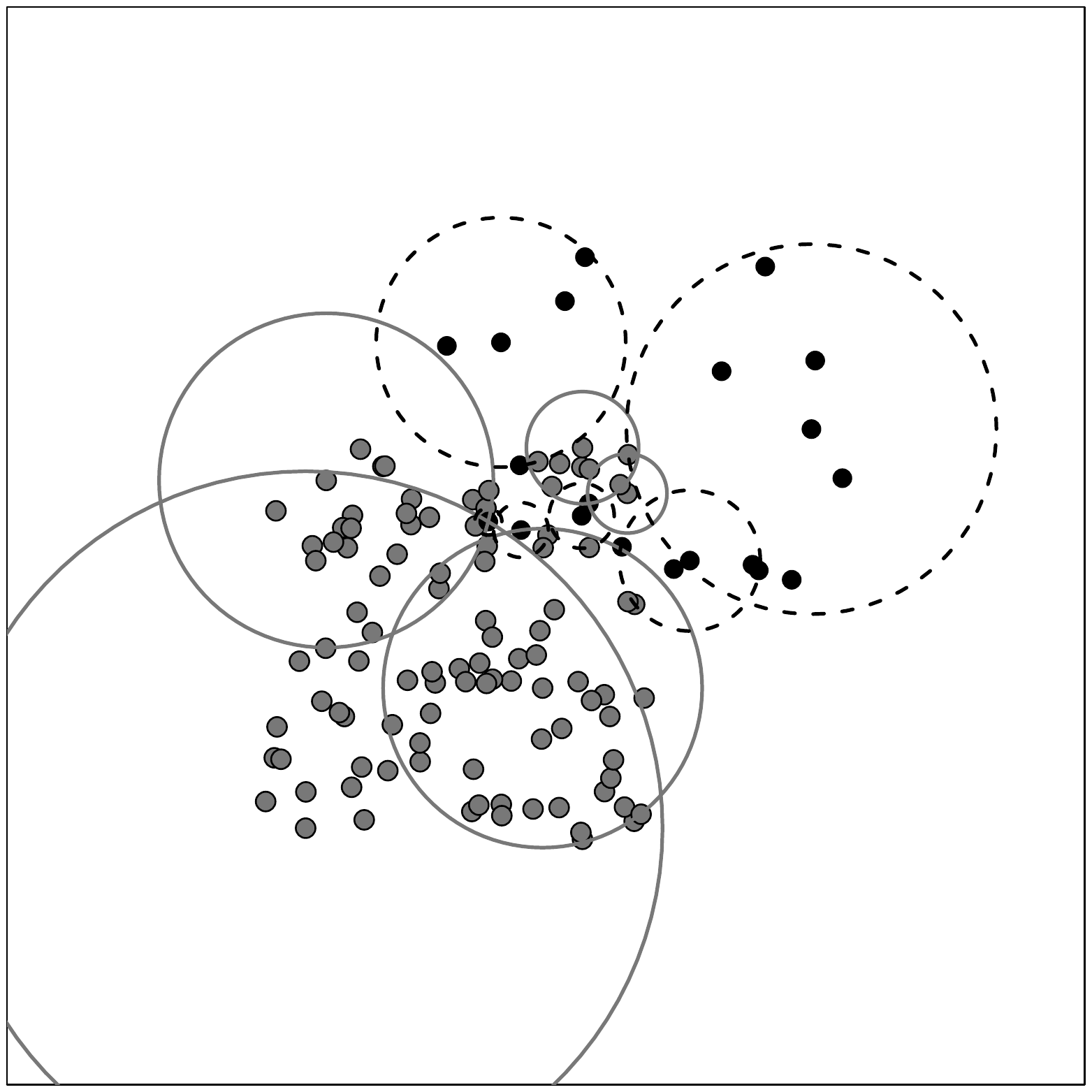} & \includegraphics[scale=0.32]{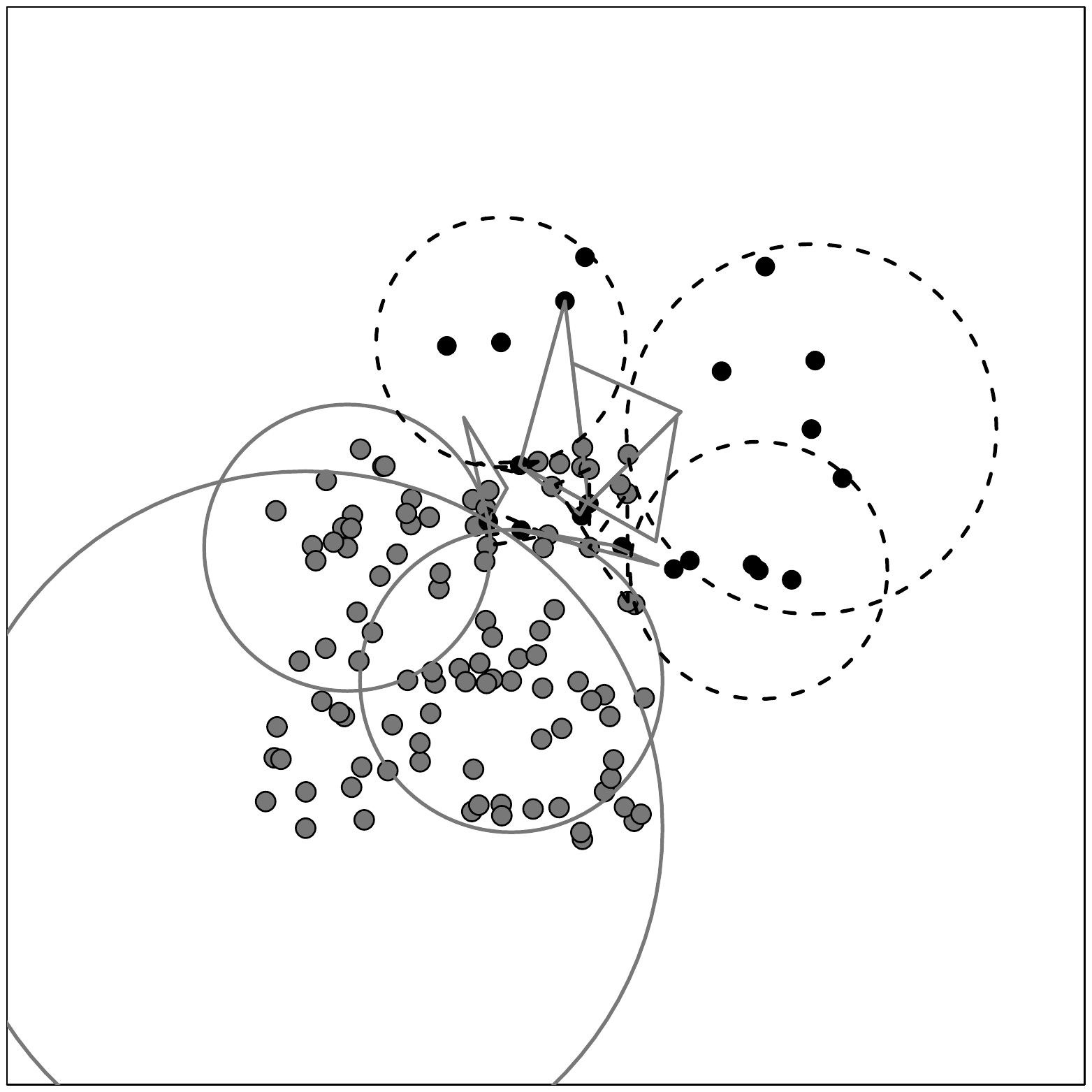} & \includegraphics[scale=0.32]{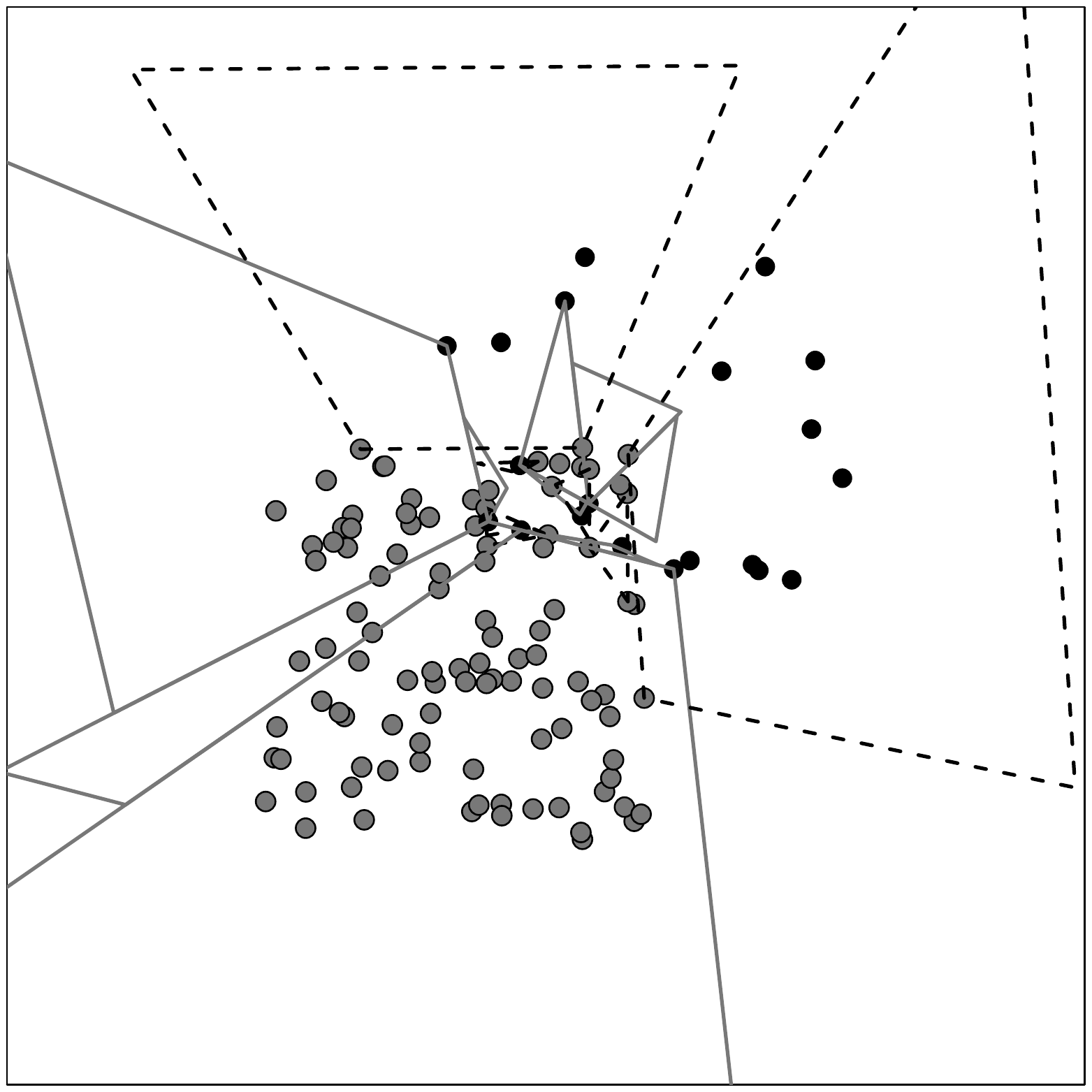} \\
(a) & (b) & (c) \\
\end{tabular}
\caption{Class covers of a data set with two-class setting in $\R^2$ where grey and black points represent points of two separate classes. The training data set is composed of two classes $\X_0$ and $\X_1$ wherein 100 and 20 samples are drawn from multivariate uniform distributions $U([0,1]^2)$ and $U([0.5,1.5]^2)$, respectively. Cover of one class is given by solid circle and solid line segments, and the cover of the other is given by dashed circle and dashed line segments. (a) Standard class covers with $\N_I(\cdot)=\N_O(\cdot)=\N_{S}(\cdot,\theta=1)$ (b) Composite class cover with $\N_I(\cdot)=\N_{PE}(\cdot,r=2)$ and $\N_O(\cdot)=\N_{S}(\cdot,\theta=1)$ (c) Standard class covers with $\N_I(\cdot)=\N_O(\cdot)=\N_{PE}(\cdot,r=2)$.}
\label{fig:cover}
\end{figure}

Our composite covers are mixtures of simplical and spherical proximity regions. Specifically, given a set of simplices and a set of spheres, the composite cover is the union of both these sets which constitute proximity regions of two separate PCD families, hence the name \emph{composite cover}. The $Q_j$ is partitioned into two: the cover $Q^{(1)}_j$ of points inside the convex hull of non-target class points, i.e., $\X_j \cap C_H(\X_{1-j})$, and the cover $Q^{(2)}_j$ of points outside, i.e., $\X_j \setminus C_H(\X_{1-j})$. Let $Q_j^{(1)}:=\cup_{x \in \X_j \cap C_H(\X_{1-j})}N_I(x)$ and $Q^{(2)}_j:=\cup_{x \in \X_j \setminus C_H(\X_{1-j})} N_O(x)$ such that $Q_j:=Q^{(1)}_j \cup Q^{(2)}_j$. Here, $\N_I(\cdot)$ and $\N_O(\cdot)$ are proximity maps associated with sets $\X_j \cap C_H(\X_{1-j})$ and $\X_j \setminus C_H(\X_{1-j})$, respectively. Hence, in composite covers, target class points inside $C_H(\X_{1-j})$ are covered with PE proximity map $\N_I(\cdot)=\N_{PE}(\cdot,r)$, and the remaining points are covered with spherical proximity map $\N_O(\cdot)=\N_{S}(\cdot,\theta)$. Given the covers $Q^{(1)}_j$ and $Q^{(2)}_j$, let $C^{(1)}_j$ and $C^{(2)}_j$ be the class covers with lower complexity associated with the dominating sets $S^{(1)}_j$ and $S^{(2)}_j$. Hence the composite cover is given by $$C_j:= C^{(1)}_j \cup C^{(2)}_j = \Bigg\{ \bigcup_{s \in S^{(1)}_j} N_I(s) \Bigg\} \bigcup \Bigg\{ \bigcup_{s \in S^{(2)}_j} N_O(s) \Bigg\}. $$ An illustration of the class covers $C_0$ and $C_1$ with $\N_I(\cdot)=\N_{PE}(\cdot,r=2)$ and $\N_O(\cdot)=\N_{S}(\cdot,\theta=1)$ is given in Figure~\ref{fig:cover}(b). 

By definition, the spherical proximity map $\N_{S}(\cdot,\theta)$ yields class covers for all points in $\X_j$. Figure~\ref{fig:cover}(a) illustrates the class covers of the map $\N_{S}(\cdot,\theta=1)$. We call such covers, that only constitute a single type of proximity map, as \emph{standard covers}. Hence the standard cover of the PE-PCD $D_j$ is a union of $d$-simplices and $d$-polytopes: $$C_j:= \bigcup_{s \in S_j} \N_{PE}(s,r). $$ Here, $\N_I(\cdot)=\N_O(\cdot)=\N_{PE}(\cdot,r)$. An illustration is given in Figure~\ref{fig:cover}(c). 

PCD covers can easily be generalized to the multi-class case with $J$ classes. To establish the set of covers $\mathcal{C} = \{C_1,C_2, \ldots, C_J\}$, the set of PCDs $\D=\{D_1,\ldots,D_J\}$, and the set of MDSs $\mcS=\{S_1,S_2\ldots,S_J\}$ associated with a set of classes $\mathfrak{X} = \{\X_1,X_2,\ldots,\X_J\}$, we gather the classes into two classes as $\X_T=\X_j$ and $\X_{NT}=\cup_{t \neq j} \X_t$ for $t,j=1,\ldots,J$. We refer to classes $\X_{T}$ and $\mathcal{X}_{NT}$ as target and non-target class, respectively. More specifically, target class is the class we want to find the cover of, and the non-target class is the union of the remaining classes. We transform the multi-class case into the two-class setting and find the cover of $j$'th class, $C_j$.

\section{Classification with PCDs} \label{sec:classification}

The elements of $S_j$ are prototypes, for the problem of modelling the class conditional discriminant regions via a collection of proximity regions (balls, simplices, polytopes, etc.). The sizes of these regions represent an estimate of the domain of influence, which is the region in which a given prototype should influence the class labelling. Our semi-parametric classifiers depends on the class covers given by these proximity regions. We define various classifiers based on the class covers (composite or standard) and some other classification methods. We approach classification of points in $\R^d$ in two ways:

\begin{description}
 \item[Hybrid classifiers:] Given the class covers $C^{(1)}_0$ and $C^{(1)}_1$ associated with classes $\X_0$ and $\X_1$, we classify a given point $z \in \R^d$ with $g_P$ if $z \in C^{(1)}_0 \cup C^{(1)}_1$, and with $g_A$ otherwise. Here, $g_P$ is the pre-classifier and $g_A$ is an alternative classifier.  
 \item[Cover classifiers:] These classifiers are constructed by class covers only; that is, a given point $z \in \R^d$ is classified as $g_C(z)=j$ if $z \in C_j \setminus C_{1-j}$ or if $\rho(z,C_j) < \rho(z,C_{1-j})$, hence class of the point $z$ is estimated as $j$ if $z$ is only in cover $C_j$, or closer to $C_j$ than $C_{1-j}$. Here, $\rho(z,C_j)$ is a dissimilarity measure between point $z$ and the cover $C_j$. Cover classifiers depend on the types of covers which are either \emph{composite} or \emph{standard} covers. 
\end{description}

	We incorporate PE-PCDs for establishing both of these types of classifiers. Hence, we will refer to them as hybrid PE-PCD and cover PE-PCD classifiers. Since the PE proximity maps were originally defined for points $\X_j \cap C_H(\X_{1-j})$, we develop hybrid PE-PCD classifiers to account for points outside of the convex hull of the non-target class in a convenient fashion. However, as we shall see later, cover PE-PCD classifiers have much more appealing properties than hybrid PE-PCD classifiers in terms of both efficiency and classification performance. Nevertheless, we consider and compare both types of classifiers, but first we define the PE-PCD pre-classifier.  

\subsection{PE-PCD Pre-classifier}

Let $\rho(z,C)$ be the dissimilarity measure between $z$ and the class cover $C$. The PE-PCD pre-classifier is given by
\begin{equation}
g_P(z):= \left\{ \begin{array}{ll}
        j & \text{if $z \in C^{(1)}_j \setminus C^{(1)}_{1-j}$ for $j=0,1$} \\
        I(\rho(z,C^{(1)}_1) < \rho(z,C^{(1)}_0)) & \text{if $z \in C^{(1)}_0 \cap C^{(1)}_1$} \\
        -1 & \text{otherwise}. \\
        \end{array}
  \right.  
\end{equation}

\begin{figure}[h]
\centering
\begin{tabular}{ccc}
\includegraphics[scale=0.35]{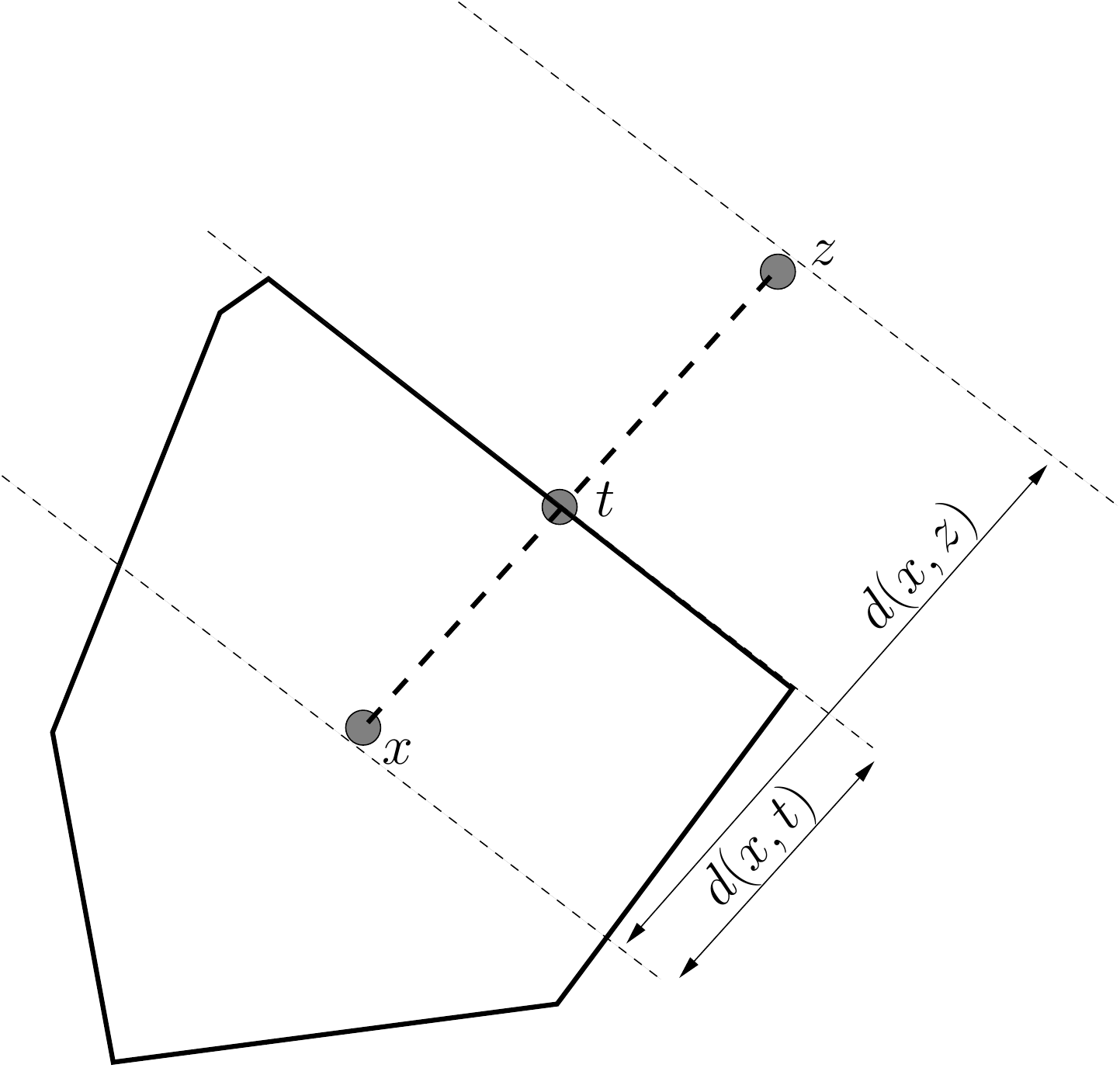} & \includegraphics[scale=0.32]{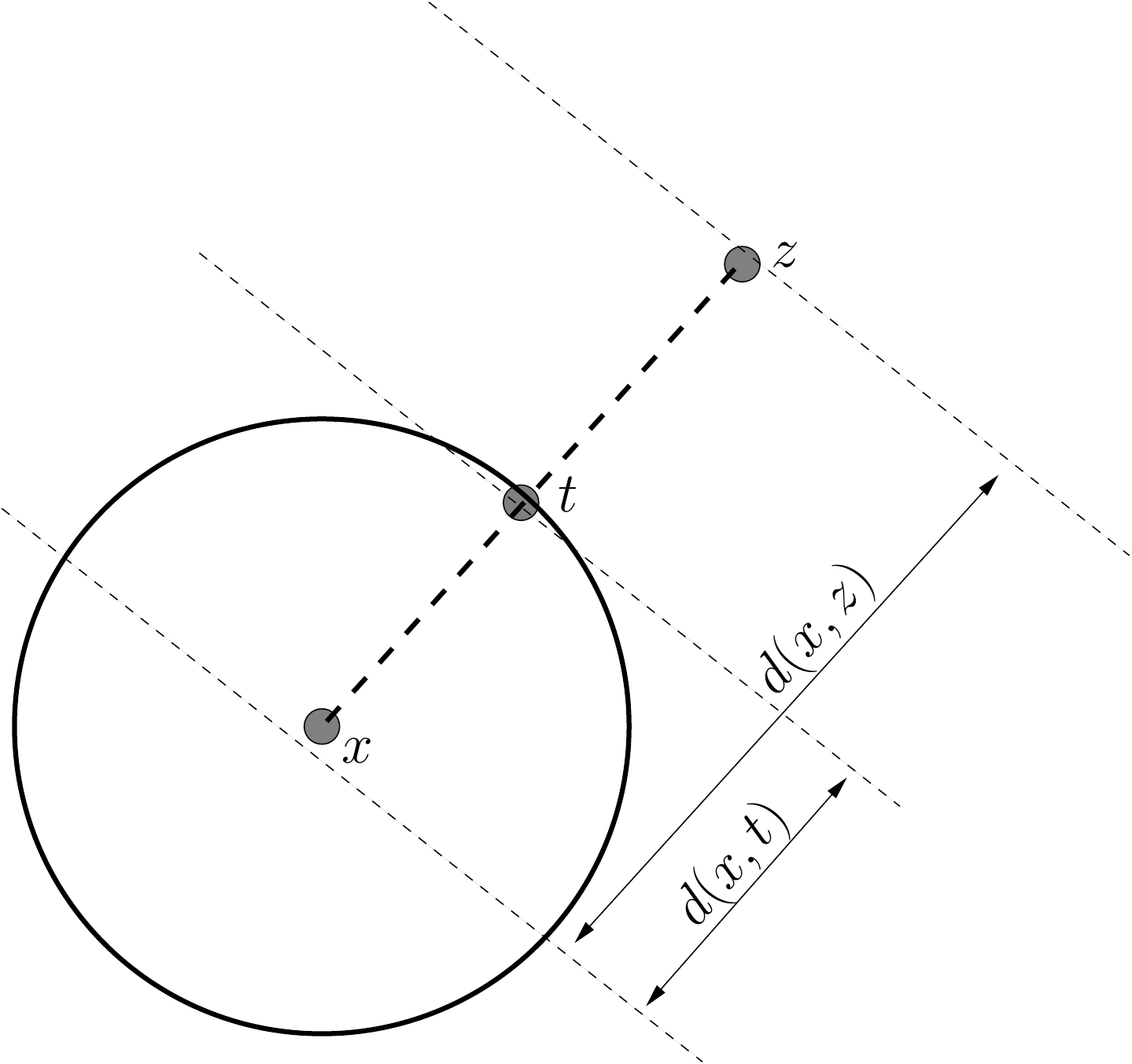} & \includegraphics[scale=0.32]{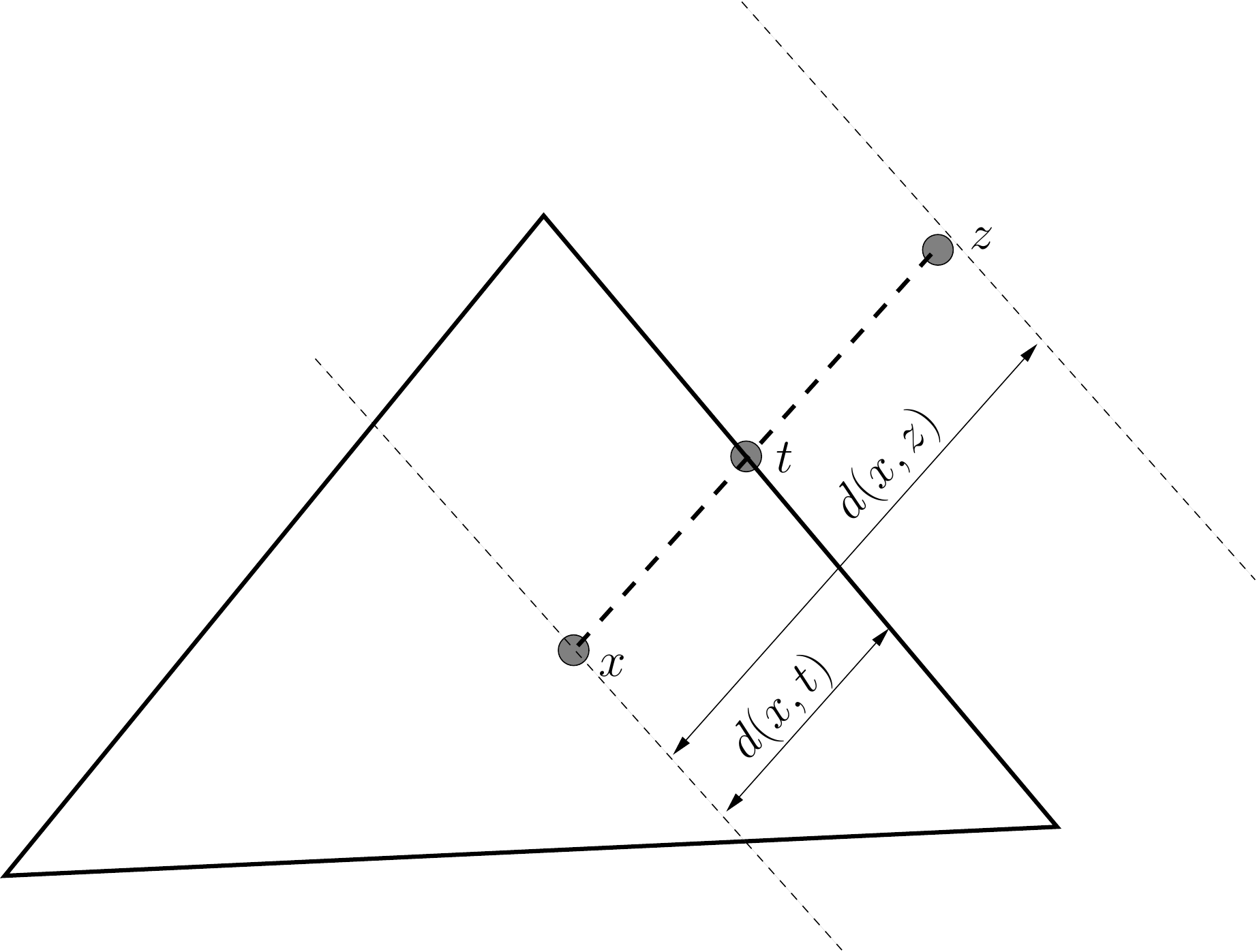} \\
(a) & (b) & (c) \\
\end{tabular}
\caption{Illustration of a convex distance between a a point $z$ and an arbitrary (a) convex set $H$, (b) ball and (c) $2$-simplex in $\R^2$.}
\label{fig:convdist}
\end{figure}

\noindent Here, $I(\cdot)$ is the indicator functional and $g_P(z)=-1$ denotes a ``no decision" case. Given that class covers $C^{(1)}_0$ and $C^{(1)}_1$ are the unions of PE proximity regions $\N_{PE}(x,r)$ of points in dominating sets  $S^{(1)}_0$ and $S^{(1)}_1$, the closest cover is found by, first, checking the proximity region of a cover closest to the point $z$: $$ \rho(z,C^{(1)}_j) = \min_{s \in S^{(1)}_j} \rho(z,N(s))$$ which is expressed based on a dissimilarity measure between a point $z$ and the region $\N(s)$. For such measures, we employ convex distance functions. Let $H$ be a convex set in $\R^d$ with center $x \in H$. The point $x$ may be viewed as the center of the set $H$. Thus, let the dissimilarity between $z$ and $H$ be defined by $$ \rho(z,H):=\frac{d(z,x)}{d(t,x)},$$ where $d(\cdot,\cdot)$ is the Euclidean distance and $t$ is a point on the line $L(x,z):=\{x+\al(z-x):\al \in [0,\infty)\}$ such that $t \in \partial(H)$, the boundary of the $H$. An illustration is given in Figure~\ref{fig:convdist} for several convex sets, including balls and simplices in $\R^2$.  

For spherical proximity map $\N_{S}(\cdot,\theta)$, the dissimilarity function is defined by the radius of that ball which is a spherical proximity region: $d(x,t)=\varepsilon_{\theta}(x)$ \citep{priebe:2003b}. However, for $d$-simplices, we characterize the dissimilarity measure in terms of barycentric coordinates of $z$ with respect to $\mathfrak S(x)=\N_{PE}(x,r)$. 

\begin{proposition} \label{prop:simpdist}
	Let $\{t_1,t_2, \ldots,t_{d+1}\} \subset \R^d$ be a set of non-collinear points that are the vertices of simplex $\mathfrak S(x)=\N_{PE}(x,r)$ with the median $M_C(x) \in \mathfrak S(x)^o$. Then, for $z \in \R^d$ and $t \in \partial(\mathfrak S(x))$, $$ \rho(z,\mathfrak S(x))=\frac{d(M_C(x),z)}{d(M_C(x),t)}=1-(d+1) w^{(k)}_{\mathfrak S(x)}(z),$$ where $w^{(k)}_{\mathfrak S(x)}(z)$ being the $k$'th barycentric coordinate of $z$ with respect to $\mathfrak S(x)$. Moreover, $\rho(z,\mathfrak S(x)) < 1$ if $z \in \mathfrak S(x)^o$ and $\rho(z,\mathfrak S(x)) \geq 1$ if $z \not\in \mathfrak S(x)^o$. 
\end{proposition}

\noindent {\bf Proof:} 
Let the line segment $L(M_C(x),z)$ and $\partial(\mathfrak S(x))$ cross at the point $t \in f_k$ for $f_k$ being the face of $\mathfrak S(x)$ opposite to $t_k$. Thus, for $\al_i \in (0,1)$ and $\beta \in (0,1)$, $$z = (1-\beta) M_C(x) + \beta t = (1-\beta) M_C(x) + \beta \left( \sum_{i=1;i \neq k}^{d+1} \al_i t_i \right).$$ Here, note that $\beta=d(M_C(x),z)/d(M_C(x),t)=\rho(z,\mathfrak S(x))$. Also, since $M_C(x)$ is the median, $$z = (1-\beta) \frac{\sum_{i=1}^{d+1} t_i}{d+1} + \beta \left( \sum_{i=1;i \neq k}^{d+1} \al_i t_i \right) = \frac{1-\beta}{d+1} t_k + \sum_{i=1;i \neq k}^{d+1} \left( \frac{1-\beta}{d+1} + \beta\al_i \right) t_i. $$ Hence $(1-\beta)/(d+1)=w^{(k)}_{\mathfrak S(x)}(z)$ which implies $\beta=1-(d+1)w^{(k)}_{\mathfrak S(x)}(z)$. Therefore, $z \in \mathfrak S(x)^o$ if and only if $\beta=1-(d+1)w^{(k)}_{\mathfrak S(x)}(z) < 1$. $\blacksquare$

For a (convex) proximity region $\N_{PE}(x,r)$, the dissimilarity measure $\rho(z,\mathfrak S(x))=\rho(z,\N_{PE}(x,r))$ indicates whether or not the point $z$ is in proximity region $\N_{PE}(x,r)$, since $\rho(z,\mathfrak S(x)) < 1$ if $z \in \N_{PE}(x,r)$ and $ \geq 1$ otherwise. Hence, the PE-PCD pre-classifier $g_P$ may simply be defined by
\begin{equation}
g_P(z):= \left\{ \begin{array}{ll}
        I(\rho(z,C^{(1)}_1) < \rho(z,C^{(1)}_0)) & \text{if $z \in C^{(1)}_0 \cup C^{(1)}_1$} \\
        -1 & \text{otherwise} \\
        \end{array}
  \right.  
\end{equation}
since $z \in C^{(1)}_0 \setminus C^{(1)}_1$ if and only if $\rho(z,C^{(1)}_0) < 1$. Let $\rho(z,x):=\rho(z,\mathfrak S(x))$ be the dissimilarity between $x$ and $z$, then the dissimilarity measure $\rho(\cdot,\cdot)$ violates the symmetry axiom of the metric since $\rho(x,z) \neq \rho(z,x)$ whenever $d(x,t(x)) \neq d(z,t(z))$ where proximity regions $\N_{PE}(x,r)$ and $\N_{PE}(z,r)$ intersect with the lines $L(M_C(x),z)$ and $L(M_C(z),x)$ at points $t(x)$ and $t(z)$, respectively.

\subsection{Classification Methods} \label{sec:methods_class}

Hybrid PE-PCD classifiers depend on both the PE-PCD pre-classifier $g_P$ and the alternative classifier $g_C$. Therefore, we use some of the well known classification methods in the literature to incorporate them as alternative classifiers. All these classifiers are well defined for all points in $\R^d$, so we use them when the PE-PCD pre-classifier fails to make a decision, i.e. $g_P(z)=-1$. In addition to considering these classifiers as alternative classifiers, we apply them to the entire training data set in our simulated and real data studies to compare them with our hybrid classifiers as well. We provide definitions to these classifiers for data sets with two classes, $\X_0$ and $\X_1$. 

One such classifier is \emph{$k$-nearest neighbor} classifier which is perhaps one of the oldest. The decision/classification rule is simple: among the $k$ closest points to point $z$, classify $z$ as the class of the majority class of the points among $k$ neighbors:
\begin{equation}
g_{knn}(z):=I\left(\frac{\sum_{i=1}^k I(x_{(i)} \in \X_1)}{k} > 0.5 \right).
\end{equation}
\noindent Here, the points $x_{(1)},x_{(2)},\ldots,x_{(k)} \in \X_0 \cup \X_1$ are the $k$ closest points to the $z$. The accuracy of the method has shown to converge to the Bayes optimal as $k \rightarrow \infty$ and $k/n \rightarrow 0$ \citep{fix1989}. Moreover, when $k=1$, it can be shown that error of $k$-NN classifier (i.e. the \emph{nearest neighbor}  classifier) becomes less then or equal to the 2 times of Bayes optimal error \citep{cover1967}. 

For many classification tasks, linear classifiers are often preferred over others. \emph{Support vector machines} (SVM) are one of the most commonly used linear classifiers in the machine learning community due to their well understood theory and high accuracy \citep{vapnik1995}. Let $\X_0$ and $\X_1$ be two sets in $\R^d$ such that there exists a hyperplane with the normal vector $\mathbf{a}$, namely a \emph{separating hyperplane}, where $\mathbf{a}^Tx<0$ if $x \in \X_1$ and $\mathbf{a}^Tx>0$ if $x \in \X_0$. Thus, a linear classifier is constructed of the form 
\begin{equation}
	g_L(z) := I(\mathbf{a}^Tz < 0) 
\end{equation}
\noindent However, there are infinitely many such separating hyperplanes, and most importantly, not all pairs of classes in $\R^d$ are \emph{linearly seperable}. Here, linear separability implies the existence of a separating hyperplane. SVM classifiers incorporate kernel functions $\phi(\cdot)$ that map points in $\R^d$ to higher dimensions where separating hyperplanes exist. Among such hyperplanes, there exists one with the normal vector $\mathbf{a}$ such that this hyperplane has the maximum \emph{margin} (the minimum distance between the training data $\X_0 \cup \X_1$ and the hyperplane) among all possible hyperplanes, and the \emph{support vectors} $x_{(1)},x_{(2)},\ldots,x_{(m)} \in \X_0 \cup \X_1$ are the points closest to this hyperplane. Thus, a SVM classifier is of the form 
\begin{equation}
	g_{svm}(z) := I\left(\sum_{i=1}^m a_{(i)} \phi(z,x_{(i)})) - b < 0\right). 
\end{equation}
Here, $a_{(i)}$ is the element of the normal vector $\mathbf{a}$ corresponding to the support vector $x_{(i)}$. 

CCCD classifiers are also well defined for all points in $\R^d$. Elements of the dominating set $S_j$ are the selected prototypes of the target class $\X_j$. The prototype set $S_j$ is provided by Algorithm~\ref{dom_greedy} using CCCD $D_j$. Hence, given the sets $S_0$ and $S_1$, the classifier is defined as 
\begin{equation}
	g_{cccd}(z) := I\left( \min_{s \in S_1} \frac{d(z,s)}{\varepsilon_{\theta}(s)} < \min_{s \in S_0} \frac{d(z,s)}{\varepsilon_{\theta}(s)} \right).
\end{equation}
\noindent Here, $\varepsilon_{\theta}(s)$ is the radii of the ball $\N_S(s,\theta)=B(s,\varepsilon_{\theta}(s))$ associated with the point $s \in S_j$. 

	Multi-class adaptation of these classifiers is straightforward. In $k$-NN classifier, the point $z$ is labeled as the label of majority class among $k$ neighbors given the class labels $j=1,2,\ldots,J$. For SVM and CCCD classifiers, either ``one-against-all" or ``one-against-one" schemes can be adapted; that is, in first, one class remains the same where the remaining are gathered into one (the scheme used in PCD and CCCD classifiers). In the latter, however, the classifier is trained $J(J-1)/2$ times for each pair of classes \citep{hsu2002}. 

\subsection{Hybrid PE-PCD Classifiers}

\begin{figure}[!h]
\centering
\begin{tabular}{ccc}
\includegraphics[scale=0.32]{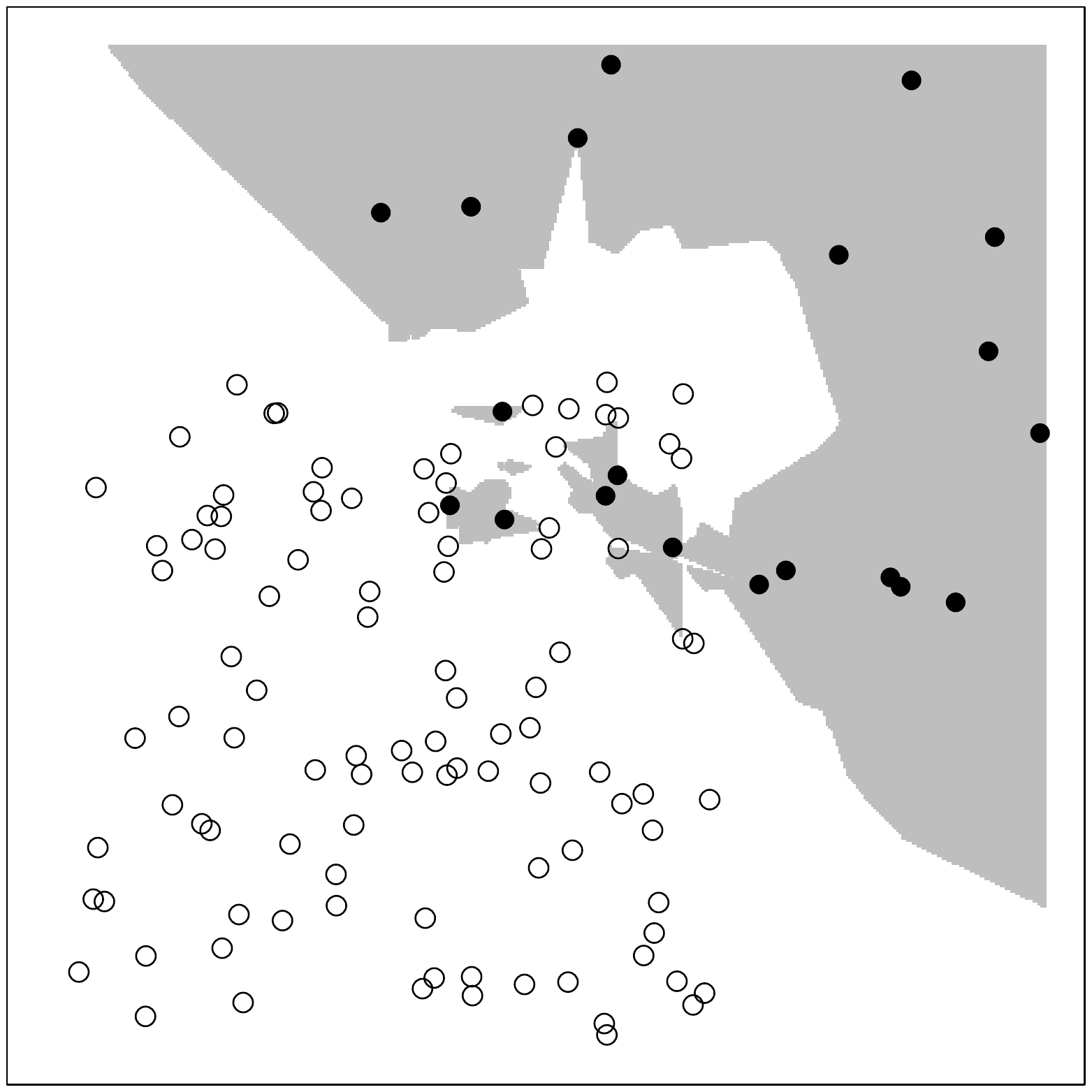} & \includegraphics[scale=0.32]{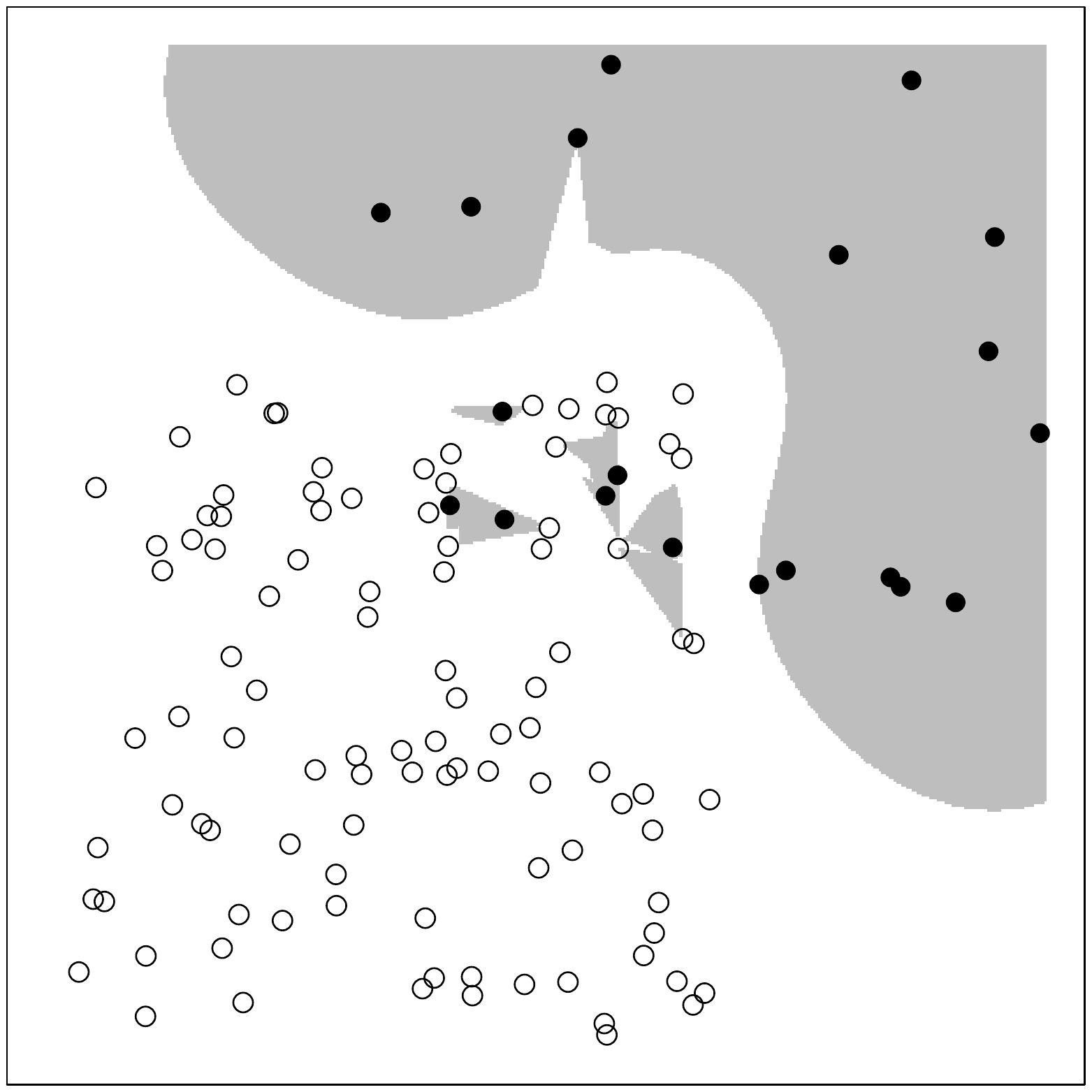} & \includegraphics[scale=0.32]{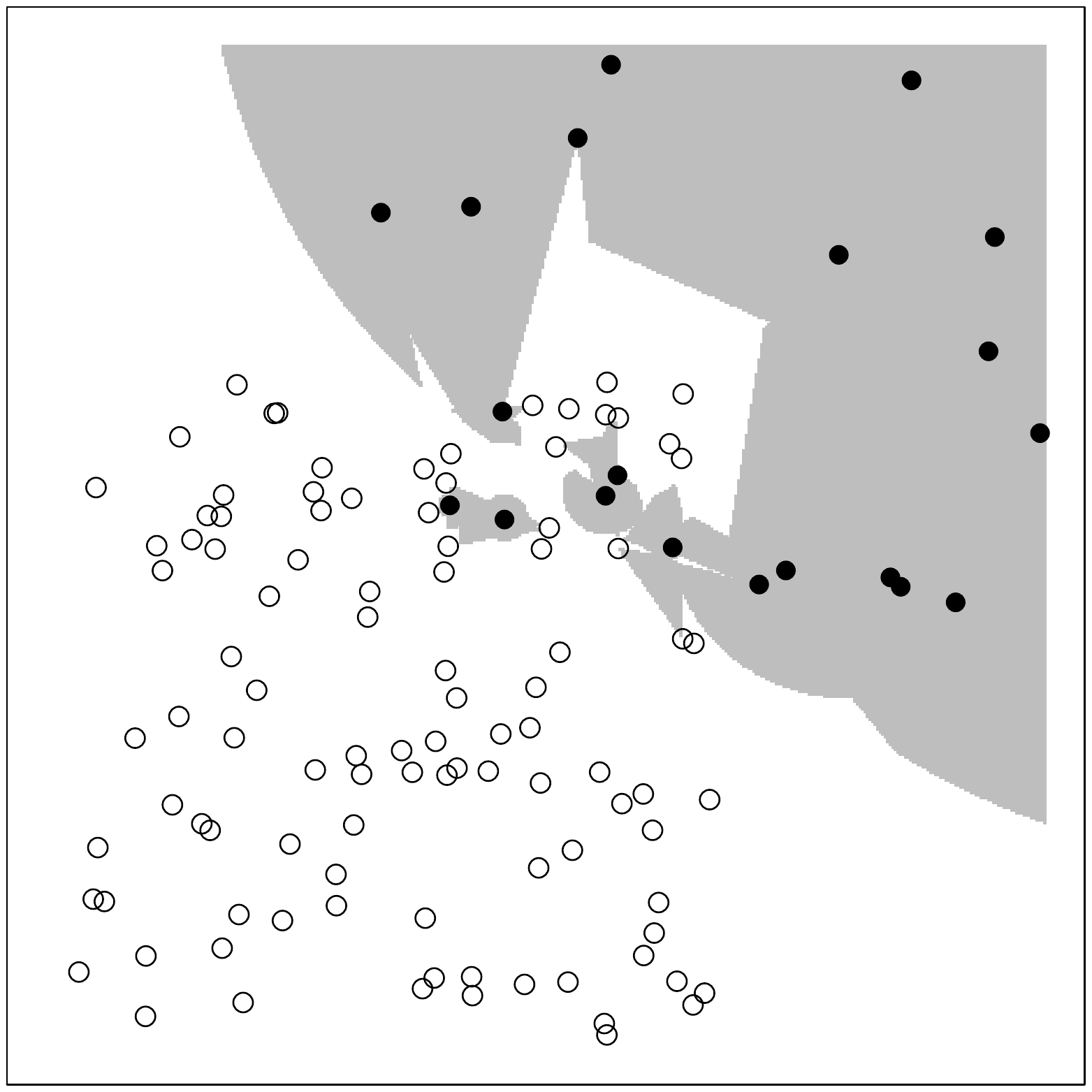} \\
(a) & (b) & (c) \\
\end{tabular}
\caption{The discriminant regions of hybrid PE-PCD classifiers with $r=2$ in a two-class setting with alternative classifier (a) $g_A=g_{knn}$ for $k=3$, (b) $g_A=g_{svm}$ for $\gamma=1$ (c) and $g_A=g_{cccd}$ for $\theta=0.5$. The grey region represents the regions where points classified as 1, i.e. $g_H(x)=1$. The training data set is composed of two classes $\X_0$ and $\X_1$ where in 100 and 20 samples are drawn from multivariate uniform distributions $U([0,1]^2)$ and $U([0.5,1.5]^2)$, respectively.}
\label{fig:discrim_hybrid}
\end{figure}

Constructing hybrid classifiers has many purposes. Some classifiers are designed to solve harder classification problems by gathering many weak learning methods (often known as \emph{ensemble} classifiers) while some others have advantages only when combined with another single classifier \citep{Wosniak20143}. Our hybrid classifiers are of the latter type. The PE-PCD pre-classifier $g_P$ is able to classify points in the overlapping region of the class supports, i.e. $s(F_0) \cap s(F_1)$, however classifying the remaining points in $\R^d$ requires incorporating an alternative classifier, often one that works for all points $\R^d$. We use the PE-PCD pre-classifier $g_P(\cdot)$ to classify all points of the test data, and if no decision are made for some of these points, we classify them with the alternative classifier $g_A$. Hence, let $g_H$ be the hybrid PE-PCD classifier such that 
\begin{equation}
g_H(z):= \left\{ \begin{array}{ll}
        g_P(z) & \text{if $z \in C^{(1)}_0 \cup C^{(1)}_1$} \\
        g_A(z) & \text{otherwise}. \\
        \end{array}
  \right.  
\end{equation}
\noindent For ``no decision" cases where $g_P(z)=-1$, we rely on the alternative classifier $g_A$; we will use the $k$-nearest neighbor, SVM and CCCD classifiers as alternative classifiers. The parameters are $k$, the number of closest neighbors to make a majority vote in the $k$-NN classifier; $\gamma$, the scaling parameter of the radial basis function (RBF) kernel of the SVM classifier; and $\theta$, the parameter of the CCCD classifier that regulates the size of each ball as described in Section~\ref{sec:cccds}. In Figure~\ref{fig:discrim_hybrid}, we illustrate the discriminant regions of three hybrid PE-PCD classifiers with expansion parameter $r=2$ where alternative classifiers are $g_A \in \{g_{knn},g_{svm},g_{cccd}\}$. The training data set is composed of two classes $\X_0$ and $\X_1$ where in 100 and 20 samples are drawn from multivariate uniform distributions $U([0,1]^2)$ and $U([0.5,1.5]^2)$, respectively.

\subsection{Composite and Standard Cover PE-PCD Classifers}

\begin{figure}[!h]
\centering
\begin{tabular}{ccc}
\includegraphics[scale=0.32]{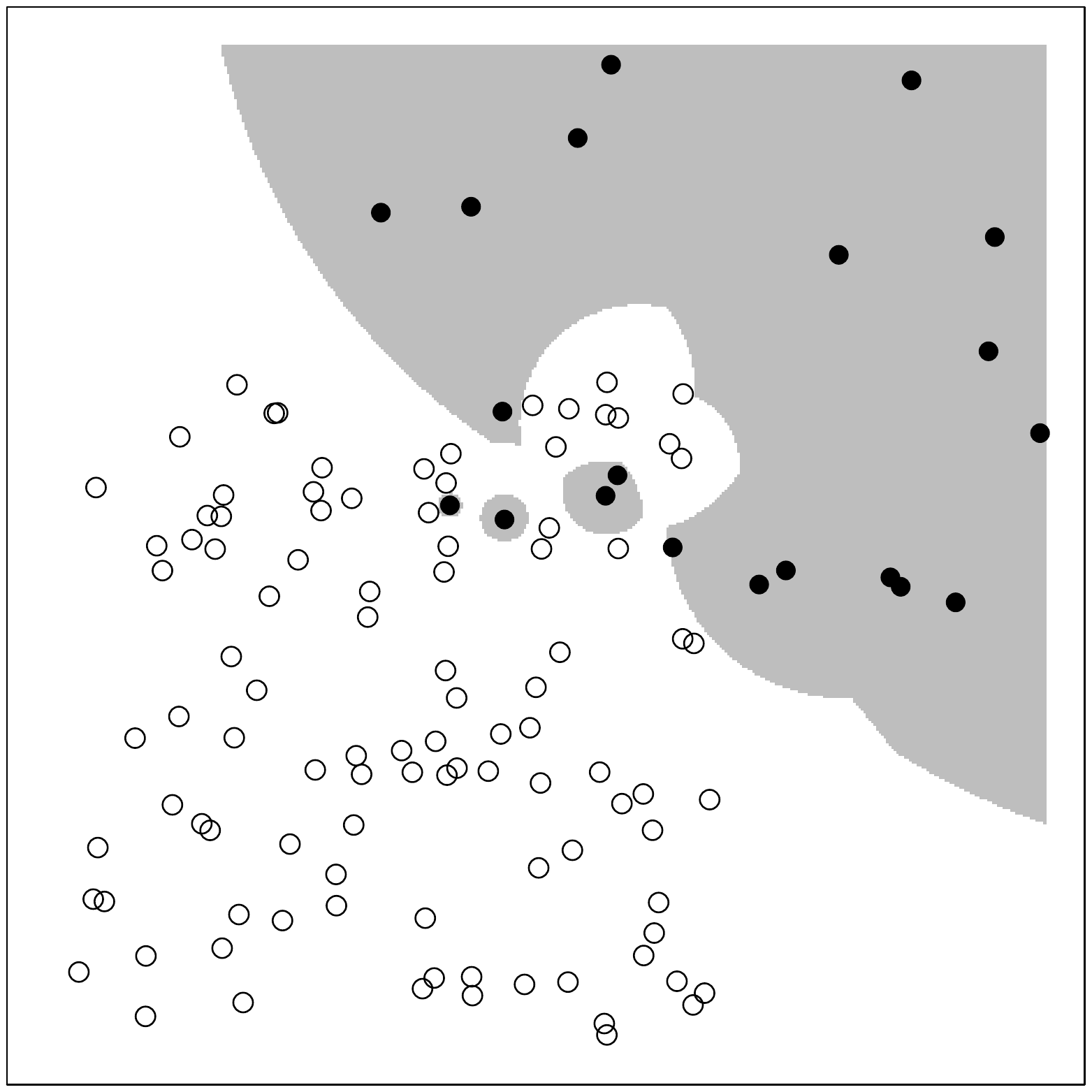} & \includegraphics[scale=0.32]{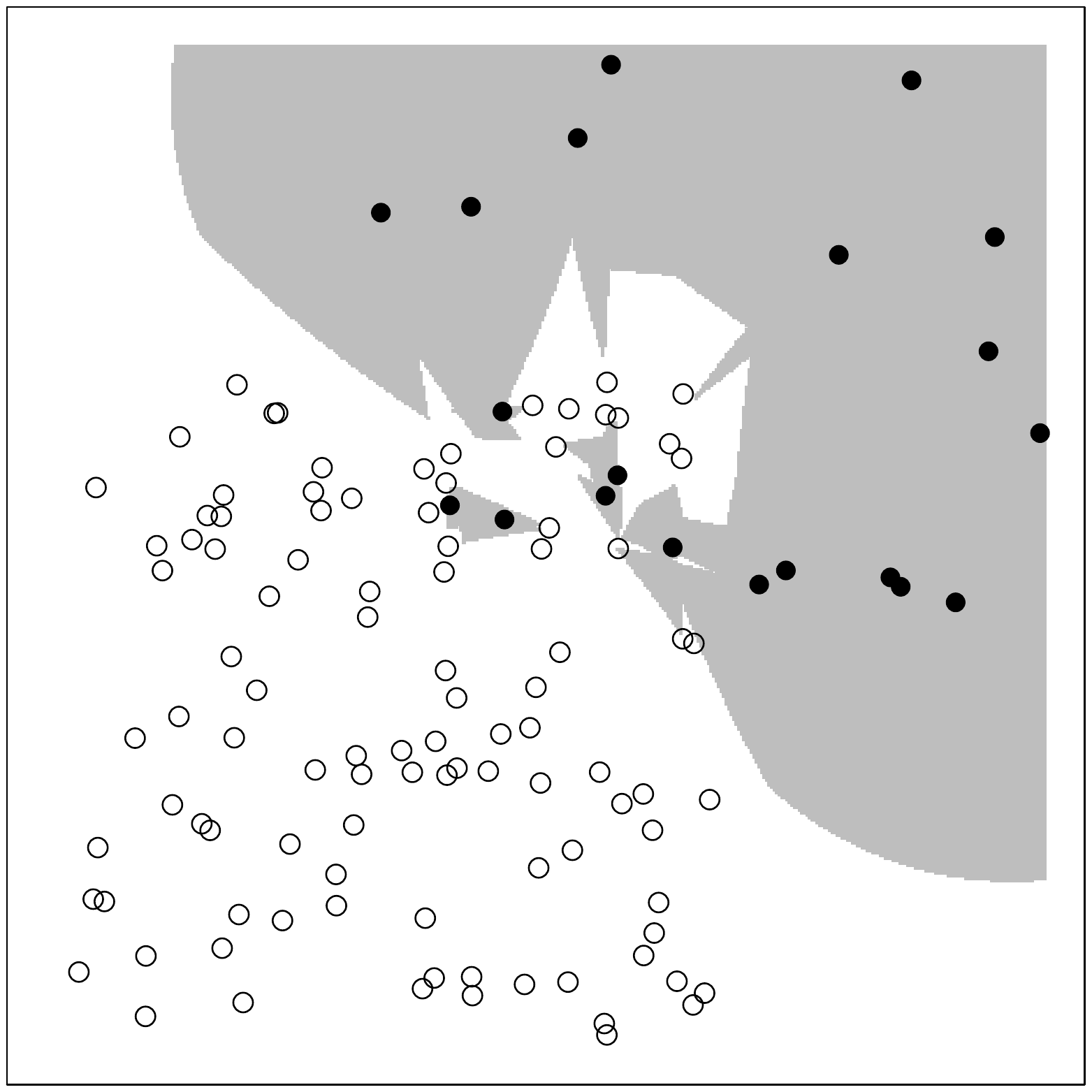} &\includegraphics[scale=0.32]{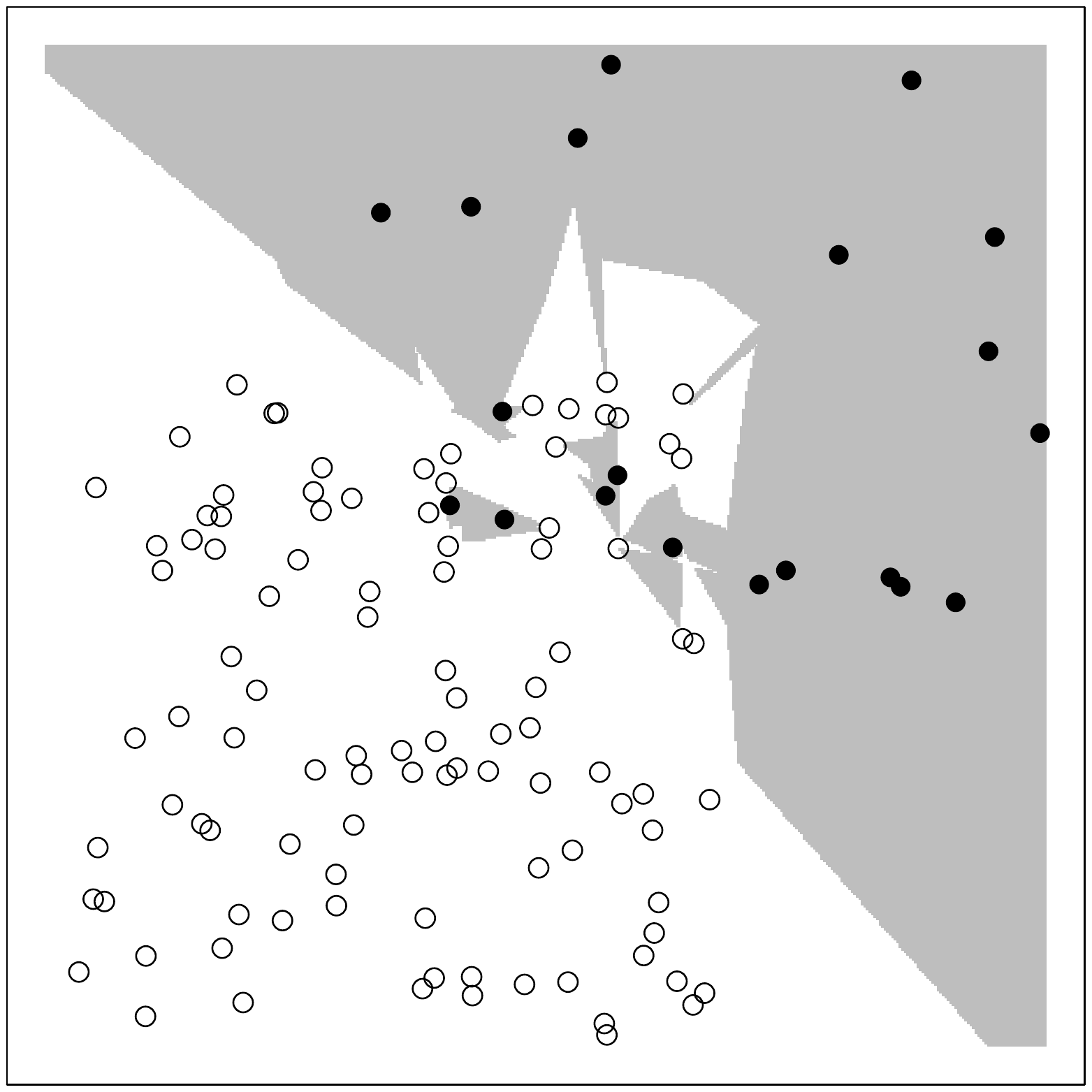} \\
\end{tabular}
\caption{The discriminant regions of cover PE-PCD classifiers. The grey region represents the regions where points classified as 1, i.e. $g_C(x)=1$. The training data set is composed of two classes $\X_0$ and $\X_1$ where in 100 and 20 samples are drawn from uniform distributions $U([0,1]^2)$ and $U([0.5,1.5]^2)$, respectively (a) The discriminant region of standard cover PE-PCD classifiers with the maps $\N_I(\cdot)=\N_O(\cdot)=\N_{S}(\cdot,\theta=1)$ (b) The discriminant region of the composite cover PE-PCD classifiers with maps $\N_I(\cdot)=\N_{PE}(\cdot,r=2)$ and $\N_O(\cdot)=\N_{S}(\cdot,\theta=1)$ (c)  The discriminant region the standard cover PE-PCD classifiers with maps $\N_I(\cdot)=\N_O(\cdot)=\N_{PE}(\cdot,r=2)$}
\label{fig:discrim_cover}
\end{figure}

We propose PE-PCD classifiers $g_C$ based on composite and standard covers. The classifier $g_C$ is defined as
\begin{equation}
g_C(z):= I(\rho(z,C_1) < \rho(z,C_0)).
\end{equation}
\noindent The cover is based on either composite covers or standard covers wherein both $\X_j \subset C_j$, hence a decision can be made without an alternative classifier. Note that composite cover PE-PCD classifiers are, in fact, different types of hybrid classifiers where the classifiers are only modelled by class covers but with multiple types of PCDs. Compared to hybrid PE-PCD classifiers, cover PE-PCD classifiers have many appealing properties. Since a reduction is done over all target class points $\X_j$, depending on the percentage of reduction, classifying a new point $z \in \R^d$ is computationally faster and more efficient, whereas an alternative classifier might not provide such a reduction. We provide the discriminant regions of cover PE-PCD classifiers with standard covers of maps $\N_{PE}(\cdot,r)$ and $\N_S(\cdot,\theta)$ used separately, and with composite covers with maps $\N_{PE}(\cdot,r)$ and $\N_S(\cdot,\theta)$ used jointly in Figure~\ref{fig:discrim_cover}.

Note that, given the multi-class prototype sets, $S_j$, the two-class cover PE-PCD classifier $g_C$ can be modified for the multi-class case as
\begin{equation}
g(z)= \underset{j \in J}{\argmin} \left( \min_{s \in S_j} \rho(z,N(s)) \right).
\end{equation}

\subsection{Consistency Analysis}

In this section, we prove some results on the consistency of both hybrid PE-PCD classifiers and cover PE-PCD classifiers when two class conditional distributions are strictly $\delta$-seperable. For $\delta \in [0,\infty)$, the regions  $A,B \subset \R^d$ are $\delta$-\emph{separable} if and only if $$\inf_{x\in A,y\in B} d(x,y) \geq \delta.$$ Moreover, let $\delta$-separable regions $A$ and $B$ be the supports of continuous distributions $F_A$ and $F_B$, respectively. Hence, $F_A$ and $F_B$ are called $\delta$-\emph{separable distributions}, and if $\delta > 0$, \emph{strictly} $\delta$-separable \citep{devroye:1996}.

We first show the consistency of cover PE-PCD classifiers, and then, we show that the hybrid PE-PCDs classifiers are also consistent. Cover classifiers are characterized by the PCDs associated with proximity regions $\N(x)$ for $x \in \R^d$, and thus, the consistency of such PCD classifiers depend on the map $\N(\cdot)$. We require the following properties for a proximity map $\N(\cdot)$ to satisfy: 
\begin{enumerate}
\item[\textbf{P1}] For all $x \in \R^d$, the proximity region $\N(x)$ is an open set, and $x$ is in the interior of $\N(x)$. 
\item[\textbf{P2}] Given data sets from two classes $\X_0$ and $\X_1$ with distributions $F_0$ and $F_1$, and supports
$s(F_0)$ and $s(F_1)$, and given that $x \in s(F_j)$ for $j=0,1$, the proximity map $\N(\cdot)$ associated with the target class $\X_j$ is a function on the non-target class points such that $\N(x) \cap \X_{1-j} = \emptyset$. 
\end{enumerate}

Note that both $\N_S(\cdot,\theta)$ for $\theta \in (0,1]$ and $\N_{PE}(\cdot,r)$ for $r \in (1,\infty)$ satisfy \textbf{P1} and \textbf{P2}. These will be useful in showing that classifiers based on our class covers attain Bayes-optimal classification performance for $\delta$-separable classes. Thus, first, we have to show that the support of a class is almost surely a subset of the class cover for sufficiently large data sets. Note that all points of the target class reside inside the class cover $C_j$, i.e. $\X_j \subset C_j$. Hence, we have the following proposition. 

\begin{proposition} \label{prop:consist}
Let $\Z_n=\{Z_1,Z_2,\ldots,Z_n\}$ be a set of i.i.d. random variables drawn from a continuous distribution $F$ whose support is $s(F) \subseteq \R^d$. Let the proximity map $\N(\cdot)$ satisfy $\textbf{P1}$, let the corresponding class cover of $\Z_n$ be denoted as $C(\Z_n)$ such that $\Z_n \subset C(\Z_n)$, and let $C^* := \liminf_{n \rightarrow \infty}  C(\Z_n)$. Hence, we have $s(F) \subset C^*$ w.p. 1 in the sense that $\lambda(s(F) \setminus C^*) \rightarrow 0$ almost surely where $\lambda(\cdot)$ is the Lebesgue measure functional. 
\end{proposition}

\noindent {\bf Proof:} Suppose, for a contradiction, $s(F) \not\subset C^*$ w.p. 1. Hence, $s(F) \setminus C^* \neq \emptyset$ w.p. $1$ in such a way that $\lambda(s(F) \setminus C^*) > 0$ w.p. $1$ since $\lambda(\N(Z))> 0$ for all $Z \in s(F)$ by \textbf{P1}. Hence, $\Z_n \cap (s(F) \setminus C^*) \neq \emptyset$ w.p. 1 as $n \rightarrow \infty$, but then some $Z \in \Z_n \cap (s(F) \setminus C^*)$  will not be in $C^*$, which contradicts the fact that $C^*$ covers $\Z_n$ including $Z$. $\blacksquare$

Proposition~\ref{prop:consist} shows that a class cover almost surely covers the support of its associated class. However, to show consistency of classifiers based on PCD class covers, we have to investigate the class covers under the assumption of separability of class supports.  

Let $\X_0$ and $\X_1$ be two classes of a data set with strictly $\delta$-separable distributions, the property \textbf{P2} of the map $\N(\cdot)$ establishes \emph{pure} class covers that include none of the points of the non-target class, i.e. $C_j \cap \X_{1-j} = \emptyset$. In this case, we have the following proposition showing that the intersection of the cover of the target class and the support of the non-target class is almost surely empty as $n_{1-j} \rightarrow \infty$. 

\begin{proposition} \label{prop:pure_consist}
 Let $\X_0=\{X_1,X_2,\ldots,X_{n_0}\}$ and $\X_1=\{Y_1,Y_2,\ldots,Y_{n_1}\}$ be two sets of i.i.d. random variables with strictly $\delta$-separable continuous distributions $F_0$ and $F_1$. For $j=0,1$, let the proximity map $\N(\cdot)$ satisfy \textbf{P1} and \textbf{P2} such that the map $\N(\cdot)$ of the target class is a function on the non-target class $\X_{1-j}$. Then, for $j=0,1$, we have $C(\X_j) \cap s(F_{1-j})=  \emptyset$ almost surely as $n_{1-j} \rightarrow \infty$ in the sense that $\lambda(C(\X_j) \setminus s(F_{1-j})) \rightarrow 0$ as $n_{1-j} \rightarrow \infty$. 
\end{proposition} 

\noindent {\bf Proof:} For $j=0,1$, note $C(\X_j)=\cup_{X \in S_j} \N(X)$ for $S_j \subset \X_j$ being the minimum prototype set of $\X_j$. We prove the proposition for $j=0$ (as the proof of case $j=1$ follow by symmetry). Hence, it is sufficient to show that (given $\N(\cdot)$ is a function on $\X_1$) $\N(x) \cap s(F_1) =  \emptyset$ w.p. $1$ as $n_1=|\X_1| \rightarrow \infty$ for all $x \in s(F_0)$. Suppose for a contradiction, $\lambda(C(\X_0) \setminus s(F_{1})) > 0$ w.p. 1 as $n_1 \rightarrow \infty$. Then, there exists $x \in s(F_0)$ such that $\N(x) \cap s(F_1) \neq \emptyset$ almost surely as $n_1 \rightarrow \infty$. Then, the region $\N(x) \cap s(F_1)$ has positive measure. Therefore, some $Y \in \X_1$ will fall in to this region w.p. 1 as $n_1 \rightarrow \infty$. This contradicts \textbf{P2} since $Y \in \N(x) \cap s(F_1)$ implies $\N(x) \cap \X_1 \neq \emptyset$. $\blacksquare$

Now, we would like to show that cover PE-PCD classifiers are consistent when class supports are strictly $\delta$-separable; that is, the error rate of the cover classifier $L(g_C)$ converges to the Bayes optimal error rate $L^*$, which is $0$ for classes with $\delta$-separable supports, as $n_0,n_1 \rightarrow \infty$ \citep{devroye:1996}. Then, we have the following theorem. 
   
\begin{theorem} \label{thm:consist}
Suppose that the samples of the data set $\X_0 \cup \X_1$ are i.i.d. with distribution $F=\pi_0\,F_0+(1-\pi_0)\,F_1$ for $\pi_0 \in [0,1]$, and let class conditional distributions $F_0$ and $F_1$ are continuous with supports $s(F_0)$ and $s(F_1)$, being finite dimensional and strictly $\delta$-separable. Then the cover classifier $g_C$ is consistent; that is, $L(g_c)\rightarrow L^*=0$ as $n_0,\,n_1 \rightarrow \infty$.
\end{theorem}

\noindent {\bf Proof:} Let $Z_j$ be a random variable with distribution $F_j$ for $j=0,1$. Then by Propositions~\ref{prop:consist} and~\ref{prop:pure_consist}, we have $P(Z_j \in C(\X_j)) \rightarrow 1$ as $n_j \rightarrow \infty$ and $P(Z_j \not\in C(\X_{1-j})) \rightarrow 1$ as $n_{1-j} \rightarrow \infty$. Hence, $$ P(Z_j \not\in C(\X_j) \text{ and } Z_j \in C(\X_{1-j})) \rightarrow 0$$ as $n_0,n_1 \rightarrow \infty$. Then, for $C_j=C(\X_j)$, 
\begin{align*}
 L(g_C) & = P(g_C(Z_0) \neq 0) \pi_0 + P(g_C(Z_1) \neq 1) \pi_1 \\
        & = P(Z_0 \not\in C_0 \text{ and } Z_0 \in C_{1}) \pi_0 + 
            P(Z_1 \not\in C_1 \text{ and } Z_1 \in C_{0}) \pi_1. \\ 
\end{align*}
\noindent Hence, $L(g_C) \rightarrow 0$ as $n_0,n_1 \rightarrow \infty$. $\blacksquare$ 

As a corollary to Theorem~\ref{thm:consist}, we have that classifier $g_C$ of standard and composite covers with maps $N_S(\cdot,\theta)$ and $N_{PE}(\cdot,r)$ for $r > 1$ are consistent. A special case occurs when $r=1$; that is, observe that $x \in \partial(\N(x))$, and hence $\N(\cdot)$ does not satisfy \textbf{P1}. 

We showed that a cover PE-PCD classifier is consistent provided that, as $n_0,n_1 \rightarrow \infty$, support of the target class is a subset of the class cover, and the PE-PCD cover excludes all points of the non-target class almost surely. However, to show that the hybrid PE-PCD classifiers are consistent, we need alternative classifiers which are consistent as well. 

\begin{theorem} \label{thm:consist_hybrid}
Suppose that the samples of data set $\X_0 \cup \X_1$ are i.i.d. with distribution $F=\pi_0\,F_0+(1-\pi_0)\,F_1$ for $\pi_0 \in [0,1]$, and let class conditional distributions $F_0$ and $F_1$ are continuous with supports $s(F_0)$ and $s(F_1)$, being finite dimensional and strictly $\delta$-separable. Then the hybrid classifier $g_H$ is consistent provided that alternative classifier $g_A$ is also consistent. 
\end{theorem}

\noindent {\bf Proof:} Note that $C_j=C_j^{(1)} \cup C_j^{(2)}$ and $C_j^{(1)} \subset C_H(\X_{1-j})$. For $j=0,1$, let $Z_j \sim F_j$. Also, let $\Upsilon_j$ be the event that $Z_j \in C_0^{(1)} \cup C_1^{(1)}$ and let $\upsilon_j := P(\Upsilon_j)$. Note that $$L(g_H) = P(g_H(Z_0) \neq 0) \pi_0 + P(g_H(Z_1) \neq 1) \pi_1. $$ Hence, for $j=0,1$;
\begin{align*}
P(g_H(Z_j) \neq j) &= P(g_H(Z_j) \neq j | \Upsilon_j) \upsilon_j + P(g_H(Z_j) \neq j | \Upsilon_j^c) (1-\upsilon_j) \\
                   &= P(g_P(Z_j) \neq j | \Upsilon_j) \upsilon_j + P(g_A(Z_j) \neq j | \Upsilon_j^c) (1-\upsilon_j).
\end{align*}
\noindent As $n_0,n_1 \rightarrow \infty$, $P(g_P(Z_j) \neq j | \Upsilon_j) \rightarrow 0$ by Theorem~\ref{thm:consist}, and $P(g_A(Z_j) \neq j) \rightarrow 0$ since the classifier $g_A$ is consistent. Then the result follows. $\blacksquare$

\section{Monte Carlo Simulations and Experiments} \label{sec:simulations}

In this section, we assess the classification performance of hybrid and cover PE-PCD classifiers. We perform simulation studies wherein observations of two classes are drawn from separate distributions where $\X_0$ is a random sample from a multivariate uniform distribution $U([0,1]^d)$ and $\X_1$ is from $U([\nu,1+\nu]^d)$ for $d = 2,3,5$ with the overlapping parameter $\nu \in [0,1]$. Here, $\nu$ determines the level of overlap between the two class supports. We regulate $\nu$ in such a way that the overlapping ratio $\zeta$ is fixed for all dimensions, i.e. $\zeta=\vol(s(F_0) \cap s(F_1))/\vol(s(F_0) \cup s(F_1))$. When $\zeta=0$, the supports are well separated, and when $\zeta=1$, the supports are identical: i.e. $s(F_0) = s(F_1)$. Hence, the closer the $\zeta$ to 1, the more the supports overlap. Observe that $\nu \in [0,1]$ can be expressed in terms of the overlapping ratio $\zeta$ and dimensionality $d$: 
\begin{equation} \label{equ:delta}
	\zeta = \frac{\vol(s(F_0) \cap s(F_1))}{\vol(s(F_0) \cup s(F_1))}=\frac{(1-\nu)^d}{2-(1-\nu)^d} \quad \Longleftrightarrow \quad \nu = 1-\left(\frac{2\zeta}{1+\zeta}\right)^{1/d}. 
\end{equation}

In this simulation study, we train the classifiers with $n_0=400$ and $n_1 = qn_0$ with the imbalance level $q=|\X_1|/|\X_0|=\{0.1,0.5,1.0\}$ and overlapping ratio $\zeta=0.5$. For values of $q$ closer to zero, classes of the data set are more imbalanced. On each replication, we form a test data with 100 random samples drawn from each of $F_0$ and $F_1$, resulting a test data set of size 200. This setting is similar to a setting used by \cite{manukyan2016}, who showed that CCCD classifiers are robust to imbalance in data sets. We intend to show that the same robustness extends to PE-PCD classifiers. Using all classifiers, at each replication, we record the area under curve (AUC) measures for the test data, and also, we record the correct classification rates (CCRs) of each class of the test data separately. We perform these replications until the standard errors of AUCs of all classifiers are below 0.0005. We refer to the CCRs of two classes as ``CCR0" and ``CCR1", respectively. We consider the expansion parameters $r=1,1.2,\ldots,2.9,3,5,7,9$ for the PE-PCD classifiers. Our hybrid PE-PCD classifiers are referred as PE-SVM, PE-$k$NN and PE-CCCD classifiers with alternative classifiers SVM, $k$-NN and CCCD, respectively.  

Before the main Monte Carlo simulation, we perform a preliminary (pilot) Monte Carlo simulation study to determine the values of optimum parameters of SVM, CCCD and $k$-NN classifiers. The same values will be used for alternative classifiers as well. We train the $g_{svm}$, $g_{cccd}$ and $g_{knn}$ classifiers, and classify the test data sets for each classifier to find the optimum parameters. We perform Monte Carlo replications until the standard error of all AUCs are below 0.0005 and record which parameter produced the maximum AUC among the set of all parameters in a trial. Specifically, on each replication, we (i) classify the test data set with each $\theta$ value (ii) record the $\theta$ values with maximum AUC and (iii) update the count of the recorded $\theta$ values. Finally, given a set of counts associated with each $\theta$ value, we appoint the $\theta$ with the maximum count as the $\theta^*$, the optimum $\theta$ (or the best performing $\theta$). Later, we use $\theta^*$ as the parameter of alternative classifier $g_{cccd}$ in our main simulations. Optimal parameter selection process is similar for classifiers $g_{knn}$ and $g_{svm}$ associated with the parameters $k$ and $\gamma$.

The optimum parameters of each simulation setting is listed in Table~\ref{tab:opt}. We consider parameters of SVM $\gamma=0.1,0.2, \ldots,4.0$, of CCCD $\theta=0,0.1,\ldots,1$ (here, $\theta=0$ is actually equivalent to $\theta=\epsilon$, the machine epsilon), and of $k$-NN $k=1,2,\ldots,30$. In Table~\ref{tab:opt}, as $q$ and $d$ increases, optimal parameters $\gamma$ and $\theta$ decrease whereas $k$ increases. \cite{manukyan2016} showed that dimensionality $d$ may affect the imbalance between classes when the supports overlap. Observe that in Table~\ref{tab:opt}, with increasing $d$, optimal parameters are more sensitive to the changes in imbalance level $q$. For the CCCD classifier, $\theta=1$ is usually preferred when the data set is imbalanced, i.e. $q=0.1$ or $q=0.5$. Bigger values of $\theta$ are better for the classification of imbalanced data sets, since with $\theta=1$, the cover of the minority class is substantially bigger which increases the domain influence of the points of the minority class. For $\theta$ closer to $0$, the class cover of the minority class is much smaller compared the class cover of the majority class, and hence, the CCR1 is much smaller. Bigger values of parameter $k$ is also detrimental for imbalanced data sets, the bigger the parameter $k$, the more likely a new point is classified as class of the majority class since the points tend to be labelled as the class of the majority of $k$ neighboring points. As for the parameter $\gamma$, support vectors have more influence over the domain as $\gamma$ decreases \citep{wang2003}. Note that $\gamma=1/(2\sigma^2)$ in the radial basis function (RBF) kernel. The smaller the $\gamma$, the bigger the $\sigma$. Hence more points are classified as the majority class with decreasing $\gamma$ since the majority class has more influence. Thus, bigger values of $\gamma$ is better for the imbalanced data sets. 

\begin{table}[ht]
\centering
\caption{Optimum parameters for SVM, CCCD and $k$-NN classifiers used in the hybrid PE-PCD classifiers. }
\begin{tabular}{ccccc}
$d$ & $q$ & $\theta$ (CCCD) & $k$ ($k$-NN) & $\gamma$ (SVM) \\
\hline
\multirow{3}{*}{2} & 0.1 &1 &1 &3.8 \\
& 0.5 &1 &1 &4.0 \\
& 1.0 &0 &3 &0.1 \\
\hline
\multirow{3}{*}{3} & 0.1 &1 &1 &2.3 \\
& 0.5 &1 &1 &0.4 \\
& 1.0 &0 &4 &0.2 \\
\hline
\multirow{3}{*}{5} & 0.1 &1 &1 &0.9 \\
& 0.5 &1 &4 &0.3 \\
& 1.0 &1 &10 &0.1 \\
\hline
\end{tabular}
\label{tab:opt}
\end{table}

\begin{figure}
\centering
\includegraphics[scale=0.63]{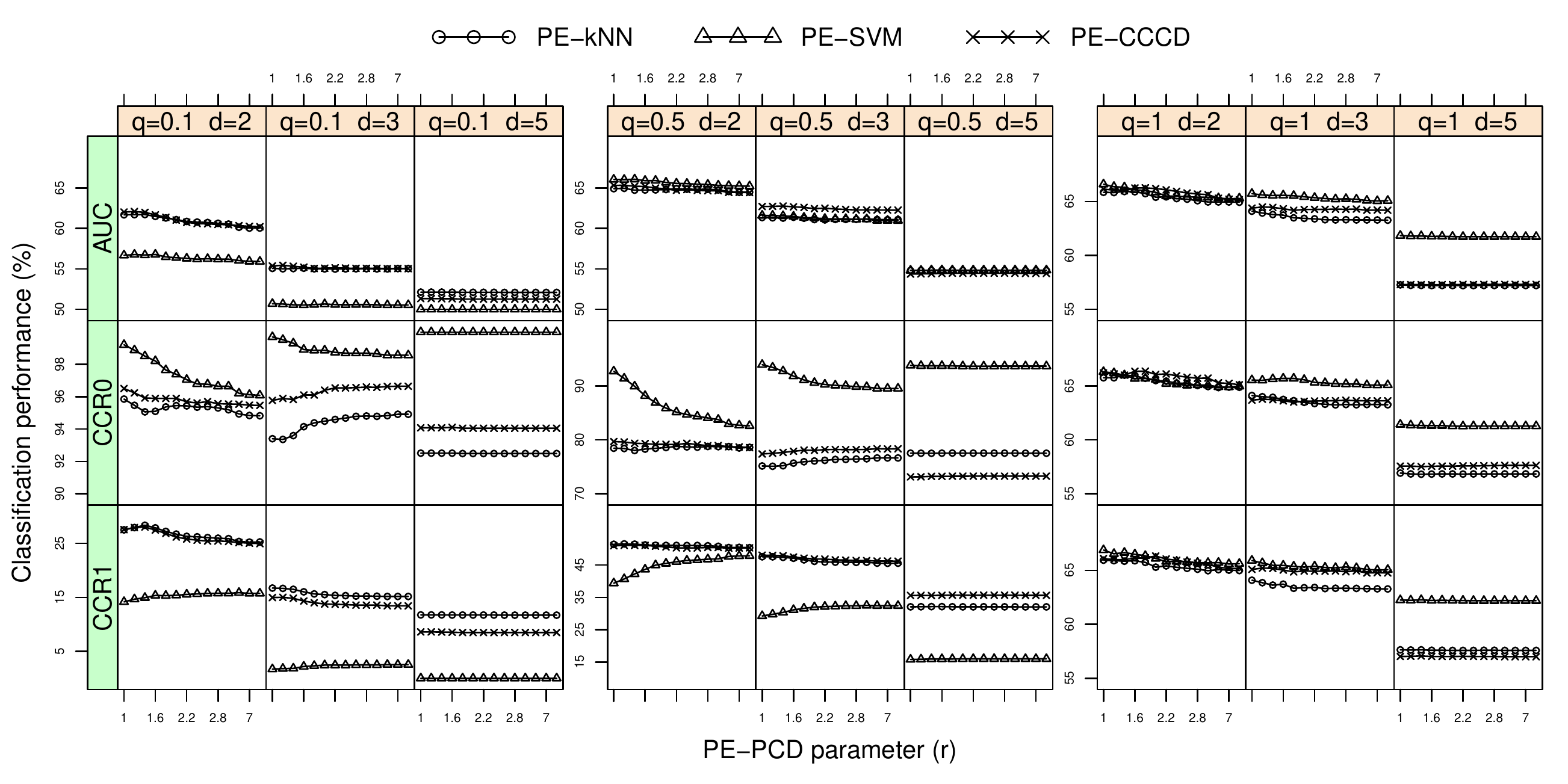}
\caption{AUCs and CCRs of the three hybrid PE-PCD classifiers versus expansion parameter $r=1,1.2,\ldots,2.9,3,5,7,9$ and the alternative classifiers: CCCD, $k$-NN and SVM. Here, the classes are drawn as $\X_0 \sim U([0,1]^d)$ and $\X_1 \sim U([\nu,1+\nu]^d)$ with several simulation settings based on $\zeta=0.5$ given the Equation~\ref{equ:delta}, imbalance level $q=0.1,0.5,1$, and dimensionality $d=2,3,5$.}
\label{fig:shift_hybrid}
\end{figure}

Average of AUCs and CCRs of three hybrid PE-PCD classifiers are presented in Figure~\ref{fig:shift_hybrid}. For $q=0.1$, the classifier PE-$k$NN, for $q=0.5$, the classifier PE-CCCD and, for $q=1.0$, the classifier PE-SVM performs better than others. Especially, when the data set is imbalanced, the CCR1 determines the performance of a classifier; that is, generally, the better a method classifies the minority class, the better the method performs overall. When the data is balanced (i.e. $q=1$), PE-SVM is expected to perform well, however it is known that SVM classifiers are confounded by the imbalanced data sets \citep{akbani2004}. Moreover, when $q=0.1$, PE-$k$NN performs better than PE-CCCD. This result contradicts the results of \cite{manukyan2016}. The reason for this is hybrid PE-PCD classifiers incorporate alternative classifiers for points outside of the convex hull and $k$NN might perform better for these points. The $k$NN classifier is prone to missclassify points closer to the decision boundary when the data is imbalanced, and we expect points outside the convex hull to be far away from the decision boundary in our simulation setting.  

In Figure~\ref{fig:shift_hybrid}, CCR1 increases while CCR0 decreases for some settings of $q$ and $d$, and vice versa for some other settings. Recall that Theorem~\ref{thm:rparam} shows a stochastic ordering of the expansion parameter $r$; that is, with increasing $r$, there is an increase in the probability of exact MDS being less than or equal to some $\kappa=1,\ldots,d+1$. Hence with increasing $r$, the proximity region $\N_{PE}(x,r)$ gets bigger and the cardinality of the prototype set $S_j$ gets lower. Therefore, we achieve a bigger cover of the minority class and more reduction in the majority class. The bigger the cover, the higher the CCR1 is in the imbalanced data sets. However, the decrease in the performance, when $r$ increases, may suggest that alternative classifiers perform better for these settings. For example, the CCR1 of PE-SVM increases as $r$ increases for $q=0.1,0.5$ and $d=2,3$, but CCR1 of PE-CCCD and PE-$k$NN decreases for $r \geq 1.6$. The higher the $r$, the more the reduction in data set. However, higher values of $r$ may confound the classification performance. Hence, we choose an optimum value of $r$. Observe that for $d=5$, the AUCs of all hybrid PE-PCD classifiers are equal for all $r$. With increasing dimensionality, the probability that a point of the target class falling in the convex hull of the non-target class decreases, hence most points remain outside of the convex hull. 

\begin{figure}
\centering
\includegraphics[scale=0.63]{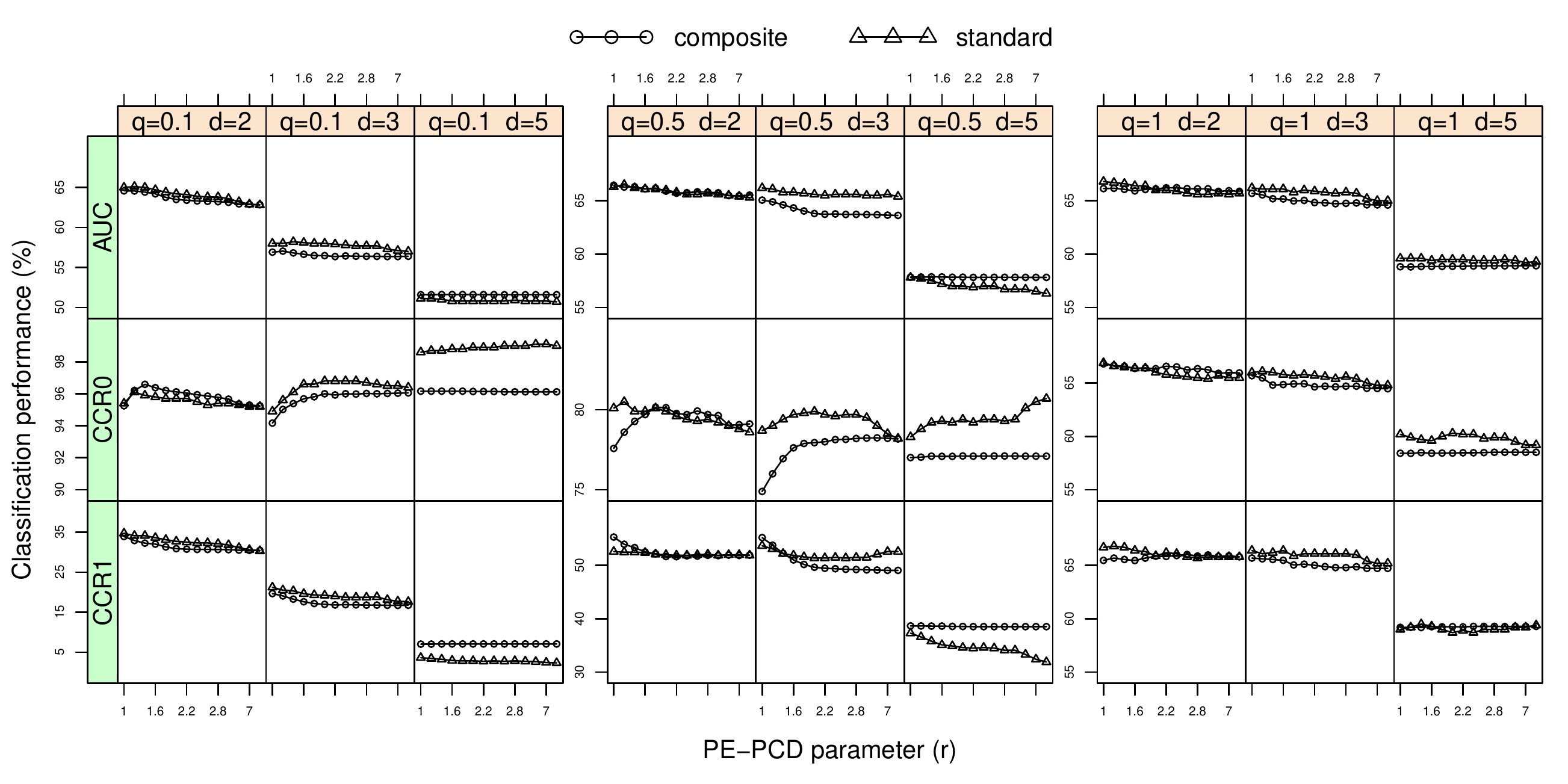}
\caption{AUCs and CCRs of the two cover PE-PCD classifiers versus expansion parameter $r=1,1.2,\ldots,2.9,3,5,7,9$ with composite and standard covers. Here, the classes are drawn as $\X_0 \sim U([0,1]^d)$ and $\X_1 \sim U([\nu,1+\nu]^d)$ with several simulation settings based on $\zeta=0.5$ given the Equation~\ref{equ:delta}, imbalance level $q=0.1,0.5,1$, and dimensionality $d=2,3,5$.}
\label{fig:shift_cover}
\end{figure}

In Figure~\ref{fig:shift_cover}, we compare the composite cover PE-PCD classifier and the standard cover PE-PCD classifier. The standard cover is slightly better in classifying the minority class, especially when there is imbalance between classes. In general, the standard cover PE-PCD classifier appear to have more CCR1 than the composite cover PE-PCD classifiers. However, the composite covers are better when $d=5$. The PE-PCD class covers are surely influenced by the increasing dimensionality. Moreover, for $q=0.1,0.5$, we see that the CCR1 of standard cover PE-PCD classifier slightly decreases with $r$, even though the data set is more reduced with increasing $r$. Hence, we should choose an optimum value of $r$ that can still be incorporated to both substantially reduce the data set and to achieve a good classification performance.  

\begin{figure}
\centering
\includegraphics[scale=0.63]{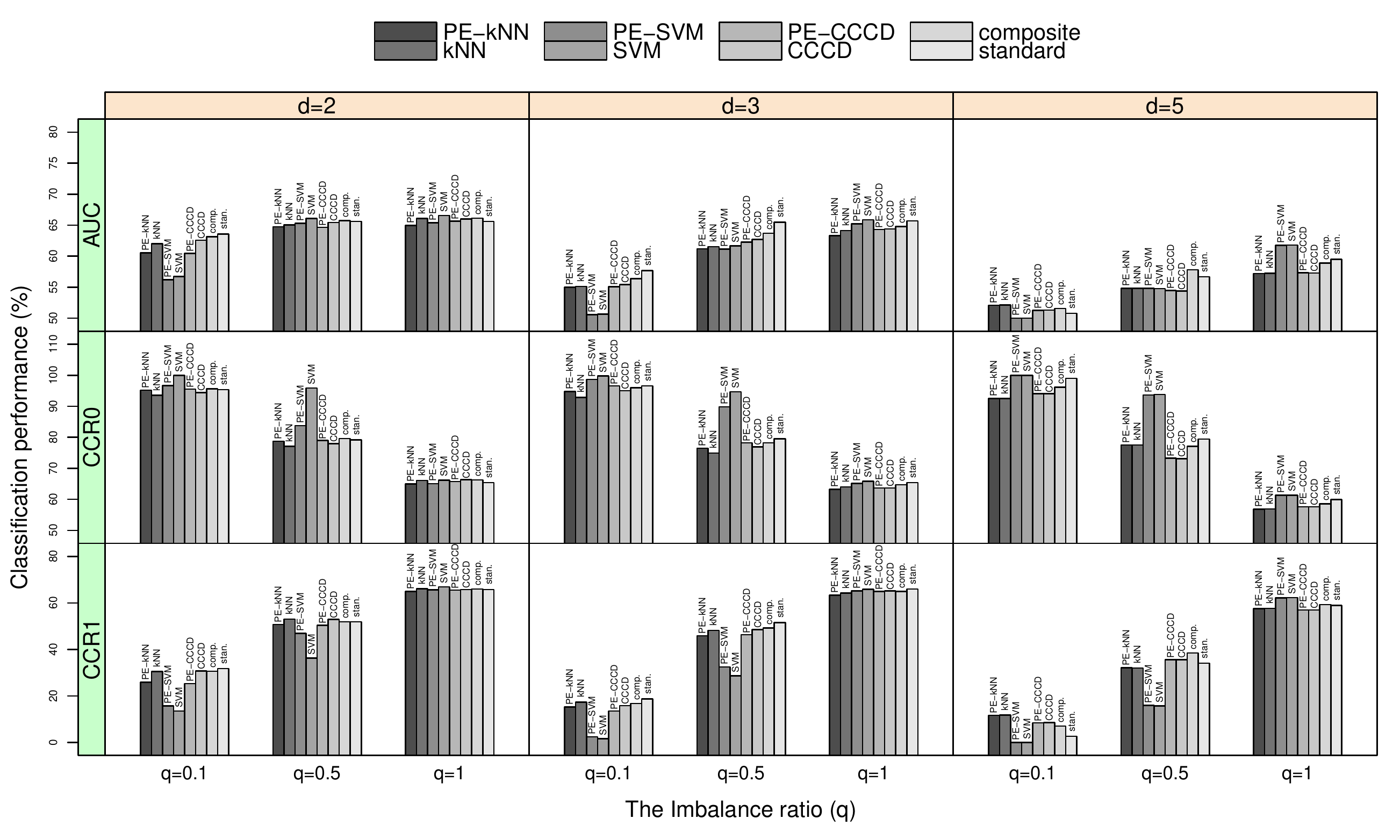}
\caption{AUCs and CCRs of the two cover, three hybrid PE-PCD classifiers with expansion parameter $r=3$, and $k$-NN, SVM and CCCD classifiers. The composite covers are indicated with ``comp." and standard covers with ``stan.". Here, the classes are drawn as $\X_0 \sim U([0,1]^d)$ and $\X_1 \sim U([\nu,1+\nu]^d)$ with several simulation settings based on $\zeta=0.5$, imbalance level $q=0.1,0.5,1$ and dimensionality $d=2,3,5$.}
\label{fig:shift_all}
\end{figure}

In Figure~\ref{fig:shift_all}, we compare all five classifiers, three hybrid and two cover PE-PCD classifiers. We consider the expansion parameter $r=3$ since, in both Figures~\ref{fig:shift_hybrid} and~\ref{fig:shift_cover}, class covers with $r=3$ perform well and, at the same time, substantially reduce the data set. For all $d=2,3,5$, it appears that all classifiers show comparable performance when $q=1$, but PE-SVM and SVM give slightly better results. However, when there is imbalance in the data sets, the performances of PE-SVM and SVM degrade, and hybrid and cover PE-PCD classifiers and CCCD classifiers have more AUC values than others. Compared to all other classifiers, on the other hand, the standard cover PE-PCD classifier is clearly the best performing one for $d=2,3$ and $q=0.1,0.5$. Observe that the standard cover PE-PCD classifier achieves the highest CCR1 among all classifiers. Apparently, the standard cover constitutes the most robust (to class imbalance) classifier. The performance of standard cover PE-PCD classifier is usually comparable to the composite cover PE-PCD classifier, but slightly better. However, for $d=5$, the performance of standard cover PE-PCD classifier degrades and composite cover PE-PCD classifiers usually perform better. These results show that cover PE-PCD classifiers are more appealing than hybrid PE-PCD classifiers. The reason for this is that the cover PE-PCD classifiers have both good classification performance and reduce the data considerably more since hybrid PE-PCD classifiers provide a data reduction for only $\X_j \cap C_H(\X_{1-j})$ whereas cover PE-PCD classifiers reduce the entire data set. The level of reduction, however, may decrease as the dimensionality of the data set increases. 

\begin{figure}
\centering
\includegraphics[scale=0.63]{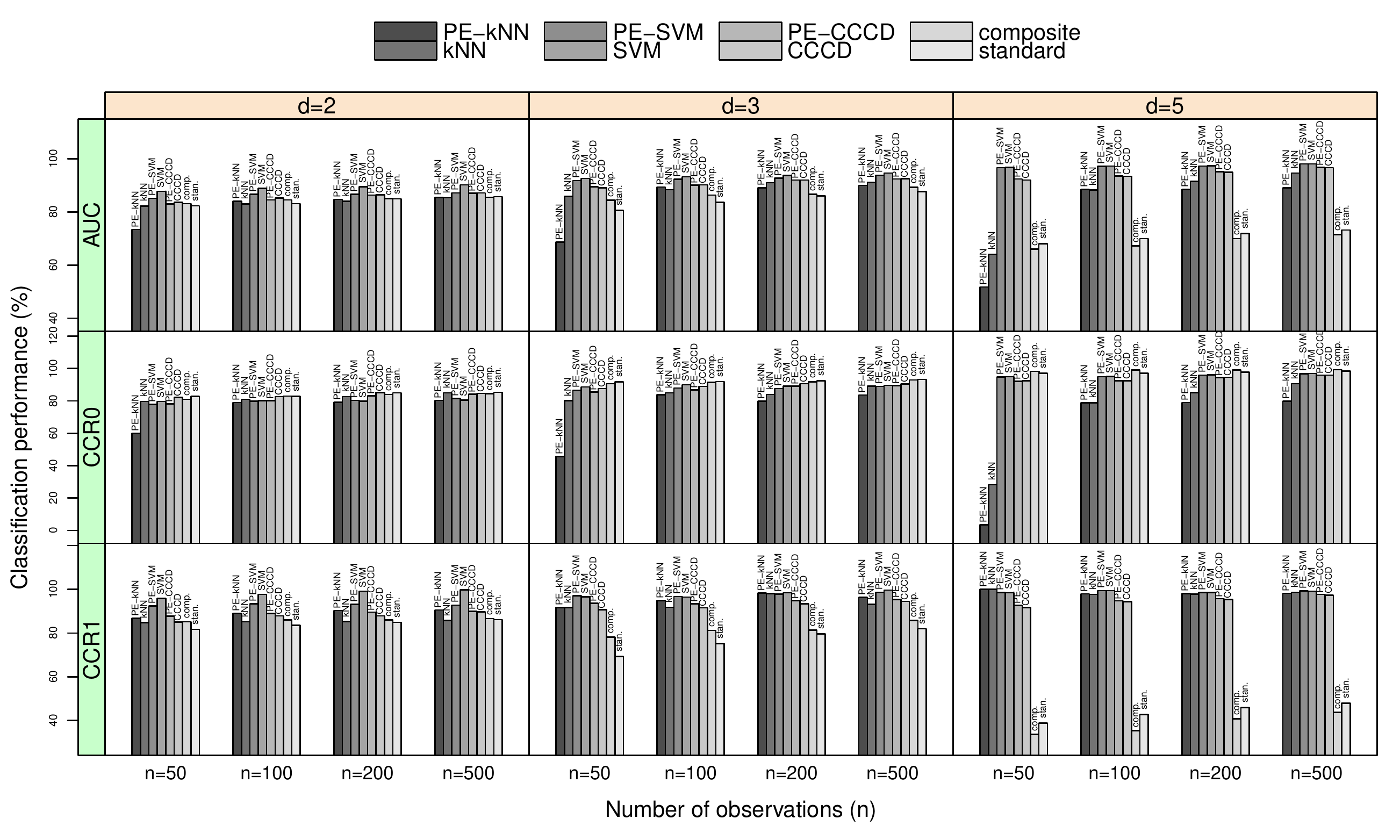}
\caption{AUCs and CCRs of the two cover, three hybrid PE-PCD classifiers with expansion parameter $r=2.2$, and $k$-NN, SVM and CCCD classifiers. The composite covers are indicated with ``comp." and standard covers with ``stan.". Here, the classes are drawn as $\X_0 \sim U([0,1]^d)$ and $\X_1 \sim U([0.3,0.7]^d)$ with several simulation settings based on number of observarions $n=50,100,200,500$ and dimensionality $d=2,3,5$.}
\label{fig:embedded_all}
\end{figure}

In Figure~\ref{fig:embedded_all}, we compare all five classifiers, three hybrid and two cover PE-PCD classifiers in a slightly different simulation setting where there exists an inherent class imbalance. We perform simulation studies wherein equal number of observations $n=n_0=n_1$ are drawn from separate distributions where $\X_0$ is a random sample from a multivariate uniform distribution $U([0,1]^d)$ and $\X_1$ is from $U([0.3,0.7]^d)$ for $d = 2,3,5$ and $n=50,100,200,500$. Observe that the support of one class in entirely inside of the other, i.e. $s(F_1) \subset s(F_1)$. The same simulation setting have been used to highlight the robustness of CCCD classifiers to imbalanced data sets \citep{manukyan2016}. In Figure~\ref{fig:embedded_all}, the performance of $k$NN and PE-$k$NN classifiers degrade as $d$ increases and $n$ decreases. With sufficiently high $d$ and low $n$, the minority class $\X_0$ is sparsely distributed around the overlapping region of class supports $s(F_1) \cap s(F_0)$ which is the support of $\X_1$. Hence, although the number of observations are equal in both classes, there exists a ``local" imbalance between classses \citep{manukyan2016}. However, CCCD and SVM classifiers, including the associated hybrid PE-PCD classifiers perform fairly good. Although the cover PE-PCD classifiers have considerably less CCR1, they perform relatively good compared to other classifiers and generally have more CCR0 than other classifiers. Similar to other simulation settings, cover PE-PCD classifiers are also affected by the increasing dimensionality of this data set.  

\begin{figure}
\centering
\includegraphics[scale=0.70]{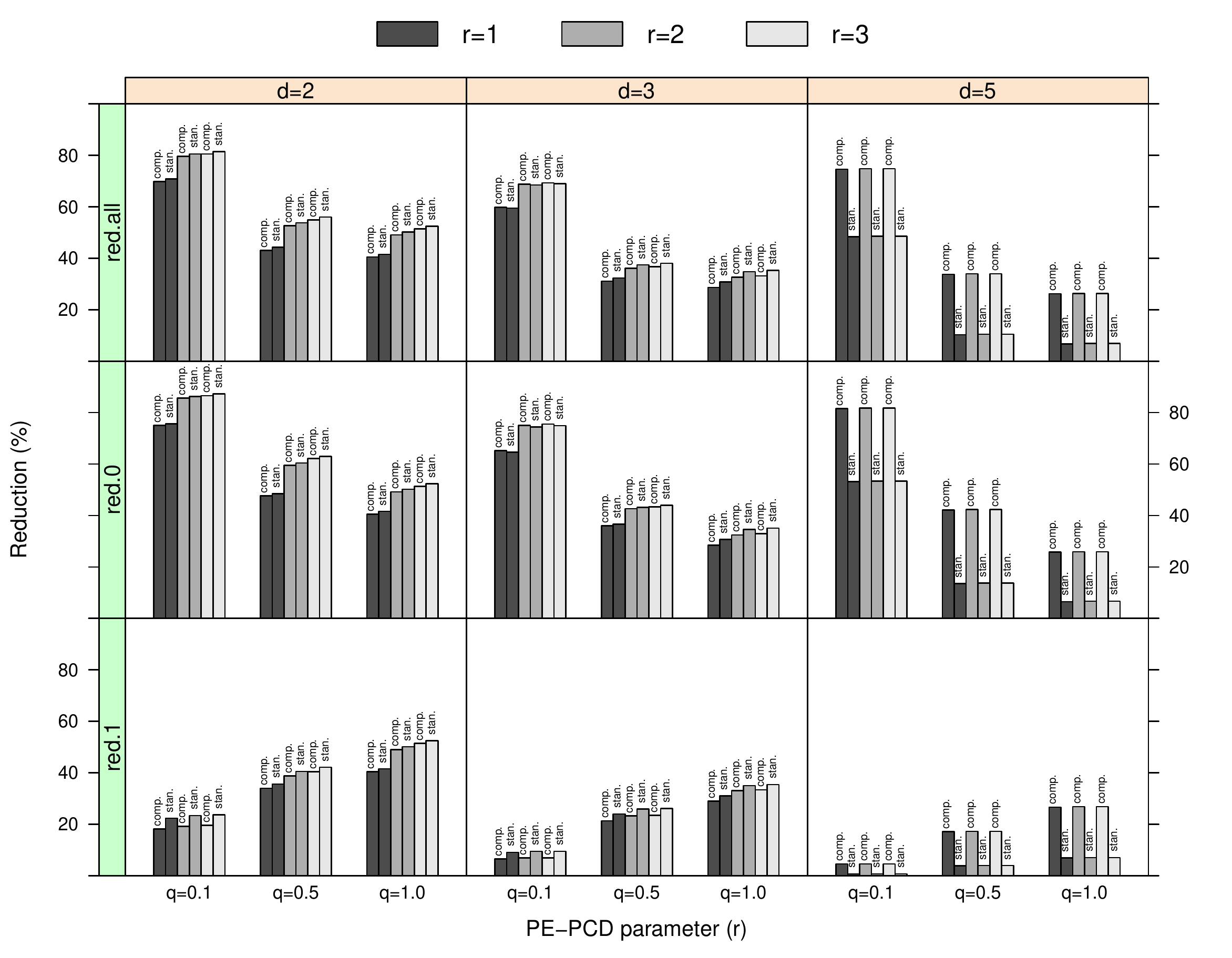}
\caption{The percentage of reduction of the composite (comp.) and standard (stan.) PE-PCD covers. The ``red.all" indicates the overall reduction in the training data set, $1-(|S_0+S_1|/(n_0+n_1))$, ``red.0" the reduction in the $\X_0$ class, $1-(|S_0|/n_0)$, and ``red.1" the reduction in the $\X_1$ class, $1-(|S_1|/n_1)$. Here, the classes are drawn as $\X_0 \sim U([0,1]^d)$ and $\X_1 \sim U([\nu,1+\nu]^d)$ with several simulation settings based on $\zeta=0.5$, imbalance level $q=0.1,0.5,1$ and dimensionality $d=2,3,5$. }
\label{fig:shift_red}
\end{figure}

Although the PE-PCD based standard cover classifiers are competitive in classification performance, a case should be made on how much they reduce the data sets during the training phase. In Figure~\ref{fig:shift_red}, we illustrate the percentage of reduction in the training data set, and separately, in both minority and majority classes, using PE-PCD for $r=1,2,3$. The overall reduction increases with $r$, which is also indicated by Theorem~\ref{thm:rparam}, and the reduction in the majority class is much more than in minority class when $q=0.1,0.5$ since proximity regions of the majority class catch more points unlike the minority class. The majority class is reduced over nearly $\%60$ when $q=0.1$, and $\%40$ when $q=0.5$. Indeed, the more the imbalance between classes, the more the reduction in the abundantly populated classes. On the other hand, as the dimensionality increases, composite covers reduce the data set more than the standard covers. The number of the facets and simplices increases exponentially with $d$, and hence the cardinality of minimum dominating set (or the prototype set) also increases exponentially with $d$ (see Theorem~\ref{thm:complexity}). As a result, composite PE-PCD covers achieve much more reduction than standard PE-PCD covers. 

\clearpage

\clearpage

\section{Real Data Examples} \label{sec:realdata}

\begin{figure}[!h]
\centering
\begin{tabular}{cc}
\includegraphics[scale=0.4]{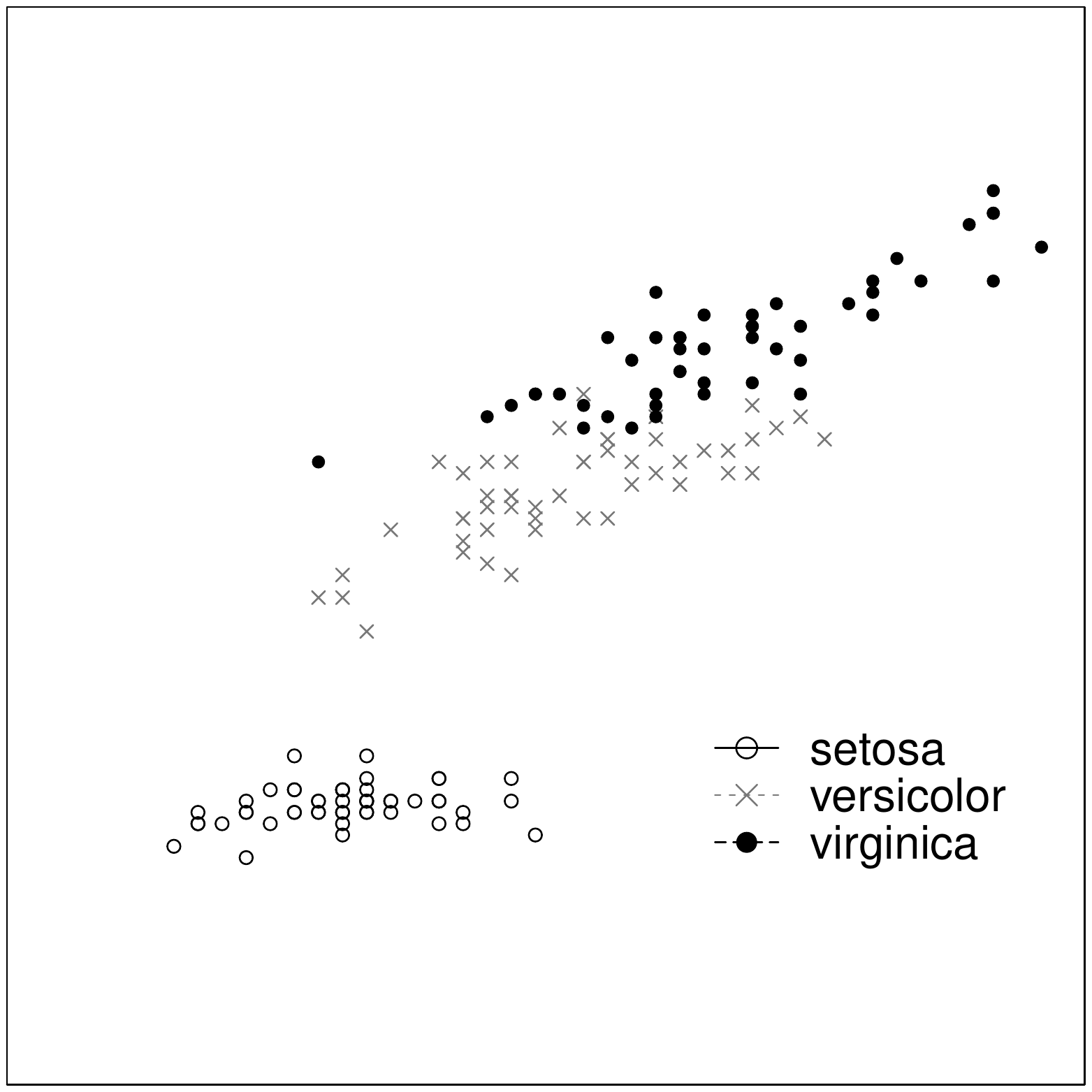} & \includegraphics[scale=0.4]{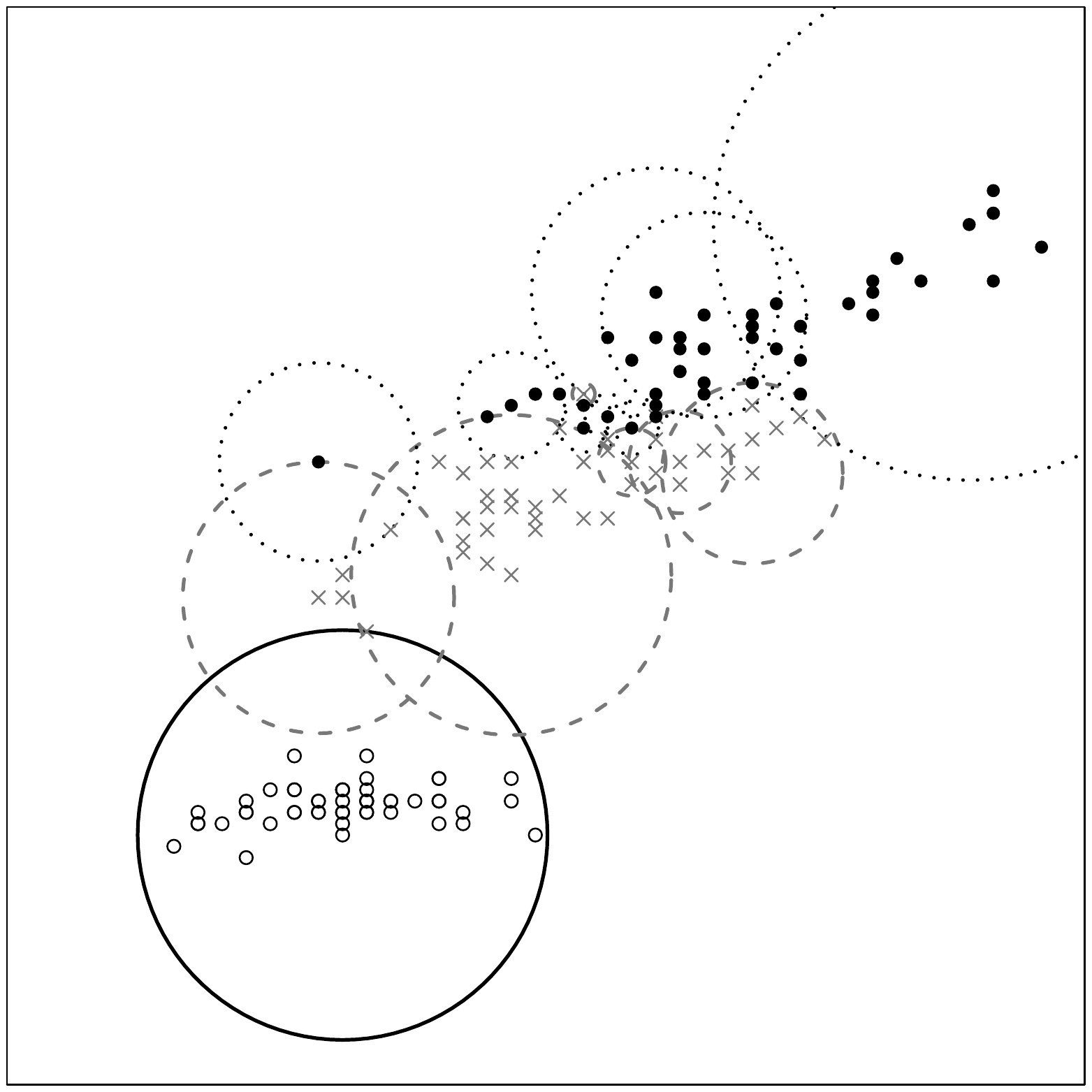} \\ 
(a) & (b) \\
\includegraphics[scale=0.4]{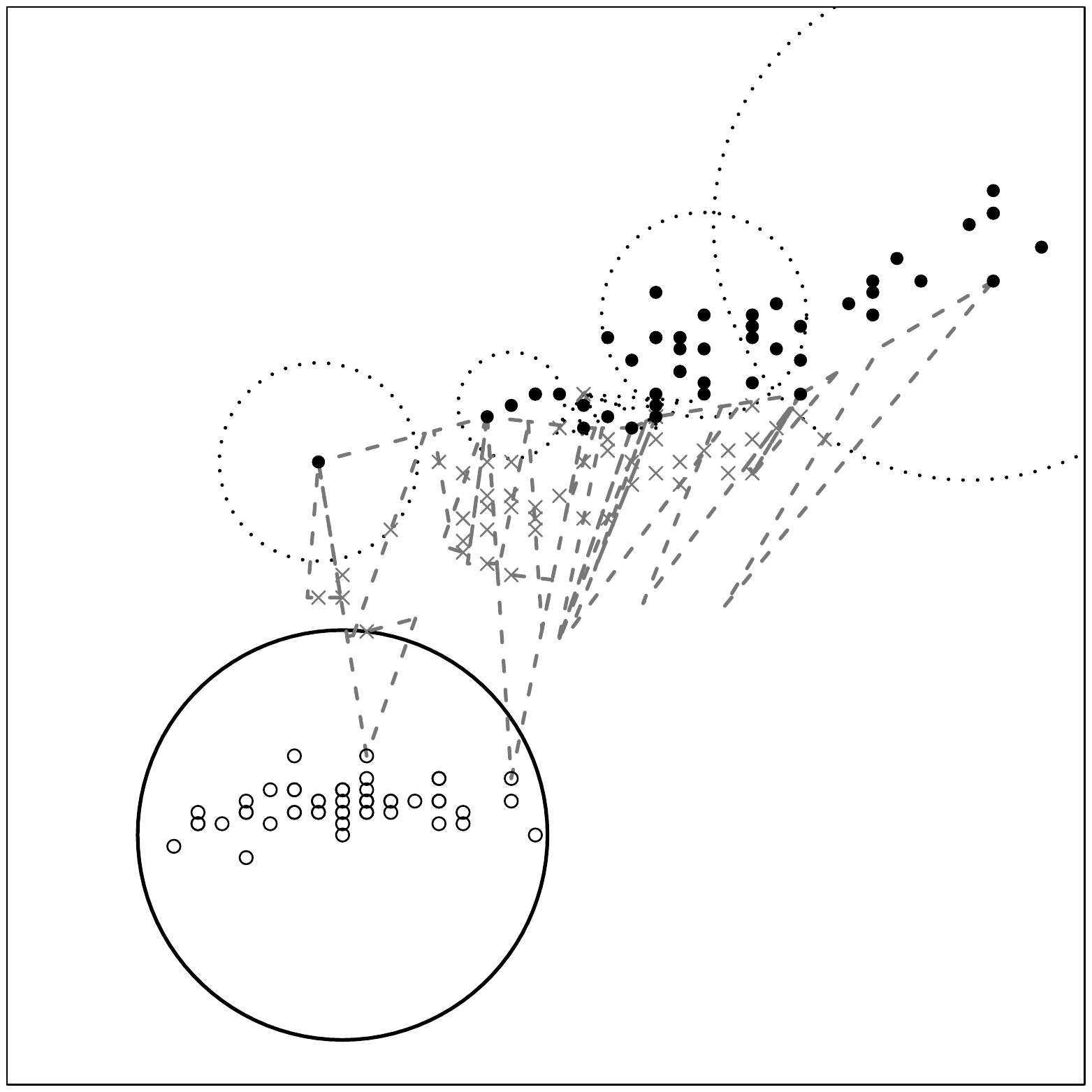} & \includegraphics[scale=0.4]{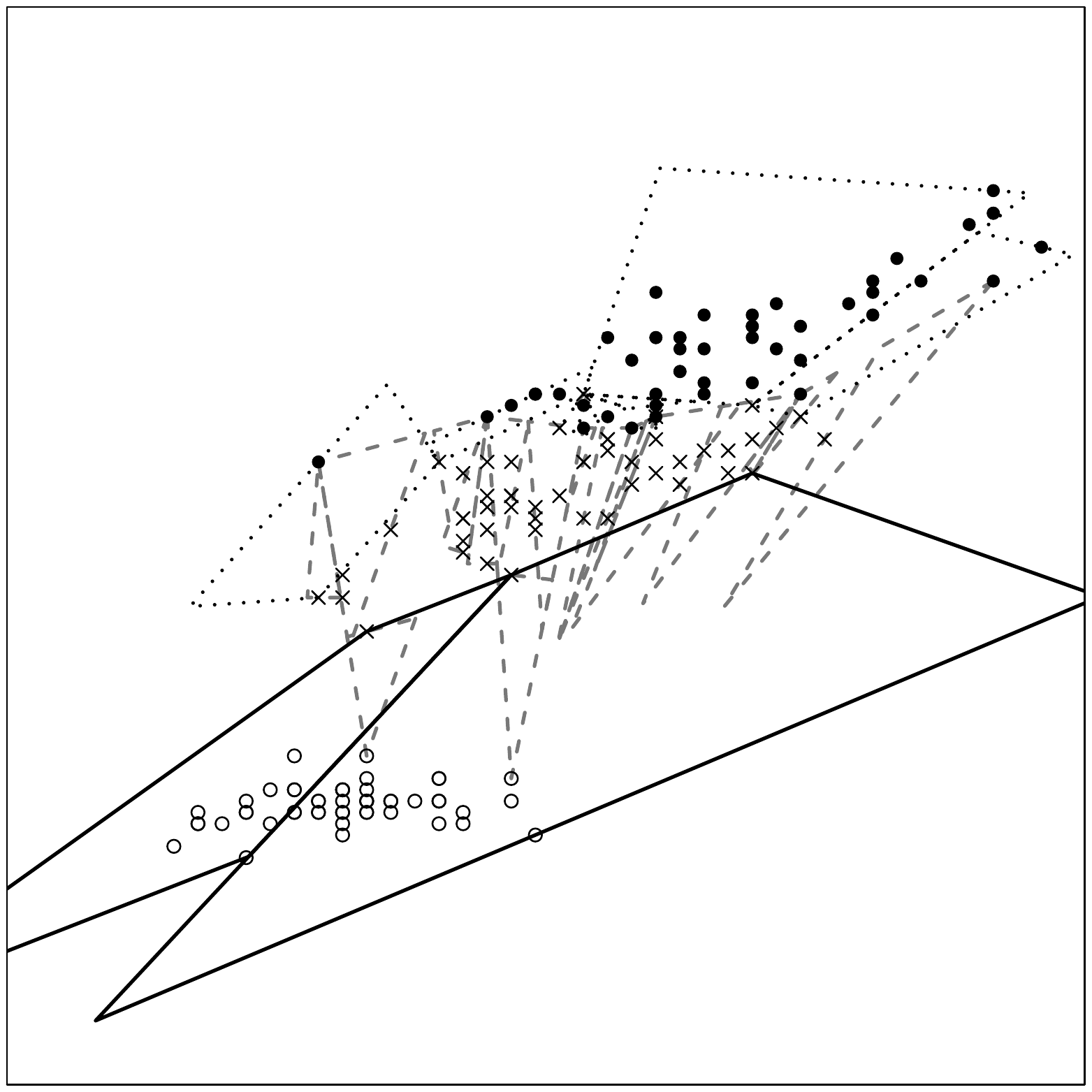} \\
(c) & (d) \\
\end{tabular}
\caption{Class covers of \texttt{iris13} data set. (a) The data set with variables sepal and petal length. (b) Standard covers with $\N_{S}(\cdot,\theta=1)$, (c) composite covers with $\N_I(\cdot)=\N_{PE}(\cdot,r=1)$ and $\N_O(\cdot)=\N_{S}(\cdot,\theta=1)$ and (d) standard covers with $\N_I(\cdot)=\N_O(\cdot)=\N_{PE}(\cdot,r=1)$.}
\label{fig:iris_covers}
\end{figure}

In this section, we apply the hybrid and cover PE-PCD classifiers on UCI and KEEL data sets \citep{BacheLichman,fdez2011}. We start with a trivial but a popular data set, \texttt{iris}. This data set is composed of 150 flowers classified into three types based on their petal and sepal lengths. Hence it constitutes a nice example for class covers of multi-class data sets. In Figure~\ref{fig:iris_covers}, we illustrate standard and composite of PE-PCD covers, and CCCD covers of the first and the third variables of iris data set, sepal and petal lengths. We refer to this data set as \texttt{iris13}. Observe that in composite covers of Figure~\ref{fig:iris_covers}(c), only a few or no triangles are used to cover the setosa and virginica classes. Points of these classes are almost all outside of the convex hull of the versicolor class points, and hence covered mostly by spherical proximity regions. However, the standard cover of Figure~\ref{fig:iris_covers}(d) covers setosa and virginica classes with polygons since these classes are in the outer triangles of the convex hull of the versicolor class. 

To test the difference between the AUC of classifiers, we employ the 5x2 paired cross validation (CV) paired $t$-test and the combined 5x2 CV $F$-test \citep[see][]{dietterich1998,alpaydm1999}. The 5x2 CV test has been devised by \cite{dietterich1998} and found to be the most powerful test among those with acceptable type-I error. However, the test statistics of 5x2 $t$-tests depend on which one of the ten folds is used. Hence, \cite{alpaydm1999} offered a combined 5x2 CV $F$-test which works as an omnibus test for all ten possible 5x2 $t$-tests (for each five repetitions there are two folds, hence ten folds in total). Basically, if a majority of ten 5x2 $t$-tests suggest that two classifiers are significantly different in terms of performance, the $F$-test also suggests a significant difference. Hence, an $F$-test with high $p$-value suggests that some of the ten $t$-tests fail to reject the null-hypothesis (i.e. they have high p-value).

Recall that the number of prototypes increases exponentially with $d$ as shown by Theorem~\ref{thm:complexity}. Simulation studies in Section~\ref{sec:simulations} also indicated that the dimensionality of a data set affects the classification performance. Hence, we apply dimension reduction to mitigate the effects of dimensionality. We use principal component analysis (PCA) to extract the principal components with high variance. For \texttt{iris}, let us incorporate the first two principal components with two highest variance. We refer to this new data set with two variables as \texttt{irispc2}. The information on the four variables of \texttt{iris} data set has projected onto two dimensions, and we expect that standard cover PE-PCD classifiers works better than that in \texttt{iris} data set. 

We give the AUC measures of all classifiers on \texttt{iris13}, \texttt{iris} and \texttt{irispc2} data set in Table~\ref{tab:CVtest-AUC} and the $p$-values of the 5x2 CV $F$-test in Table~\ref{tab:CVtest-pval}. All classifiers perform well in classifying all three \texttt{iris} data sets. Although hybrid PE-PCD classifier (PE-$k$NN, PE-SVM and PE-CCCD) perform comparable to other $k$NN, SVM and CCCD classifiers, they seem to perform slightly better than the hybrid PE-PCD classifiers. Since \texttt{iris} data set and its variants in Table~\ref{tab:CVtest-AUC} are well separated and the classes are balanced, it is not surprising that $k$NN and SVM performs better. In \texttt{iris13} data set, standard cover PE-PCD classifier produces comparable AUC to other hybrid and cover PE-PCD classifiers. For example, standard cover PE-PCD classifier has nearly 0.05 AUC less than PE-$k$NN classifier in CV repetitions 1 and 3; but, on the other hand, 0.05 more AUC than PE-$k$NN in repetition 5. However, in \texttt{iris} data set, standard cover PE-PCD classifier has significantly much less AUC (about 0.1 AUC less) than other classifiers. Observe that $d=2$ in \texttt{iris13} data set, but $d=4$ in \texttt{iris} data set. Since the complexity of the class cover increases with dimensionality, the class cover of the standard cover PE-PCD classifier becomes less appealing. Although the composite cover PE-PCD classifier has substantially more AUC than standard cover PE-PCD classifier for \texttt{iris} data set, it still performs worse than the CCCD classifier. However, in \texttt{irispc2}, observe that AUC of the standard cover PE-PCD classifier has substantially increased compared to that in  \texttt{iris} data set. Obviously, the increase in the performance of standard cover PE-PCD classifiers is a result of the low dimensionality. The lower the dimension, the less the complexity of the class cover and the fewer the number of prototype sets, and thus better the classification performance. Moreover, we also report on the optimum parameters of all classifiers in Table~\ref{tab:CVtest-AUC}. It appears that, in general, $\theta$ increases, and $k$ and $\gamma$ decrease as expansion parameter $r$ increases. As reviewed in Section~\ref{sec:simulations}, the smaller the values of $k$ and $\gamma$, the higher the values of $\theta$ and $r$.

	Cover PE-PCD classifiers perform better if the data has low dimensionality. Hence, we reduce the dimensionality of data sets by means of say, PCA, and then classify the data set with the cover PE-PCD classifiers trained over this data set in the reduced dimension. The \texttt{Ionosphere} data set has 34 variables. We refer to the \texttt{Ionosphere} data set with two principal components of two highest variance as \texttt{Ionopc2}, and also, with three principal components as \texttt{Ionopc3}, and with five as \texttt{Ionopc5}. We give the AUC measures of all classifiers on these dimensionaly reduced Ionosphere data sets in Table~\ref{tab:CVtest-AUC} and the $p$-values of the 5x2 CV $F$-test in Table~\ref{tab:CVtest-pval}. In all three data sets, SVM classifiers seem to have the highest AUC values. Hybrid PE-PCD classifiers perform slightly worse compared to their corresponding classifiers which are used as alternative classifiers. However, for \texttt{Ionopc2} data set, both composite and standard cover PE-PCD classifiers have comparable performance to other classifiers. For \texttt{Ionopc3} and \texttt{Ionopc5}, on the other hand, the AUC of composite and standard cover PE-PCD classifiers relatively deteriorate compared to other classifiers. Although PE-PCD classifiers have computationally tractable MDSs and potentially have comparable performance to those other classifiers, the high dimensionality of the data sets are detrimental for these classifiers based on PE-PCD class covers. 
	
	In Table~\ref{tab:CVtest_AUC_all}, we reduce the dimensionality and classify eleven KEEL and UCL data sets with all classifiers. All data sets, except Yeast6, achieved maximum AUC when reduced to two dimensions, and for these dimensionally low data sets, standard cover PE-PCD classifiers perform, in general, comparable to other classifiers. Observe that low dimensionality mitigates the effects on the complexity of the standard cover, and hence, a relatively good classification performance is achieved. Hybrid PE-PCD classifiers usually perform slightly worse then their alternative classifier counterparts. However, the hybrid PE-PCD classifier PE-$k$NN increases the AUC of $k$-NN 0.01 AUC more.
	
\begin{table}
\centering
\caption{AUC measures of the best performing (ones with their respective optimum paramaters) hybrid and cover PE-PCD classifiers for three variants of both \texttt{iris} and \texttt{Ionosphere} data sets.}
\resizebox{\textwidth}{!}{
\begin{tabular}{cccccccccccccccccccc}
     & & \multicolumn{2}{c}{PE-$k$NN} & \multicolumn{2}{c}{$k$NN} & \multicolumn{2}{c}{PE-SVM} & \multicolumn{2}{c}{SVM} & \multicolumn{2}{c}{PE-CCCD} & \multicolumn{2}{c}{CCCD} & \multicolumn{2}{c}{Composite} & \multicolumn{2}{c}{Standard} \\
\hline \rule{0pt}{3ex}
Data & & Fo. 1 & Fo. 2 & Fo. 1 & Fo. 2 & Fo. 1 & Fo. 2 & Fo. 1 & Fo. 2 & Fo. 1 & Fo. 2 & Fo. 1 & Fo. 2 & Fo. 1 & Fo. 2 & Fo. 1 & Fo. 2\\
\hline \rule{0pt}{3ex}
     & opt. & \multicolumn{2}{c}{$r=2.8$  $k=10$} & \multicolumn{2}{c}{$k=10$} & \multicolumn{2}{c}{$r=2.8$  $\gamma=3.1$} & \multicolumn{2}{c}{$\gamma=3.1$} & \multicolumn{2}{c}{$r=2.8$  $\theta=0.1$} & \multicolumn{2}{c}{$\theta=0.1$} & \multicolumn{2}{c}{$r=2.8$  $\theta=0.1$} & \multicolumn{2}{c}{$r=2.8$} \\
\multirow{5}{*}{\texttt{iris13}}  & 1 & 0.95& 0.97& 0.93& 0.97& 0.95& 0.90& 0.96& 0.90& 0.92&  0.96&  0.92&  0.96&  0.95&  0.92&  0.92&  0.92 \\
 & 2 & 0.88& 0.96& 0.92& 0.99& 0.88& 0.96& 0.92& 0.99& 0.85&  0.96&  0.89&  0.96&  0.89&  0.97&  0.87&  0.94 \\
 & 3 & 0.96& 0.86& 0.99& 0.93& 0.96& 0.86& 1.00& 0.95& 0.96&  0.86&  0.99&  0.89&  0.97&  0.88&  0.95&  0.90 \\
 & 4 & 0.91& 0.96& 0.96& 0.96& 0.92& 0.93& 0.96& 0.93& 0.88&  0.96&  0.93&  0.96&  0.92&  0.96&  0.91&  0.95 \\
 & 5 & 0.96& 0.88& 0.95& 0.93& 0.91& 0.88& 0.87& 0.92& 0.96&  0.88&  0.92&  0.89&  0.95&  0.88&  0.93&  0.91 \\
\hline \rule{0pt}{3ex}
     & opt. & \multicolumn{2}{c}{$r=2$  $k=8$} & \multicolumn{2}{c}{$k=8$} & \multicolumn{2}{c}{$r=2$  $\gamma=0.1$} & \multicolumn{2}{c}{$\gamma=0.1$} & \multicolumn{2}{c}{$r=2$  $\theta=0.6$} & \multicolumn{2}{c}{$\theta=0.6$} & \multicolumn{2}{c}{$r=2$  $\theta=0.6$} & \multicolumn{2}{c}{$r=2$} \\
\multirow{5}{*}{\texttt{iris}}  & 1 & 0.96& 0.97& 0.97& 0.97& 0.93& 0.99& 0.95& 0.99& 0.96&  0.97&  0.97&  0.97&  0.97&  0.92&  0.76&  0.76 \\
 & 2 & 0.95& 0.97& 0.95& 0.97& 0.95& 0.97& 0.93& 0.97& 0.92&  0.97&  0.92&  0.97&  0.91&  0.97&  0.71&  0.70 \\
 & 3 & 0.97& 0.92& 0.97& 0.95& 0.97& 0.92& 0.97& 0.94& 0.97&  0.92&  0.97&  0.95&  0.97&  0.85&  0.84&  0.81 \\
 & 4 & 0.96& 0.96& 0.96& 0.97& 0.95& 0.92& 0.95& 0.92& 0.96&  0.92&  0.96&  0.93&  0.96&  0.96&  0.71&  0.76 \\
 & 5 & 0.96& 0.93& 0.97& 0.93& 0.95& 0.93& 0.96& 0.93& 0.91&  0.92&  0.91&  0.92&  0.92&  0.95&  0.75&  0.77 \\
\hline \rule{0pt}{3ex}
     & opt. & \multicolumn{2}{c}{$r=4$  $k=3$} & \multicolumn{2}{c}{$k=3$} & \multicolumn{2}{c}{$r=4$  $\gamma=0.8$} & \multicolumn{2}{c}{$\gamma=0.8$} & \multicolumn{2}{c}{$r=4$  $\theta=0.8$} & \multicolumn{2}{c}{$\theta=0.8$} & \multicolumn{2}{c}{$r=4$  $\theta=0.8$} & \multicolumn{2}{c}{$r=4$} \\
\multirow{5}{*}{\texttt{irispc2}} & 1 & 0.94& 0.89& 0.97& 0.99& 0.94& 0.86& 0.97& 0.96& 0.94&  0.89&  0.96&  0.97&  0.93&  0.91&  0.87&  0.89 \\
 & 2 & 0.92& 0.93& 0.95& 0.97& 0.92& 0.92& 0.92& 0.96& 0.92&  0.93&  0.95&  0.97&  0.92&  0.96&  0.93&  0.89 \\
 & 3 & 0.90& 0.95& 0.96& 0.96& 0.90& 0.95& 0.96& 0.96& 0.90&  0.93&  0.97&  0.96&  0.95&  0.95&  0.91&  0.95 \\
 & 4 & 0.88& 0.96& 0.94& 0.96& 0.88& 0.93& 0.92& 0.93& 0.88&  0.96&  0.94&  0.97&  0.95&  0.96&  0.95&  0.95 \\
 & 5 & 0.97& 0.90& 0.97& 0.95& 0.97& 0.90& 0.99& 0.95& 0.95&  0.90&  0.95&  0.95&  0.93&  0.93&  0.91&  0.93 \\
\hline \rule{0pt}{3ex}
     & opt. & \multicolumn{2}{c}{$r=1.3$  $k=9$} & \multicolumn{2}{c}{$k=9$} & \multicolumn{2}{c}{$r=1.3$  $\gamma=0.9$} & \multicolumn{2}{c}{$\gamma=0.9$} & \multicolumn{2}{c}{$r=1.3$  $\theta=0.1$} & \multicolumn{2}{c}{$\theta=0.1$} & \multicolumn{2}{c}{$r=1.3$  $\theta=0.1$} & \multicolumn{2}{c}{$r=1.3$} \\
\multirow{5}{*}{\texttt{Ionopc2}} & 1 & 0.75& 0.73& 0.76& 0.75& 0.77& 0.76& 0.78& 0.77& 0.76&  0.74&  0.76&  0.75&  0.72&  0.70&  0.76&  0.72 \\
 & 2 & 0.72& 0.74& 0.73& 0.78& 0.71& 0.76& 0.73& 0.79& 0.71&  0.76&  0.74&  0.76&  0.71&  0.74&  0.73&  0.76 \\
 & 3 & 0.80& 0.73& 0.82& 0.72& 0.79& 0.72& 0.82& 0.73& 0.74&  0.72&  0.74&  0.72&  0.75&  0.68&  0.78&  0.70 \\
 & 4 & 0.74& 0.76& 0.78& 0.77& 0.76& 0.73& 0.79& 0.72& 0.73&  0.75&  0.77&  0.74&  0.71&  0.73&  0.71&  0.72 \\
 & 5 & 0.75& 0.74& 0.78& 0.75& 0.75& 0.76& 0.78& 0.77& 0.74&  0.72&  0.75&  0.72&  0.74&  0.74&  0.75&  0.72 \\
\hline \rule{0pt}{3ex}
     & opt. & \multicolumn{2}{c}{$r=1.9$ $k=6$} & \multicolumn{2}{c}{$k=6$} & \multicolumn{2}{c}{$r=1.9$ $\gamma=2$} & \multicolumn{2}{c}{$\gamma=2$} & \multicolumn{2}{c}{$r=1.9$ $\theta=0.4$} & \multicolumn{2}{c}{$\theta=0.4$} & \multicolumn{2}{c}{$r=1.9$ $\theta=0.4$} & \multicolumn{2}{c}{$r=1.9$} \\
\multirow{5}{*}{\texttt{Ionopc3}} & 1 & 0.87& 0.83& 0.88& 0.84& 0.88& 0.82& 0.89& 0.82& 0.86&  0.80&  0.86&  0.80&  0.86&  0.81&  0.88&  0.80 \\
 & 2 & 0.81& 0.83& 0.81& 0.85& 0.83& 0.82& 0.84& 0.83& 0.81&  0.83&  0.83&  0.83&  0.84&  0.79&  0.81&  0.80 \\
 & 3 & 0.83& 0.81& 0.84& 0.81& 0.82& 0.86& 0.84& 0.86& 0.81&  0.84&  0.83&  0.83&  0.85&  0.84&  0.83&  0.84 \\
 & 4 & 0.79& 0.86& 0.80& 0.87& 0.86& 0.86& 0.86& 0.86& 0.83&  0.84&  0.84&  0.84&  0.84&  0.84&  0.80&  0.83 \\
 & 5 & 0.81& 0.81& 0.84& 0.81& 0.80& 0.80& 0.83& 0.80& 0.82&  0.78&  0.84&  0.79&  0.80&  0.80&  0.81&  0.78 \\
\hline \rule{0pt}{3ex}
     & opt. & \multicolumn{2}{c}{$r=1.9$ $k=4$} & \multicolumn{2}{c}{$k=4$} & \multicolumn{2}{c}{$r=1.9$ $\gamma=4$} & \multicolumn{2}{c}{$\gamma=4$} & \multicolumn{2}{c}{$r=1.9$ $\theta=0$} & \multicolumn{2}{c}{$\theta=0$} & \multicolumn{2}{c}{$r=1.9$ $\theta=0$} & \multicolumn{2}{c}{$r=1.9$} \\
\multirow{5}{*}{\texttt{Ionopc5}} & 1 & 0.88& 0.84& 0.88& 0.84& 0.94& 0.89& 0.94& 0.90& 0.92&  0.83&  0.92&  0.83&  0.87&  0.81&  0.86&  0.84 \\
 & 2 & 0.85& 0.85& 0.85& 0.85& 0.91& 0.89& 0.91& 0.89& 0.93&  0.86&  0.93&  0.86&  0.91&  0.83&  0.88&  0.83 \\
 & 3 & 0.86& 0.86& 0.86& 0.86& 0.87& 0.90& 0.87& 0.90& 0.88&  0.90&  0.88&  0.90&  0.89&  0.87&  0.84&  0.78 \\
 & 4 & 0.85& 0.88& 0.85& 0.88& 0.91& 0.89& 0.91& 0.89& 0.89&  0.87&  0.89&  0.87&  0.84&  0.88&  0.80&  0.85 \\
 & 5 & 0.84& 0.86& 0.84& 0.86& 0.91& 0.94& 0.91& 0.95& 0.89&  0.84&  0.89&  0.84&  0.84&  0.84&  0.81&  0.78 \\
\hline 
\end{tabular}
}
\label{tab:CVtest-AUC}
\end{table}

\begin{table}
\centering
\footnotesize
\caption{The $p$-values of the 5x2 CV $F$ test of AUC values in Figure~\ref{tab:CVtest-AUC}. The $p$-values below 0.1 are given in boldface.}
    \begin{tabular}{cccccccccc}
    \hline \rule{0pt}{3ex}
    \multirow{9}[0]{*}{iris13} &       & PE-$k$NN & $k$NN   & PE-SVM & SVM   & PE-CCCD & CCCD  & Composite & Standard \\
          & PE-$k$NN &       & 0,315 & 0,442 & 0,404 & 0,454 & 0,690 & 0,389 & 0,526 \\
          & $k$NN   &       &       & 0,227 & 0,498 & 0,251 & 0,545 & 0,439 & 0,506 \\
          & PE-SVM &       &       &       & 0,285 & 0,367 & 0,549 & 0,305 & 0,270 \\
          & SVM   &       &       &       &       & 0,315 & 0,420 & 0,540 & 0,447 \\
          & PE-CCCD &       &       &       &       &       & 0,482 & 0,384 & 0,434 \\
          & CCCD  &       &       &       &       &       &       & 0,780 & 0,719 \\
          & Composite &       &       &       &       &       &       &       & 0,656 \\
    \hline \rule{0pt}{3ex}
    \multirow{9}[0]{*}{iris} &       & PE-$k$NN & $k$NN   & PE-SVM & SVM   & PE-CCCD & CCCD  & Composite & Standard \\
          & PE-$k$NN &       & 0,403 & 0,627 & 0,635 & 0,402 & 0,708 & 0,628 & \textbf{0,005} \\
          & $k$NN   &       &       & 0,535 & 0,617 & 0,227 & 0,391 & 0,532 & \textbf{0,003} \\
          & PE-SVM &       &       &       & 0,433 & 0,350 & 0,641 & 0,706 & \textbf{0,020} \\
          & SVM   &       &       &       &       & 0,120 & 0,309 & 0,576 & \textbf{0,014} \\
          & PE-CCCD &       &       &       &       &       & 0,389 & 0,756 & \textbf{0,010} \\
          & CCCD  &       &       &       &       &       &       & 0,793 & \textbf{0,005} \\
          & Composite &       &       &       &       &       &       &       & \textbf{0,008} \\
    \hline \rule{0pt}{3ex}
    \multirow{9}[0]{*}{irispr2} &       & PE-$k$NN & $k$NN   & PE-SVM & SVM   & PE-CCCD & CCCD  & Composite & Standard \\
          & PE-$k$NN &       & 0,219 & 0,535 & 0,307 & 0,535 & 0,327 & 0,628 & 0,695 \\
          & $k$NN   &       &       & 0,184 & 0,205 & 0,122 & 0,386 & \textbf{0,066} & \textbf{0,081} \\
          & PE-SVM &       &       &       & 0,224 & 0,535 & 0,279 & 0,484 & 0,694 \\
          & SVM   &       &       &       &       & 0,178 & 0,356 & \textbf{0,038} & 0,196 \\
          & PE-CCCD &       &       &       &       &       & 0,186 & 0,495 & 0,676 \\
          & CCCD  &       &       &       &       &       &       & 0,133 & 0,117 \\
          & Composite &       &       &       &       &       &       &       & 0,535 \\
    \hline \rule{0pt}{3ex}
    \multirow{9}[0]{*}{Ionopr2} &       & PE-$k$NN & $k$NN   & PE-SVM & SVM   & PE-CCCD & CCCD  & Composite & Standard \\
          & PE-$k$NN &       & 0,324 & 0,528 & 0,515 & 0,328 & 0,438 & \textbf{0,000} & \textbf{0,093} \\
          & $k$NN   &       &       & 0,282 & 0,521 & 0,294 & 0,424 & \textbf{0,028} & \textbf{0,038} \\
          & PE-SVM &       &       &       & 0,398 & 0,439 & 0,435 & \textbf{0,045} & 0,343 \\
          & SVM   &       &       &       &       & 0,419 & 0,434 & 0,137 & 0,301 \\
          & PE-CCCD &       &       &       &       &       & 0,589 & 0,130 & 0,574 \\
          & CCCD  &       &       &       &       &       &       & 0,182 & 0,467 \\
          & Composite &       &       &       &       &       &       &       & 0,118 \\
    \hline \rule{0pt}{3ex}
    \multirow{9}[0]{*}{Ionopr3} &       & PE-$k$NN & $k$NN   & PE-SVM & SVM   & PE-CCCD & CCCD  & Composite & Standard \\
          & PE-$k$NN &       & 0,430 & 0,638 & 0,507 & 0,727 & 0,693 & 0,655 & 0,656 \\
          & $k$NN   &       &       & 0,672 & 0,620 & 0,542 & 0,594 & 0,631 & 0,479 \\
          & PE-SVM &       &       &       & 0,434 & 0,420 & 0,617 & 0,610 & 0,154 \\
          & SVM   &       &       &       &       & \textbf{0,074} & 0,108 & 0,350 & \textbf{0,014} \\
          & PE-CCCD &       &       &       &       &       & 0,578 & 0,732 & 0,486 \\
          & CCCD  &       &       &       &       &       &       & 0,659 & 0,282 \\
          & Composite &       &       &       &       &       &       &       & 0,584 \\
    \hline \rule{0pt}{3ex}
    \multirow{9}[0]{*}{Ionopr5} &       & PE-$k$NN & $k$NN   & PE-SVM & SVM   & PE-CCCD & CCCD  & Composite & Standard \\
          & PE-$k$NN &       & 0,500 & \textbf{0,022} & \textbf{0,020} & 0,548 & 0,548 & 0,618 & 0,223 \\
          & $k$NN   &       &       & \textbf{0,022} & \textbf{0,020} & 0,548 & 0,548 & 0,618 & 0,223 \\
          & PE-SVM &       &       &       & 0,535 & 0,324 & 0,324 & \textbf{0,096} & \textbf{0,062} \\
          & SVM   &       &       &       &       & 0,338 & 0,338 & \textbf{0,094} & \textbf{0,061} \\
          & PE-CCCD &       &       &       &       &       & 0,500 & 0,452 & 0,168 \\
          & CCCD  &       &       &       &       &       &       & 0,452 & 0,168 \\
          & Composite &       &       &       &       &       &       &       & \textbf{0,076} \\
    \hline \rule{0pt}{3ex}
    \end{tabular}
\label{tab:CVtest-pval}
\end{table} 

\begin{table}
  \centering
  \footnotesize
  \caption{Average of ten folds of 5x2 CV $F$-test AUC values of all classifiers on eleven KEEL and UCL data sets. The symbol ``*" indicate a difference with the AUC of standard cover PE-PCD classifier at significant level of 0.1, and ``**" at level 0.05. ``PC$d$" indicates the number of principal components used. }
    \begin{tabular}{cccccccccc}
    Data  & PC $d$ & PE-$k$NN & $k$NN & PE-SVM & SVM   & PE-CCCD & CCCD  & Composite & Standard \\
    \hline
    \texttt{iris} & 2     & 0,924 & 0,962* & 0,918 & 0,952 & 0,920 & 0,959 & 0,939 & 0,918 \\
    \texttt{Ionosphere} & 2     & 0,747* & 0,763** & 0,752 & 0,767 & 0,737 & 0,746 & 0,720 & 0,735 \\
    \texttt{New-Thyroid1} & 2     & 0,962 & 0,965 & 0,947 & 0,950 & 0,960 & 0,963 & 0,966 & 0,963 \\
    \texttt{New-Thyroid2} & 2     & 0,977 & 0,977 & 0,948 & 0,948 & 0,986 & 0,986 & 0,986 & 0,969 \\
    \texttt{Segment0} &       &       &       &       &       &       &       &       &  \\
    \texttt{Shuttle0vs4} & 2     & 0,995 & 0,995 & 1,000 & 1,000 & 0,998 & 0,998 & 0,998 & 0,997 \\
    \texttt{Wine} & 2     & 0,974** & 0,975** & 0,971** & 0,972** & 0,965 & 0,965 & 0,955 & 0,950 \\
    \texttt{Yeast4} & 2     & 0,579 & 0,588 & 0,555 & 0,504** & 0,569 & 0,564 & 0,562 & 0,553 \\
    \texttt{Yeast5} & 2     & 0,711 & 0,721 & 0,675 & 0,624 & 0,688 & 0,683 & 0,668 & 0,666 \\
    \texttt{Yeast6} & 3     & 0,687* & 0,676* & 0,621 & 0,557 & 0,655 & 0,641 & 0,594 & 0,613 \\
    \texttt{Yeast1289vs7} & 2     & 0,559 & 0,547 & 0,548 & 0,503 & 0,552 & 0,535 & 0,546 & 0,549 \\
    \hline
    \end{tabular} 
    
    \bigskip
    
    \begin{tabular}{cccccccc}
    Data  & $N$ & $d$ & $q=m/n$ & $k$ & $\gamma$ & $\theta$ & $r$ \\
    \hline
    \texttt{iris} & 150   & 4     & 2,00  & 3     & 0,8   & 0,8   & 4,0 \\
    \texttt{Ionosphere} & 351   & 35    & 1,78  & 9     & 0,9   & 0,1   & 1,3 \\
    \texttt{New-Thyroid1} & 215   & 5     & 5,14  & 5     & 2,5   & 1,0   & 2,5 \\
    \texttt{New-Thyroid2} & 215   & 5     & 5,14  & 4     & 3,5   & 1,0   & 4,0 \\
    \texttt{Segment0} & 2308  & 20    & 6,02  &       &       &       &  \\
    \texttt{Shuttle0vs4} & 1829  & 10    & 13,87 & 1     & 0,1   & 0,2   & 1,1 \\
    \texttt{Wine} & 178   & 13    & ~2,00 & 24    & 1,0   & 0,4   & 1,4 \\
    \texttt{Yeast4} & 1484  & 9     & 28,10 & 1     & 0,7   & 1,0   & 7,0 \\
    \texttt{Yeast5} & 1484  & 9     & 32,70 & 6     & 2,5   & 1,0   & 1,0 \\
    \texttt{Yeast6} & 1484  & 9     & 41,40 & 1     & 2,3   & 1,0   & 9,0 \\
    \texttt{Yeast1289vs7} & 1484  & 9     & 30,70 & 1     & 4,0   & 0,3   & 4,0 \\
    \hline
    \end{tabular}

  \label{tab:CVtest_AUC_all}
\end{table}

\section{Summary and Discussion} \label{sec:discussion}

We use \emph{proximity catch digraphs} (PCDs) to construct semi-parametric classifiers. These families of random geometric digraphs constitute class covers of a class of interest (i.e. the target class) in order to generate decision-boundaries for classifiers. PCDs are generalized versions of Class Cover Catch Digraphs (CCCDs). For imbalanced data sets, CCCDs showed better performance than some other commonly used classifiers in previous studies \citep{manukyan2016,devinney2002}. CCCDs are actually examples of PCDs with spherical proximity maps. Our PCDs, however, are based on simplical proximity maps, e.g. proportional-edge (PE) proximity maps. Our PCD, or PE-PCD, class covers are extended to be unions of simplical and polygonal regions whereas original PE-PCD class covers were composed of only simplicial regions. The most important advantage of these family of PE proximity maps is that their respective digraphs, or namely PE-PCDs, have computationally tractable minimum dominating sets (MDSs). The class covers of such digraphs are minimum in complexity, offering maximum reduction of the entire data set with comparable and, potentially, better classification performance. 

The PE-PCDs are defined on the Delaunay tessellation of the non-target class (i.e. the class not of interest). PE-PCDs, and associated proximity maps, were only defined for the points inside of the convex hull of the non-target class, $C_H(\X_{1-j})$, in previous studies. Here, we introduce the outer simplices associated with facets of $C_H(\X_{1-j})$ and thus extend the definition of the PE proximity maps to these outer simplices. Hence, the class covers of PE-PCDs apply for all points of the target class $\X_j$. PE-PCDs are based on the regions of simplices associated with the vertices of these simplices, called $M$-vertex regions. We characterize these vertex regions with barycentric coordinates of target class points with respect to the vertices of the $d$-simplices. However, the barycentric coordinates only apply for the target class points inside the $C_H(\X_{1-j})$. For those points outside the convex hull, we may incorporate the generalized barycentric coordinates of, for example, \cite{warren1996}. Such coordinate systems are convenient for locating points outside $C_H(\X_{1-j})$  since outer simplices are similar to convex $d$-polytopes even though they are unbounded. However, generalized barycentric coordinates of the points with respect to these convex polytopes are not unique. Hence, properties on MDSs and convex distance measures are not well-defined. 

PE-PCD class covers are low in complexity; that is, by finding the MDSs of these PE-PCDs, we can construct class covers with minimum number of proximity regions. The minimum dominating set, or the prototype set, is viewed as a reduced data set that potentially increases the testing speed of a classifier. CCCDs have the same properties, but only for data sets in $\R$. By extending outer intervals, i.e. intervals with infinite end points, to outer simplices in $\R^d$ for $d>1$, we established classifiers having the same appealing properties of CCCDs in $\R$. The expansion parameter $r$ of the PE proximity maps substantially decreases the cardinality of the minimum dominating set, but the classification performance decreases for very large $r$. Hence, an optimal choice of $r$ value is in order. On the other hand, the complexity of the prototype set increases exponentially with $d$, the dimensionality of the data set. This fact is due to the Delaunay tessellation of the non-target class since the number of simplices and facets increases exponentially on $d$ (see Theorem~\ref{thm:complexity}). Therefore, these class covers become inconvenient for modelling the support of the class for high $d$. We employ dimensionality reduction, e.g. principal components analysis, to mitigate the effects of the dimensionality. Hence, the classification performance substantially increases with these dimensionally reduced data sets as shown in Section~\ref{sec:realdata}. The Monte Carlo simulations and experiments in Section~\ref{sec:simulations} also indicate that PE-PCDs have good reduction percentage in lower dimensions. 

We define two types of classifiers based on PE-PCDs, namely hybrid and cover PE-PCD classifiers. In hybrid PE-PCD classifiers, alternative classifiers are used when PE-PCD pre-classifiers are unable to make a decision on a query point. These pre-classifiers are only defined by the simplices provided in the Delaunay tesselation of the set $\X_{1-j}$, hence only for target class points in $C_H(\X_{1-j})$. We considered alternative classifiers $k$-NN, SVM and CCCD. The cover PE-PCD classifiers, on the other hand, are based on two types of covers: \emph{composite} covers where the target class points inside and outside of the convex hull of the non-target class are covered with separate proximity regions, and \emph{standard} covers where all points are covered with regions based on the same family of proximity maps. For composite covers, we consider a composition of spherical proximity maps (used in CCCDs) and PE proximity maps. Results on both hybrid and cover PE-PCD classifiers indicate that when the dimensionality is low and classes are imbalanced, standard cover PE-PCD classifiers achieve either comparable or slightly better classification performance than others. We show that these classifiers are better in classifying the minority class in particular. This makes cover PE-PCD classifiers appealing since they present slightly better performance than other classifiers (including hybrid PE-PCD classifiers) with a high reduction in the data set. 

PE-PCDs offer classifiers of (exact) minimum complexity based on estimation of the class supports. The MDSs are computationally tractable, and hence, the maximum reduction is achieved in polynomial time (on the size of the training data set). This property of PE-PCDs, however, achieved by partitioning of $\R^d$ by Delaunay tessellation, and as a result, the number of the simplices and facets of the convex hull of the non-target class determines the complexity of the model which increases exponentially fast with the dimensionality of the data set. Indeed, this leads to an overfitting of the data set. We employ PCA to extract the features with the most variation, and thus reduce the dimensions to mitigate the effects of dimensionality. PCA, however, is one of the oldest dimensionality reduction method, and there are many dimension reduction methods in the literature that may potentially increase the classification performance of PCD classifiers. Moreover, PE-PCDs are one of many family of PCDs using simplicial proximity maps investigated in \cite{ceyhan2010}. Their construction is also based on the Delaunay tessellations of the non-target class, and similar to PE-PCDs, they enjoy some other properties of CCCDs in $\R$, and they can also be used to establish PCD classifiers.  However, our work proves the idea that relatively good performing classifiers with minimum prototype sets can be provided with PCDs based on partitioning schemes (e.g. Delaunay tesselations), but we believe an alternative partitioning method, say for example a rectangular partitioning scheme, that produces less partitioning than a Delaunay tessellation would be more appealing for the class cover. Such schemes could also have computationally tractable MDSs. Such classifiers and their classification performance are topics of ongoing research. 

\section*{Acknowledgments}
Most of the Monte Carlo simulations presented in this article were executed at Ko\c{c} University High Performance Computing Laboratory.

\section{Appendix}

\subsection{Proof of Theorem~\ref{thm:vertex_rd_bary}}

We prove this theorem by induction on dimension $d$. The proof of the case $d=1$ is trivial. For $\mathfrak S(\Y)=(\y_1,\y_2) \subset \R$ and $\y_1 < \y_2$, the vertex regions $R_M(\y_1)$ and $R_M(\y_2)$ are the intervals $(\y_1,M)$ and $(M,\y_2)$, respectively ($\{x=M\}$ and $\{x=\y_i\}$ have zero $\R$-Lebesgue measure). For $\al_1 \in (0,1)$ and $\al_2=1-\al_1$, let $\al_1 \y_1 + \al_2 \y_2$ be the convex (or barycentric) combination of $x \in \mathfrak S(\Y)$. Hence, $x \in (\y_1,M)=R_M(\y_1)$ if and only if $\al_1 / \al_2 > m_1/m_2$. The case $d=2$ is proved in Proposition~\ref{prop:vertex_r2_bary}. Thus, there only remains the case $d>2$. We suppose the statement is true for all faces of the $d$-simplices which are $d-1$ dimensional, and by that, we will show that the statement is also true for the $d$-simplex which is $d$ dimensional. 

It is sufficient to show the result for $\y_1$ (as the others follow by symmetry). Let $x \in R_M(\y_1)$ and note that the elements of the set of $(d-1)$-faces, $\{f_j\}^{d+1}_{j=2}$, are adjacent to $\y_1$. Each of these faces are of $d-1$ dimensions. Hence, they are  $(d-1)$-simplices and they also have their own vertex regions. Thus, let $R_{M_i}(\y_j,f_i)$ be the vertex region of $\y_j$ with respect to $(d-1)$-simplex $f_i$ for $j \neq i$. Note that $M_i$ is the center of $f_i$. Now, let $w_{f_i}(z,\y_j) = w_{ij} $ be the barycentric coordinate of point $z$ corresponding to $\y_j$ with respect to the $f_i$. Observe that $w_{ii}$ is not defined since $\y_i$ is not a vertex of the face $f_i$. 

Moreover, let $\mathbf{m}'=(m'_1,\ldots,m'_{i-1},m'_{i+1},\ldots,m'_{d+1})$ be the barycentric coordinates of $M_i$ with respect to $f_i$, and note that $M_i$ is a linear combination of $M$ and $\y_i$. Also, observe that $m'_i$ is not defined since the vertex $\y_i$ is not a vertex of $f_i$. Hence, for $\beta \geq 1$,
\begin{align} \label{equ:M_i}
M_i &= \beta M + (1-\beta) \y_i = \beta \left( \sum_{t=1;t \neq i}^{d+1} m_t \y_t \right) + (1-\beta) \y_i. 
\end{align}
Therefore, by the uniqueness of barycentric coordinates, $m'_t=\beta m_t$ for $t=1,\ldots,d+1$ and $t \neq i$. Note that $(1-\beta)=0$ since $M_i \in f_i$ and also $f_i \subset \partial(\mathfrak S(\Y))$. Hence, $\beta=1$ which implies $m'_t=m_t$ for all $t \neq i$. Then, $m'_1/m'_j=m_1/m_j$ for $j=2,3,\ldots,d+1$ and $j \neq i$. We use this result on our induction hypothesis. 

Now, for $i=2,\ldots,d+1$, let the face $f_i$ and line defined by $x$ and $\y_i$ cross at the point $z_i$. Observe that $z_i \in f_i$, and since $f_i$ is a $(d-1)$-simplex and $x \in R_M(\y_1)$, see that $z_i \in R_{M_i}(\y_1,f_i)$. By induction hypothesis and (\ref{equ:M_i}), we observe that $z_i \in R_{M_i}(\y_1,f_i)$ if and only if $w_{i1} > (m'_1/m'_j) w_{ij}$ if and only if $w_{i1} > (m_1/m_j) w_{ij}$ for $j=2,3,\ldots,d+1$ and $j \neq i$. Since the point $x$ is the convex (and linear) combination of $z_i$ and $\y_i$, for $\alpha \in (0,1)$, we have
\begin{align*}
	x = (1-\al) \y_i + \al z_i = (1-\al) \y_i + \al \left( \sum_{k=1;k \neq i}^{d+1} w_{ik} \right).
\end{align*}
\noindent By the uniqueness property of barycentric coordinates, it follows that $w_{\mathfrak S}^{(1)}(x) =  \al w_{i1}$ and $w_{\mathfrak S}^{(j)}(x) =  \al w_{ij}$. Hence,  
\begin{equation} \label{equ:w_x}
	\frac{w_{\mathfrak S}^{(1)}(x)}{w_{\mathfrak S}^{(j)}(x)} = \frac{w_{i1}}{w_{ij}} > \frac{m_1}{m_j}.
\end{equation}
\noindent Since (\ref{equ:w_x}) is true for all $i=2,\ldots,d+1$, we see that $x \in R_M(y_1)$ if and only if $ w_{\mathfrak S}^{(1)}(x) > (m_1/m_i) w_{\mathfrak S}^{(i)}(x)$. Hence, the result follows. $\blacksquare$

\section{Acronyms and Abbreviations}

\begin{tabular}{ll}
\textbf{PCD} & Proximity catch digraph \\
\textbf{CCCD} & Class cover catch digraph \\
\textbf{PE} & Proportional edge \\
\textbf{PE-PCD} & Proportional edge PCD \\
\textbf{CCP} & Class cover problem \\
\textbf{MDS} & Minimum dominating set \\
\textbf{SVM} & Support vector machine \\
\textbf{RBF} & Radial basis function \\
\textbf{AUC} & Area under curve \\
\textbf{CCR} & Correct Classification Rate \\
\textbf{$k$-NN} & $k$ nearest neighbor \\
\textbf{CV} & Cross validation \\
\textbf{PCA} & Principal Components Analysis \\
\textbf{PE-$k$NN} & Hybrid PE-PCD classifier with alternative classifier $k$-NN \\
\textbf{PE-SVM} & Hybrid PE-PCD classifier with alternative classifier SVM \\
\textbf{PE-CCCD} & Hybrid PE-PCD classifier with alternative classifier CCCD \\
\end{tabular}

\bibliography{References2}
\bibliographystyle{apalike}

\end{document}